\documentclass{article}
\usepackage{xcolor}
\usepackage[utf8]{inputenc}
\usepackage{tabularx}
\usepackage{booktabs}
\usepackage{graphicx}
\usepackage{amssymb,amsfonts,amsmath,amsthm,bm}
\usepackage{hyperref}
\usepackage{cleveref}
\usepackage{subcaption}
\usepackage{booktabs}
\usepackage{microtype}
\usepackage{titletoc}
\usepackage{graphicx,array,pifont}
\usepackage[dvipsnames]{xcolor}

\usepackage{tabularx}
\newcolumntype{Y}{>{\raggedright\arraybackslash}X}
\usepackage{multirow}
\usepackage{enumitem}
\usepackage{wrapfig}
\usepackage{adjustbox}
\usepackage[table]{xcolor}
\let\SavedAddContentsLine\addcontentsline  
\usepackage{iclr2027_conference,times}
\let\addcontentsline\SavedAddContentsLine 
\usepackage[table]{xcolor}
\usepackage{tabularray}
\usepackage{float}
\UseTblrLibrary{booktabs}
\title{Learned Image Compression for\\ Vision-Language-Action Models}

\iclrfinalcopy % Uncomment for camera-ready version, but NOT for submission.
\newcommand{\authormail}[1]{\makebox[0.42\textwidth][l]{\texttt{#1}}}

\author{%
Hyeonjun Kim \\ POSTECH \\ \authormail{kim.hyeonjun@postech.ac.kr}
\And
Jegwang Ryu \\ POSTECH \\ \authormail{jegwang.ryu@postech.ac.kr}
\AND
Sangbeom Ha \\ POSTECH \\ \authormail{sangbeomha@postech.ac.kr}
\And
Junhyeok Lee \\ Soongsil University \\ \authormail{wnsx0000@gmail.com}
\AND
Jun-Hyuk Kim \\ Chung-Ang University \\ \authormail{junhyukkim@cau.ac.kr}
\And
Hyemin Ahn \\ POSTECH \\ \authormail{hmahn@postech.ac.kr}
\AND
Jaeho Lee \\ POSTECH \\ \authormail{jaeho.lee@postech.ac.kr}
}

\usepackage{amsmath,amsfonts,bm}

\def\eqref#1{equation~\ref{#1}}
\def\1{\bm{1}}

\DeclareMathAlphabet{\mathsfit}{\encodingdefault}{\sfdefault}{m}{sl}
\SetMathAlphabet{\mathsfit}{bold}{\encodingdefault}{\sfdefault}{bx}{n}

\usepackage{url}

\begin{document}
\maketitle
%===============================================================================
%\input{Figures_tex/intro}
\begin{abstract}
Vision-language-action (VLA) models increasingly rely on high-frame-rate multi-camera observations, making visual communication a major bottleneck for real-time robotic control in bandwidth-constrained or distributed deployment settings. Existing image and video codecs, however, are designed to preserve generic visual fidelity rather than the control performance of downstream VLA policies. In this work, we introduce SPARC (\underline{SP}atially \underline{A}daptive \underline{R}ate \underline{C}ontrol), a learned image compression framework tailored for VLA-driven robots. Our key observation is that the importance of visual information varies substantially across both camera views and spatial regions within an image frame. Based on this observation, SPARC employs a lightweight temporal mask selector that uses the current and preceding frames to decide which latent entries to prioritize, thereby allocating bitrate across views and regions according to task relevance. We further introduce a tilted rate loss that stabilizes training by reducing the tendency of entropy-based rate objectives to over-suppress rare yet task-critical visual patterns. Experiments on RoboCasa365, VLABench, and LIBERO with three VLA models show that SPARC consistently achieves better bitrate-success tradeoff than conventional image/video codecs and recent learned compression methods. We additionally demonstrate real-world deployment benefits in remote-control settings, where SPARC substantially improves the bitrate-success tradeoff.

%We introduce a novel image compression method to compress visual signals for downstream vision-language-action models (VLAs). Although VLA models have been developed for serving generalist policies, their real-time deployment often requires high-frequency multi-camera observations, which can be difficult to transmit under limited communication bandwidth or resource-constrained client settings. To address this problem, we focus on neural image compression and formalize our new problem that compresses images for robots. Based on our formulation, we introduce our new algorithm \textbf{SPARC} (Spatially Adaptive Rate Control For Robots) that filters out unnecessary information in images. In image compression for vision--language--action models, specifically, we consider that adaptively controlling compression rates on (i) multiple cameras and (ii) unnecessary pixel-level information in images are important for maintaining the performance of VLA models. Our result shows that our codec is above other neural codec baselines in robotics benchmarks, such as VLABench, Robocasa, and LIBERO. 
\end{abstract}

% We introduce \textbf{SPARC} (Latent Compression for Robots), a neural image compression framework for transmitting visual signals to downstream vision--language--action (VLA) models. Although VLA models enable generalist robotic policies, their real-time deployment often requires high-frequency multi-camera observations, which can be difficult to transmit under limited communication bandwidth or resource-constrained client settings. This motivates a VLA-oriented compression problem: reducing visual transmission cost while preserving the information necessary for action prediction. Unlike generic image compression, this setting requires adaptive compression across camera views and spatial regions, since not all visual information is equally relevant to the VLA policy. Based on this insight, SPARC operates in the latent space of a neural image codec and selectively suppresses task-irrelevant information before transmission. In particular, we propose a \textit{temporal mask selector}, which learns to identify unimportant latent components for the VLA model and adaptively controls the compression rate over time. Our framework provides a practical direction for efficient visual communication in robot systems that rely on remote or bandwidth-limited VLA inference.

\section{Introduction} \label{sec:intro}

%\subsection{Emphasize that server-client communication framework is essential for vla communication}

% 문단1: VLA기반 로봇제어가 각광받고 있는데, 여기에서 Server-Client간 통신이 필수불가결함. 왜냐하면 robot 상에 gpu 올리는 게 어려운 케이스… 카메라와 gpu가 리모트한 경우도 있음…
% Emphasize that server-client communication framework is essential for vla communication
% kick off sentence: How much visual information does a robot truly need to act effectively?
% 문단2: 통신비용이 극심하게 높은 (내지는 critical한) 경우도 있음. 예컨대:
% Extremely Fast: Low-latency로 동작하는 경우 (e.g., safety-critical)
% Extremely Large data: 멀티 카메라
% Extremeley Low Channel Capacity: 통신 밴드윗쓰 자체가 작은 경우
% State that communication budget is extremely high at certainc cases.

How much visual information does a robot truly need to act effectively? This question is becoming increasingly important as vision-language-action (VLA) models emerge as a promising paradigm for general-purpose robotic control. Modern VLAs consume high-frame-rate visual streams, often from multiple cameras, to generate actions in real time~\citep{octo,openvla-oft,OpenVLA,pi0,pi05, gr00tn1_2025}.

%카메라와 GPU가 떨어져 있는게 더 맞는 statement ==> 그리고 robot이랑 카메라가 떨어져 있더는 것을 보여준다. Split-computting은 예시로 보낸다.
% In practical deployments, VLA inference is frequently split between an edge robot and a remote GPU server due to limited onboard compute, centralized infrastructure, weight confidentiality, and wireless camera setup. As a result, robots are required to continuously stream visual observations to the server, making visual communication a major bottleneck. As illustrated in \Cref{fig:motivation}, this challenge is particularly severe in three regimes: large-scale multi-robot deployments with network contention such as in factories, multi-camera robotic systems with high visual throughput, and communication-constrained environments such as underwater or space robotics. In these settings, reducing visual transmission cost is essential for scalable real-time control. Despite this, robotic visual observations are still typically compressed using generic image or video codecs such as JPEG~\citep{wallace1992jpeg} or H.265~\citep{sullivan2012overview}, which are designed to preserve perceptual image fidelity rather than downstream control performance.

\begin{figure}[t] % <-- 시작 부분 괄호 수정
    \centering
    \begin{subfigure}[t]{0.4\linewidth}
        \centering
        \includegraphics[width=\linewidth]{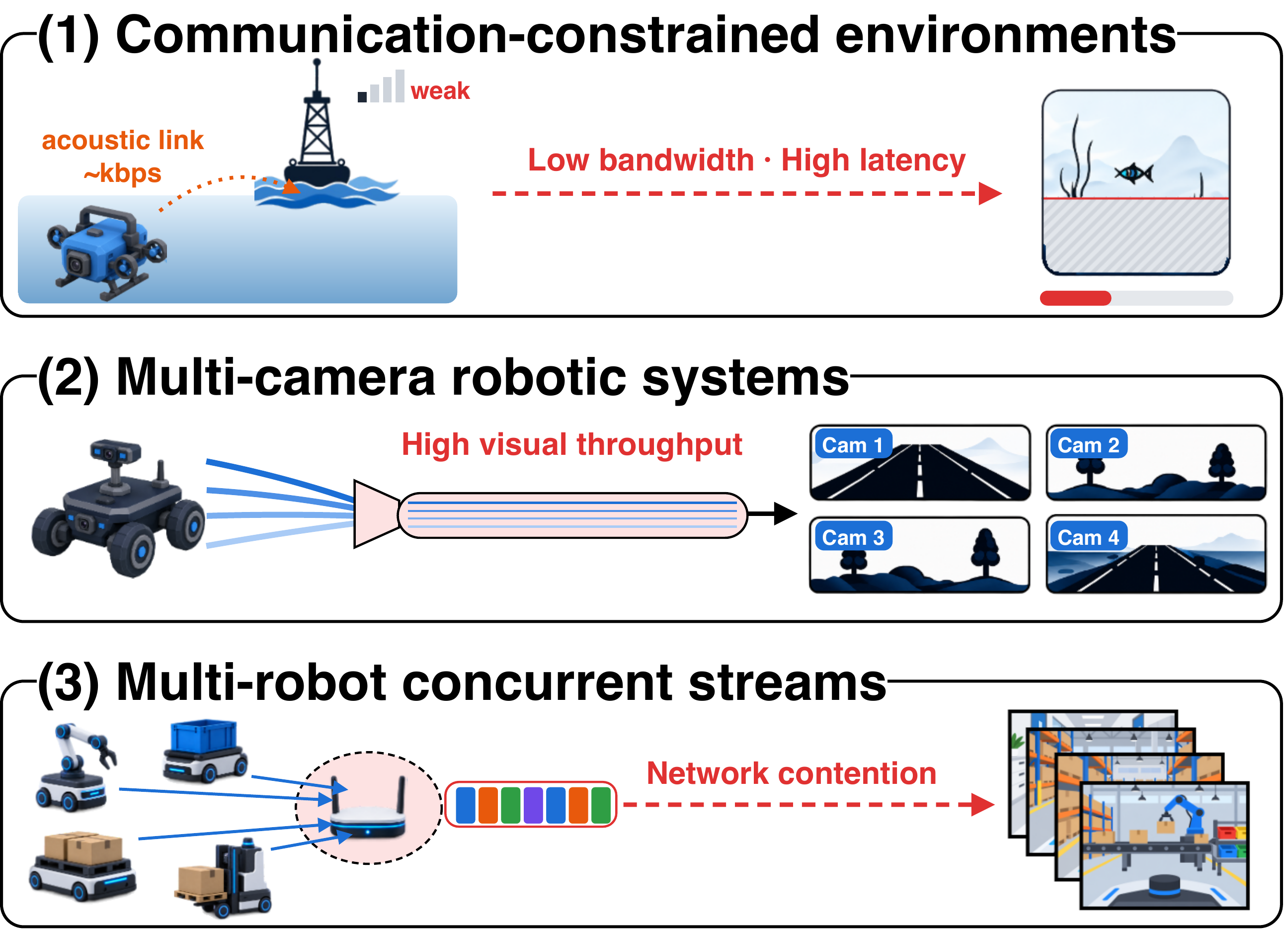}
        \caption{Motivating scenarios}
        \label{fig:motivation}
    \end{subfigure}%
    \hfill%
    \begin{subfigure}[t]{0.58\linewidth}
        \centering
        \captionsetup{justification=centering}
        \includegraphics[width=\linewidth]{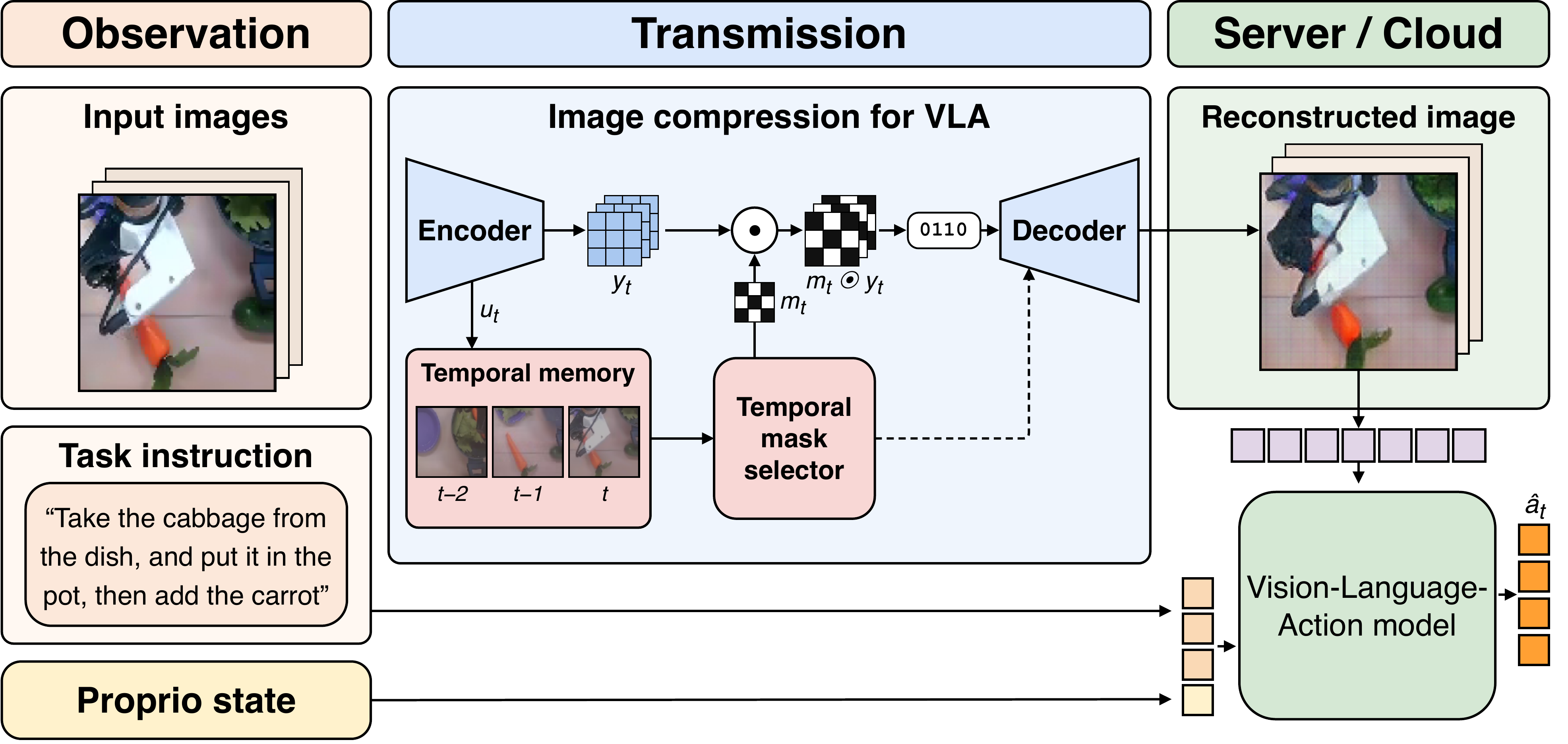}
        \caption{Framework of learned image compression for VLA}
        \label{fig:motivation_framework}
    \end{subfigure}
    \caption{\textbf{Motivation and overview of SPARC.} (a) Visual transmission becomes a bottleneck in communication-constrained environments, multi-camera robotic systems, and multi-robot deployments sharing a network. (b) Trained with the action loss rather than pixel-level distortion, SPARC predicts a task-aware latent mask from temporal memory and transmits only the unmasked entries.}
    \label{fig:motivation_overview}
\end{figure}

In many practical scenarios, VLA inference relies on cameras that stream visual observations to a remotely located computing unit. As illustrated in \Cref{fig:motivation}, transmitting these visual streams incurs substantial transmission costs that hinder scalable real-time control, particularly in three scenarios: communication-constrained environments such as underwater or space robotics, robotic systems utilizing massive number of camera inputs, and multi-robot deployments with network contention such as in factories. In these settings, reducing the transmission cost of visual signals is essential for scalable real-time control. Yet, robotic visual observations are still compressed using generic image or video codecs such as JPEG~\citep{wallace1992jpeg} or H.265~\citep{sullivan2012overview}. These codecs na\"{\i}vely allocate more bitrate to regions with more details to preserve perceptual image fidelity, while spatial regions with high complexity may not necessarily be relevant for the robotic task. 
% Recent studies verified task-relevant regions for robotic control \citep{xu2026vlacache}.
Consequently, the bitrate allocation strategies of existing codecs may be inefficient for downstream control.

% 문단3: 지금은 꽤 나이브하게 하고 있음. 로봇태스크의 특성을 살리지 못하고 일반적인 이미지와 같이 압축함. 구체적으로, 요런요런 기회들이 낭비되고 있음:
% Camera-wise: 
% Pixel-wise: 
% Currently, JPEG, H.264, or any other standards.. methods are used, but they are naive, not specifically optimal when it comes to communication in robotics (Camera-wise, Pixel-wise)

%첫 문장을 문제제기로 하기. 더 room이 있을 것 같다. Head room이 있다. 더 잘 할 수 있다는 주장 등등...
%Downstream performance으로 더 잘 할수 있게 하는 여지가 있을까 ==> Recent observation에 따르면 task에 따라서 다른 중요도를 가져간다.

%다음문단 부터는 정황증거가 어떤게 있는지, 어떻게 이를 이용하는지 보이기.
% We argue that compression for VLAs should prioritize information relevant for action generation to the generation quality of images. This distinction is particularly important because visual importance in VLAs is highly non-uniform. Modern VLAs often operate on multi-camera inputs, where different viewpoints contribute unequally to control \citep{mandlekar2021what}. Moreover, within each image, only a subset of spatial regions may be relevant for the current task \citep{xu2026vlacache}. In light of these observations, we believe that current image compression methods are not optimal for VLAs.
We believe that visual compression for VLAs should consider two major opportunities: localized task importance across cameras and spatial regions, and temporal sparsity due to action chunking. These points are particularly important because visual information a VLA relies on is highly non-uniform in both space and time. Spatially, modern VLAs often operate on multi-camera inputs, where different viewpoints contribute unequally to control \citep{mandlekar2021what}. Moreover, within each camera input, only a subset of spatial regions may be relevant to the current task \citep{xu2026vlacache}. Temporally, action chunking means new observations are available every few control steps during inference. The resulting gaps between transmitted frames make inter-frame change large and highly non-uniform, breaking the frame-to-frame continuity that conventional video codecs rely on.

We propose \textbf{SPARC} (\underline{SP}atially \underline{A}daptive \underline{R}ate \underline{C}ontrol), a learned image compression framework tailored for VLA systems.
%SPARC라는 모델은 importance를 잘 알고 있는 mask selector를 활용한다. 이걸 먼저
Reflecting highly localized task importance, SPARC performs policy-aware masking directly in the latent space of a neural image codec to allocate high bitrates to task-relevant regions. Moreover, it conditions on previously transmitted observations to compensate for the sparse temporal sampling induced by action chunking.  In practice, it employs a lightweight temporal mask selector that predicts element-wise latent masks ($h$$\times$$w$$\times$$c$) from the temporal context, allowing adaptive bitrate allocation between camera streams and image regions. To train this adaptive masking mechanism stably under a VLA-guided rate objective, we introduce a tilted rate loss that prevents entropy-based optimization from aggressively suppressing statistically rare yet task-critical visual patterns. Built on top of a pretrained neural image codec, SPARC is trained end-to-end to directly optimize downstream action quality under a communication budget.

%Unlike prior codecs that optimize generic image fidelity~{\color{red}[X,X,..]} or rely on external spatial priors~{\color{red}[X,X,..]}, SPARC adaptively allocates bitrate according to the needs of the downstream VLA policy. Its core component is a temporal mask selector that predicts binary masks over image latents, enabling the codec to transmit only task-relevant latent regions. Built on top of a pretrained neural image codec, SPARC is trained with a VLA-guided objective that jointly minimizes action distortion and the total transmission rate.

We validate SPARC on both simulation environments and real-world remote control deployments. On simulation benchmarks, such as RoboCasa365, VLABench, and LIBERO, SPARC consistently achieves stronger bitrate-success tradeoffs than conventional image/video codecs and recent learned compression methods. In real-world deployments, we employ split-computing architecture between local CPU and remote GPU server to verify that SPARC consistently achieves better success rates than MS-ILLM while using substantially lower bitrate. We show that reduced communication overhead translates into lower end-to-end latency in bandwidth-constrained deployments.

% 다음과 같은 문장으로 result 결과 설명 시작.
% We verify the generalizability of this learning framework..
% We validate the effectiveness of this data by ... leading to improvement..

% result 등 추가, pi0과 pi0.5 등에서 같은 성공률 기준 약 3배에서 ~ 5배 정도의 BPP 감소를 기록하였다.

% \input{Figures_tex/intro}

% 문단 6: 커다란 implication 혹은 3-bullet 컨트리뷰션 요약
% Explain results and 3-bullet summary
% we propose로 시작, key considerations를 고려해.. temporal mask selector로 camera-wise, spatially region-wise인 codec을 제안했다. In our best knowledge, 이렇게 adaptive compression을 for VLAs로 적용한 시도는 이게 처음이다.
% 어떤 실험을 통해 ~한 점에서 이점이 있다는 것을 보였다.
% VLA-guided objective를 사용해 downstream VLA task가 need for하는 부분에 adaptive하게 bitrate을 주기로 했다.

% best tradoff
Our contributions are summarized as follows:

\begin{itemize}[leftmargin=*,topsep=0pt,parsep=0pt]
    \item We propose SPARC, the first learned image compression framework tailored for VLAs.
    \item We introduce a temporally conditioned latent masking framework for compression, combining a temporal mask selector for adaptive bitrate allocation and a tilted rate loss for stable optimization.
    \item We demonstrate superior bitrate-success tradeoffs and lower end-to-end latency across simulation and real-world robotic deployments.
\end{itemize}

\section{Related Work} \label{sec:related_work}

\textbf{Neural compression of visual signals.}
Compared with traditional codecs such as JPEG~\citep{wallace1992jpeg} and H.265~\citep{sullivan2012overview}, learned compression methods utilize neural networks optimized under rate-distortion objectives~\citep{balle2017end,yang2023introduction,balle2018variational,minnen2018joint,mentzer2020high,yang2023lossy}. For example, prior work has improved compression quality through learned hyperpriors~\citep{balle2018variational,minnen2018joint} and generative reconstruction techniques~\citep{mentzer2020high,yang2023lossy}. However, many of these methods are trained for a fixed compression rate, making it difficult to support multiple rates with a single model. To this end, variable-rate compression methods train a single model that supports multiple bitrates using iterative refinement, conditioning, or latent selection~\citep{toderici2015variable,toderici2017full,choi2019variable,lee2022selective}. Beyond global rate control, spatially adaptive codecs allocate different bit budgets to different image regions using external priors such as ROI masks or quality maps~\citep{ma2021variable,song2021variable}. Yet, this reliance on external priors limits the applicability of existing methods to real-time VLA systems, where task-relevant regions should ideally be inferred automatically. SPARC therefore learns VLA-guided bit allocation across image regions without relying on external spatial priors.

\textbf{Neural image compression for machines.}
A growing body of work adapts image compression for machine vision by using downstream-task losses instead of reconstruction distortion, enabling codecs to preserve task-relevant information rather than pixel fidelity~\citep{duan2020video}. Representative approaches include content-adaptive codecs, ROI-based bit allocation with segmentation priors, and unified human-machine compression~\citep{le2021learned,10647785,feng2026difficmh}. Nevertheless, prior codecs for machines mainly target perception models that map images to labels, regions, or text, rather than policies that predict continuous robot actions. In contrast, to the best of our knowledge, SPARC is the first neural image codec designed specifically for VLA systems.

\section{Problem Formulation} \label{sec:formulation}

We formalize image compression for VLA models as follows.
Suppose we have $K$ cameras that capture a set of images $\mathbf{x}_t = \{\mathbf{x}_{t,i}\}_{i=1}^K$ at each timestep $t = 1,\ldots,T$. Let $\mathbf{l}$ denote the language instruction for the task. At time $t$, the VLA policy generates an action $\mathbf{a}_t$ according to $ \pi(\cdot | \mathbf{x}_{1:t}, \mathbf{l},\mathbf{a}_{1:t-1})$. The policy is typically trained to minimize a loss measuring the quality of the generated action sequence $\mathbf{a}_{1:T}$. For example, given expert demonstrations $\mathbf{a}_{1:T}^*$, an off-policy objective may be
\begin{align}
\min_{\pi} \frac{1}{T}\sum_{t=1}^T \mathbb{E}[\ell(\mathbf{a}_t,\mathbf{a}^*_t)] \quad \mathrm{where} \quad \mathbf{a}_t \sim \pi(\cdot|\mathbf{x}_{1:t},\mathbf{l},\mathbf{a}^*_{1:t-1}),
\end{align}
where $\ell(\cdot,\cdot)$ is a loss function on actions, such as the flow matching loss.

Now consider \textit{compressing} the image stream $\mathbf{x}_t$ before it is consumed by the VLA. Specifically, let $(f_{\theta},g_{\phi})$ denote an encoder-decoder pair parameterized by $(\theta,\phi)$. The encoder maps each image into a compact discrete latent representation $\mathbf{y}_{t,i} = f_{\theta}(\mathbf{x}_{t,i})$, and the decoder reconstructs the image as $\hat{\mathbf{x}}_{t,i} = g_{\phi}(\mathbf{y}_{t,i})$. In practice, the encoder may run on an edge device co-located with the cameras, while the decoder may run on a GPU server hosting the VLA model.

Our objective is to minimize the VLA loss under the compressed image stream $\{\hat{\mathbf{x}}_{t}\}_{t=1}^T$ while enforcing a compact latent representation $\mathbf{y}$. We formulate this objective as a Lagrangian relaxation of the rate-distortion objective under the off-policy setting:
\begin{align}
\min_{\theta,\phi, \psi} \frac{1}{T}\sum_{t=1}^T \mathbb{E}\Big[\underbrace{\ell(\hat{\mathbf{a}}_t,\mathbf{a}^*_t)}_{\text{action distortion}}
+
\lambda
\underbrace{
\Big(\sum_{i=1}^K-\log p_{\psi}(\mathbf{y}_{t,i})\Big)
}_{\text{total rate}}
\Big],\qquad
\begin{aligned}
% \mathbf{a}_t \sim \pi(\cdot|\mathbf{x}_{1:t},\mathbf{l},\mathbf{a}_{1:t-1})\\ \quad \hat{\mathbf{a}}_t \sim \pi(\cdot|\hat{\mathbf{x}}_{1:t},\mathbf{l},\mathbf{a}_{1:t-1})
\hat{\mathbf{a}}_t \sim \pi(\cdot|\hat{\mathbf{x}}_{1:t},\mathbf{l},\hat{\mathbf{a}}_{1:t-1})
\end{aligned} \label{eq:lag}
\end{align}
where $p_{\psi}$ is a learnable entropy model over the latent, and $\lambda > 0$ controls the tradeoff between compression rate and action quality. Note that logarithms are base 2 throughout the paper.

% \textbf{Properties.} A key distinction from prior work is twofold. First, the compression quality is measured by the action distortion $\ell(\mathbf{a}_t,\hat{\mathbf{a}}_t)$, rather than image-level distortion $d(\mathbf{x}_t,\hat{\mathbf{x}}_t)$ or distortions computed from non-sequential downstream predictors $d(h(\mathbf{x}_{t}),h(\hat{\mathbf{x}}_{t}))$. Second, we minimize the aggregated total rate over the cameras, instead of applying the same rate individually. 

\textbf{Properties.} A key distinction from prior work is twofold. First, the compression quality is measured by the action distortion $\ell(\hat{\mathbf{a}}_t,\mathbf{a}^*_t)$, rather than image-level distortion $d(\mathbf{x}_t,\hat{\mathbf{x}}_t)$ or distortions computed from non-sequential downstream predictors $d(h(\mathbf{x}_{t}),h(\hat{\mathbf{x}}_{t}))$. Second, we minimize the aggregated total rate over the cameras, instead of applying the same rate individually.

\section{Method} \label{sec:method}
\begin{figure}[!t] % <-- 시작 부분 괄호 수정
    \centering
    \begin{subfigure}[t]{0.68\linewidth}
        \centering
        \includegraphics[width=\linewidth]{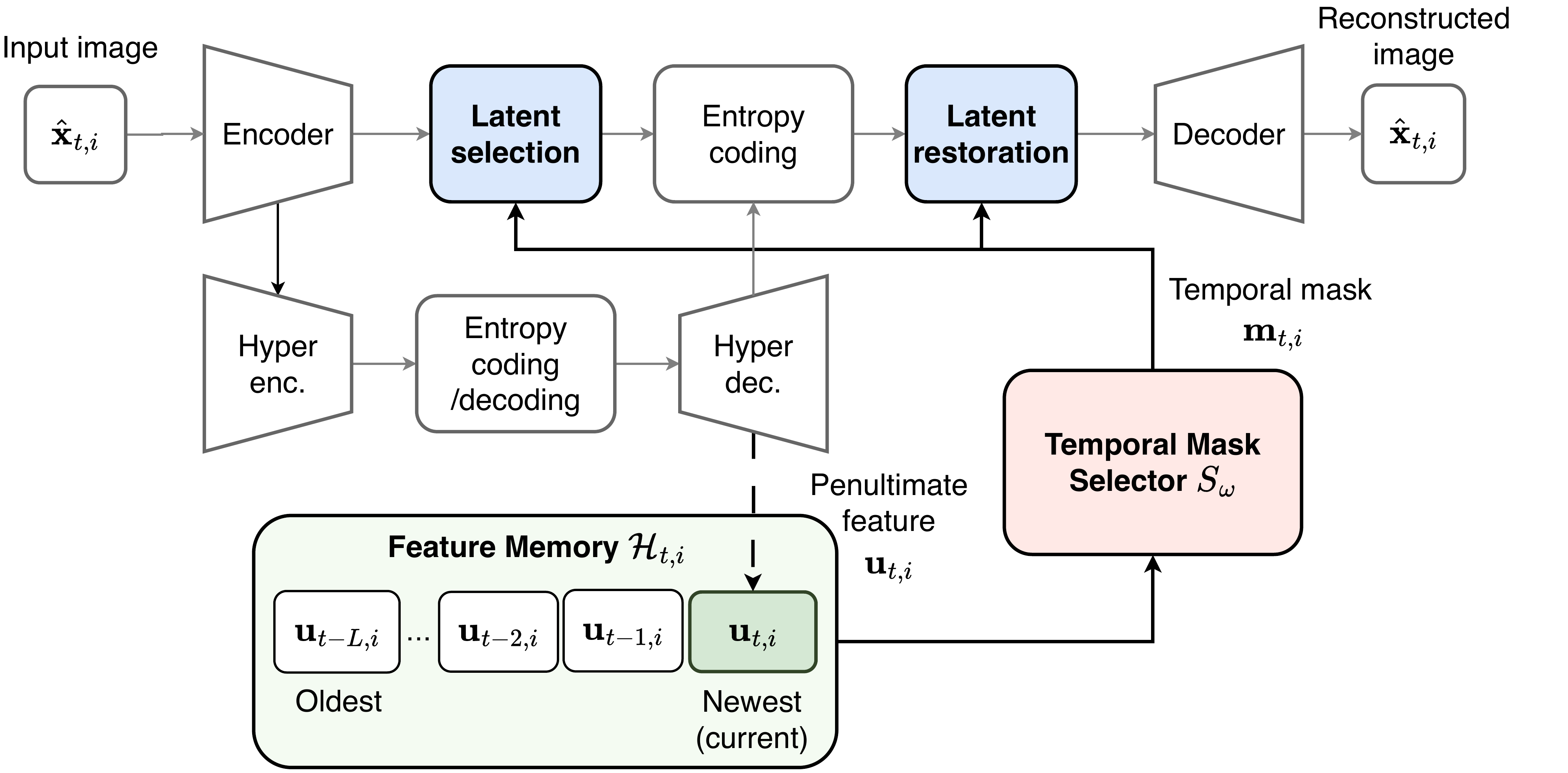}
        \caption{Overall architecture of SPARC}
        \label{fig:architecture_overall}
    \end{subfigure}%
    \hfill%
    \begin{subfigure}[t]{0.30\linewidth}
        \centering
        \captionsetup{justification=centering}
        \includegraphics[width=0.77\linewidth]{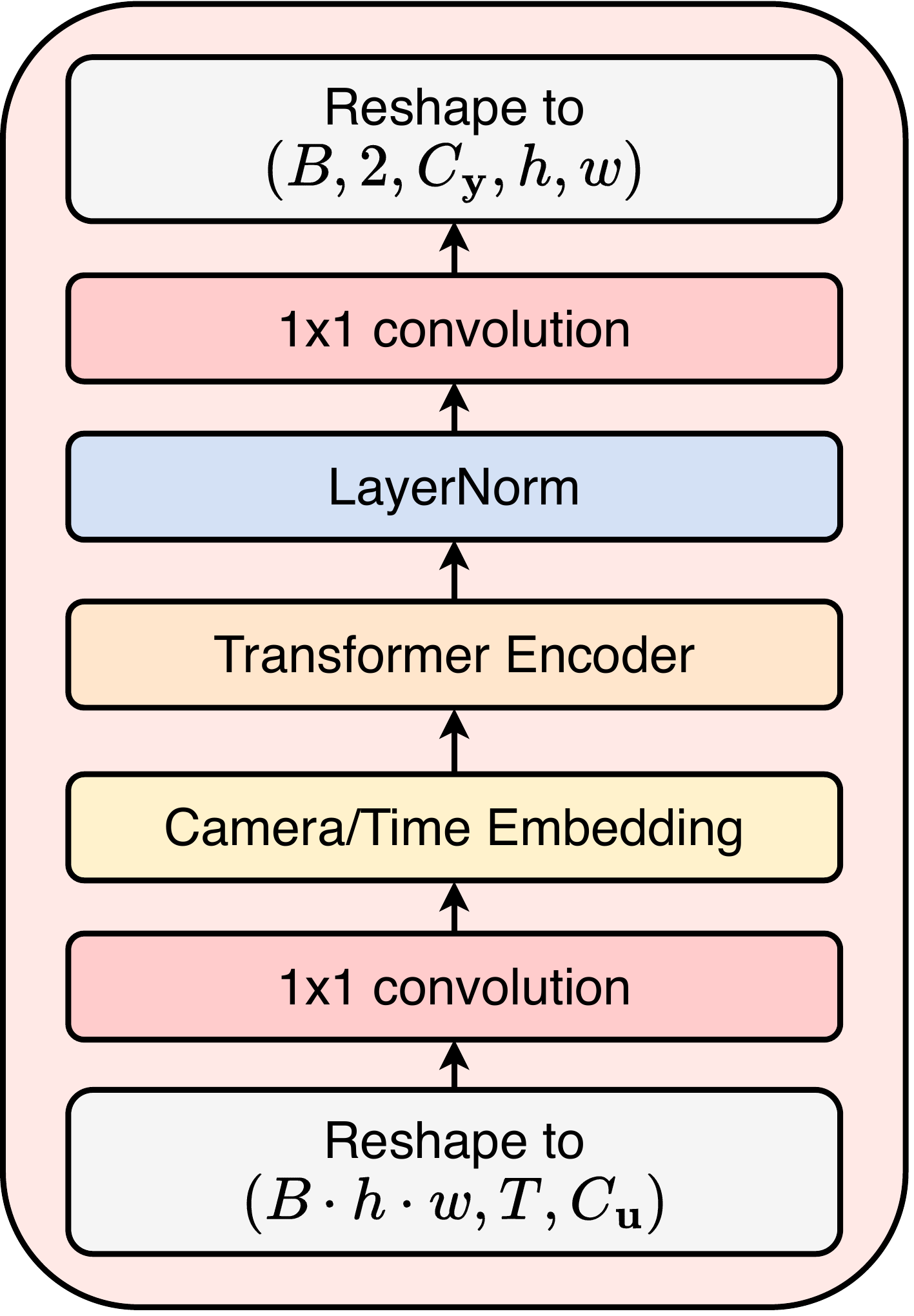}
        \caption{Temporal mask selector}
        \label{fig:selector_internals}
    \end{subfigure}
    \caption{
        \textbf{Architecture of SPARC.} 
        (a) SPARC builds on an autoencoder architecture with hyperpriors, incorporating a temporal mask selector which takes a history of hyperprior features ($\mathcal{H}_{t,i}$) and generates a spatial binary mask ($\mathbf{m}_{t,i}$) to selectively mask the latent. 
        (b) Temporal mask selector is composed of a lightweight transformer, incorporating the camera and time embedding.
        }
    \label{fig:architecture_diagram}
\end{figure}

We propose a learned image codec framework for VLAs: \textbf{SPARC} (\underline{SP}atially \underline{A}daptive \underline{R}ate \underline{C}ontrol).
Our design is motivated by the characteristics of images commonly encountered in VLA systems, as illustrated in \Cref{fig:architecture_diagram}. In particular, we observe that the importance of visual information varies significantly both across cameras and across spatial regions within an image. To leverage this property, SPARC adaptively assigns different compression rates to different regions based on their task relevance. Concretely, SPARC employs a lightweight temporal mask selector operating on latent representations (\Cref{ssec:architecture}), trained with a specialized objective (\Cref{ssec:training}).

% \begin{figure}[!t]
%     \centering
%     \includegraphics[trim=1cm 0cm 1cm 0cm, clip, width=1.0\linewidth]{"Figures/temporal_mask_selector_architecture.drawio (7).png"}
%     \caption{Temporary Main Architecture Design}
%     \label{fig:architecture}
% \end{figure}

\subsection{Architecture} \label{ssec:architecture} 

Motivated by their success in neural image compression, SPARC adopts an autoencoder architecture with hyperpriors (\Cref{fig:architecture_diagram}). Specifically, we build upon the architecture of \citet{balle2018variational}, and introduce a temporal mask selector that modulates the latent representation on a per-region basis.

\textbf{Temporal mask selector.} Given a latent representation $\mathbf{y}_{t,i} \in \mathbb{R}^{h \times w \times c}$, the temporal mask selector $S_{\omega}$ produces a binary mask $\mathbf{m}_{t,i} \in \{0,1\}^{h \times w \times c}$ of the same shape, indicating which entries of $\mathbf{y}_{t,i}$ should be retained. We then apply elementwise masking, $\mathbf{m}_{t,i} \odot \mathbf{y}_{t,i}$, so that only the unmasked entries are stored and transmitted, reducing the required bitrate.

To generate the mask, $S_{\omega}$ leverages information from past observations. Specifically, we maintain a feature memory $\mathcal{H}_{t,i} = \{\mathbf{u}_{t-L,i}, \ldots, \mathbf{u}_{t-1,i}, \mathbf{u}_{t,i}\}$, where $L$ denotes the history length, and $\mathbf{u}_{t,i}$ denotes penultimate-layer feature of hyperprior decoder. Such features are known to contain informative signals about image content \citep{lee2022selective}. The features in $\mathcal{H}_{t,i}$ are concatenated, projected, and augmented with learned temporal and camera embeddings. The resulting representation is processed by a lightweight transformer. Finally, the output is binarized into $\mathbf{m}_{t,i}$ using Gumbel-Softmax \citep{jang2017categorical}, allowing gradients to propagate through the discretization step during training.

\textbf{Transmission.} During transmission, SPARC sends two components: the entropy-coded image latent containing only the unmasked entries, and the hyper-latent. The binary mask $\mathbf{m}_{t,i}$ itself is not transmitted, avoiding additional communication overhead. Instead, the mask is deterministically rematerialized at the receiver side (i.e., GPU side) from the hyper-latent, using the same temporal mask selector and feature memory $\mathcal{H}_{t,i}$, producing the identical mask \(\mathbf{m}_{t,i}\) at both the sender and receiver. The receiver then restores the sparse latent by inserting zeros at masked locations, yielding $\mathbf{m}_{t,i}\odot\mathbf{y}_{t,i}$. This procedure is applied independently to each camera stream.

\subsection{Training} \label{ssec:training}
To leverage a codec pretrained on large-scale and diverse visual data, we initialize SPARC from a pretrained neural image codec, namely MS-ILLM \citep{muckley2023improving}. Specifically, we augment the pretrained model with the temporal mask selector $S_\omega$ and the feature memory $\mathcal{H}$.

% Due to the limited availability of robot-centric image datasets, we initialize SPARC from a pretrained neural image codec, namely MS-ILLM \citep{muckley2023improving}, which was trained on a large-scale image dataset. Specifically, we augment the pretrained model with the temporal mask selector $S_\omega$ and the feature memory $\mathcal{H}$, and train the resulting system in two phases.

% \textbf{Phase 1: Decoder warm-up.} We first warm up the decoder to align it with the target VLA before introducing any masking-induced rate pressure. Concretely, we train the decoder for $W$ steps while freezing all other modules and disabling masking. During this phase, the training objective reduces to the action distortion loss $\mathbb{E}[\ell(\mathbf{a}_t,\hat{\mathbf{a}}_t)]$ from Equation~\ref{eq:lag}. Empirically, we find that this warm-up stage significantly improves training stability during subsequent optimization.

\textbf{Joint optimization with tilted rate loss.}
% After warm-up, 
We jointly optimize the codec decoder and the temporal mask selector using both the action distortion and the rate loss. To stabilize training, we introduce a modified rate objective, termed the \textbf{\textit{tilted rate loss}}, which prevents the masking policy from being overly biased toward regions with high local bitrate.

Let $\mathbf{y} \in \mathbb{R}^{N}$ denote a (flattened) latent representation with $N = h \times w \times c$ entries. We define the local bit vector $\mathbf{b} \in \mathbb{R}^N$ as the number of bits required to encode each latent entry:
\begin{align}
\mathbf{b} = \big(-\log p_{\psi}(y_1),\ldots,-\log p_{\psi}(y_N)\big),
\end{align}
where $p_{\psi}$ is the entropy model and $y_j$ is the $j$-th entry of $\mathbf{y}$. Using this notation, the original rate loss is $\ell_{\mathrm{rate}}(\mathbf{y}) := \tfrac{1}{N}\sum_{j=1}^N b_j$. Applying a binary mask $\mathbf{m} \in \{0,1\}^{N}$ yields the masked rate loss $\ell_{\mathrm{rate}}(\mathbf{y};\mathbf{m}) := \frac{1}{N}\sum_{j=1}^N m_j b_j$.

However, this objective implicitly encourages the temporal mask selector to suppress entries with the largest $b_j$, i.e., entries that are least likely under the entropy model. Such entries often correspond to rare or unusual visual patterns, which can nevertheless be highly important for robotic decision-making. Excessively masking these regions may therefore lead to unstable or degraded training.

To mitigate this issue, we introduce the tilted rate loss. Specifically, we define an exponentiated version of local bit vector:
\begin{align}
\mathbf{b}^{(\alpha)} = \big(b_1^{\alpha},\ldots,b_N^{\alpha}\big), \qquad \alpha \in [0,1],
\end{align}
which compresses the dynamic range of the bit penalties. We then replace the original penalty $\mathbf{b}$ with the scaled penalty $\mathbf{b}^{(\alpha)}$ when computing the masked rate:
\begin{align}
\ell_{\mathrm{rate}}^{(\alpha)}(\mathbf{y};\mathbf{m}) = \left(\frac{1}{N}\sum_{j=1}^N m_j b^{\alpha}_j\right) \cdot \frac{\sum_{j=1}^N b_j}{\sum_{j=1}^N b_j^\alpha}. \label{eq:bpp_masked}
\end{align}
The second term acts as a normalization factor, ensuring that the overall scale of the rate loss remains comparable across different values of $\alpha$. The tilted rate loss is computed independently for each camera stream, summed, and incorporated into the overall optimization objective in Equation \ref{eq:lag}.

\section{Experiments} \label{sec:experiment}
To validate our ideas, we conduct several experiments on diverse simulation benchmarks, compression models and VLA models, as described in \Cref{ssec:setup}. We compare our method with several traditional and learned image compression methods and video compression methods (\Cref{ssec:simulation}) and present several visual examples (\Cref{ssec:qualitative}). Additional ablation results are provided in \Cref{sec:ablations}.
\subsection{Experimental Setup}\label{ssec:setup}
% \begin{figure}[t]
%     \centering
%     \includegraphics[width=0.96\linewidth]{Figures/all_codec_bpp_success_tradeoffs_legend_boxed.pdf}
%     \caption{\textbf{Bitrate--success rates on diverse simulation benchmarks and models.} We report average simulation success rates over a range of average bpps (bits per pixel). To maintain the fairness of comparisons, we calculate average bpp by aggregating bpp statistics of all frames in every episode. For SPARC, we average over 3 seeds to guarantee the reliability of SPARC's performance. }
%     \label{fig:main_result}
% \end{figure}
% \usepackage{subcaption}   % cleveref보다 먼저
% \usepackage{booktabs}

\begin{figure}[t]
\centering
\small
\begin{subfigure}[b]{\linewidth}
  \centering
  \begin{tabular}{@{}lccccc@{}}
    \toprule
    & \multicolumn{3}{c}{RoboCasa365} & VLABench & LIBERO \\
    \cmidrule(lr){2-4} \cmidrule(lr){5-5} \cmidrule(l){6-6}
    VLA model & $\pi_{0.5}$ & $\pi_{0}$ & GR00T N1.6 & $\pi_{0.5}$ & $\pi_{0.5}$ \\
    \midrule
    SPARC vs.\ MS-ILLM & $-43.2\%$ & $-70.4\%$ & $-24.0\%$ & $-17.2\%$ & $-18.7\%$ \\
    \bottomrule
  \end{tabular}
  \caption{BD-rate of SPARC relative to MS-ILLM. Negative values indicate bitrate savings.}
  \label{fig:main_result_bdrate}
\end{subfigure}
\\[1.0em]
\begin{subfigure}[b]{\linewidth}
  \centering
  \includegraphics[width=0.96\linewidth]{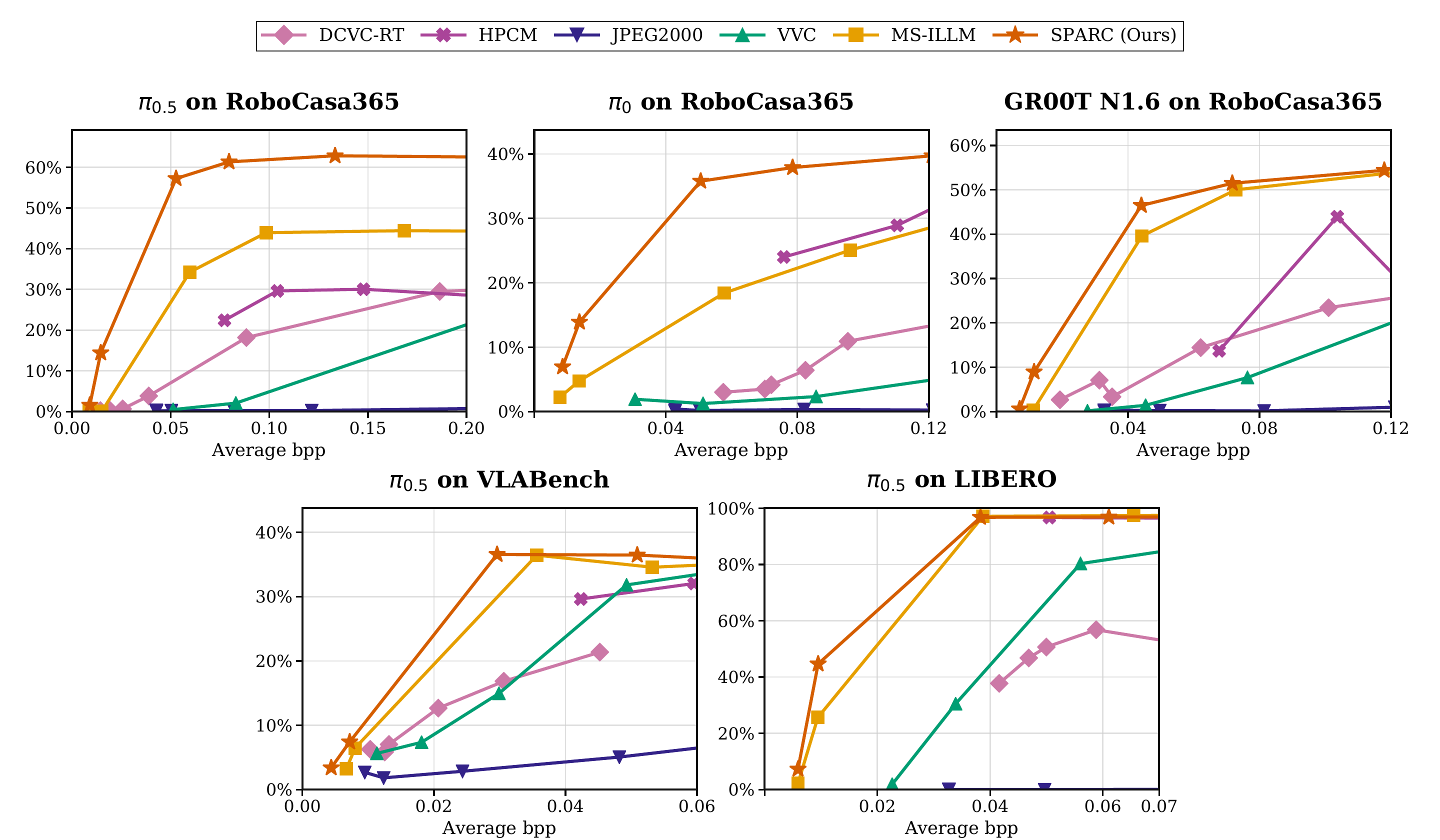}
  \caption{Bitrate--success trade-off across benchmarks and VLA models.}
  \label{fig:main_result_curves}
\end{subfigure}

\caption{\textbf{Bitrate--success rates on diverse simulation benchmarks and models.}
We report average simulation success rates over a range of average bpps (bits per pixel).
To maintain the fairness of comparisons, we calculate average bpp by aggregating bpp
statistics of all frames in every episode. For SPARC, we average over 3 seeds to guarantee
the reliability of SPARC's performance.}
\label{fig:main_result}
\end{figure}
\textbf{Benchmarks.} We evaluate SPARC on three standard simulation benchmarks: (1) RoboCasa365~\citep{robocasa365}: We use a VLA model fine-tuned on the Human300 pre-training tasks and report the average success rate across all 24 atomic tasks. (2) VLABench~\citep{zhang2025vlabench}: We evaluate the official VLA checkpoint fine-tuned on the 10 primitive tasks under the in-distribution setting. (3) LIBERO~\citep{libero}: We employ a checkpoint fine-tuned on all four task suites (Spatial, Object, Goal, and Long) and report the overall average success rate across them.

\textbf{Baselines.} We compare SPARC against multiple compression methods, categorized into two groups: (1) Image codecs: JPEG2000~\citep{christopoulos2000jpeg2000}, a traditional handcrafted codec, VVC~\citep{bross2021overview}, a traditional standard codec, as well as MS-ILLM~\citep{muckley2023improving}, and HPCM~\citep{Li_2025_ICCV}, which are recent learned neural codecs, and (2) Video codecs: DCVC-RT~\citep{jia2025towards}, a powerful learned real-time codec. For downstream VLA models, we use $\pi_{0}$~\citep{pi0}, $\pi_{0.5}$~\citep{pi05} and GR00T N1.6~\citep{gr00tn1_2025}. See \Cref{sec:method:impl} for more details.

\textbf{Training details.} 
% We initialize our codec from a pretrained MS-ILLM model and fine-tune it using our training recipe. 
Across all datasets, we use a batch size of 12 and train for 15,000 steps. We report bpp based on the size of latent and hyper-latent representations. Also, VLA model-specific pre-processing is applied after visual signals are compressed. For each benchmark, SPARC is trained on the same dataset used to train the corresponding VLA model and the action distortion loss is equivalent to the action loss (i.e. flow-matching loss) used for fine-tuning VLA models. In terms of hyper-parameters, we tune $\alpha$ over ${0.0, 0.2, 0.4, 0.6, 0.8, 1.0}$ and set the temporal history to $1$. For the rate-distortion trade-off, we use \(\lambda \in \{1.0, 10.0, 20.0, 50.0\}\), with larger \(\lambda\) values fine-tuned from pre-trained checkpoints targeting higher-quality operating points to reach lower-bpp regimes. See \Cref{ssec:add_exp_details} for more details. 

\textbf{Inference details.} 
For MS-ILLM and HPCM, we utilize the official publicly available checkpoints. For both DCVC-RT and VVC, we select the best GOP from \{1, 64\}. VVC uses VVenC 1.13.0/VVdeC 3.0.0. See \Cref{ssec:add_exp_details} for more details.
\subsection{Simulation Results}\label{ssec:simulation} 
In \Cref{fig:main_result}, SPARC consistently achieves higher simulation success rates under comparable bpp budgets across diverse VLA models and benchmarks. Compared with conventional codecs and learned codecs, SPARC reaches strong performance at lower bpps, indicating that preserving task-relevant visual information is more important than generic image fidelity. For RoboCasa365 benchmark, SPARC shows strong performance on $\pi_0$ and $\pi_{0.5}$. For VLABench and LIBERO, SPARC remains competitive and consistently lies on the upper envelope of the bitrate-success curve.
\begin{figure}[t]
    \centering
    \includegraphics[width=1.0\linewidth]{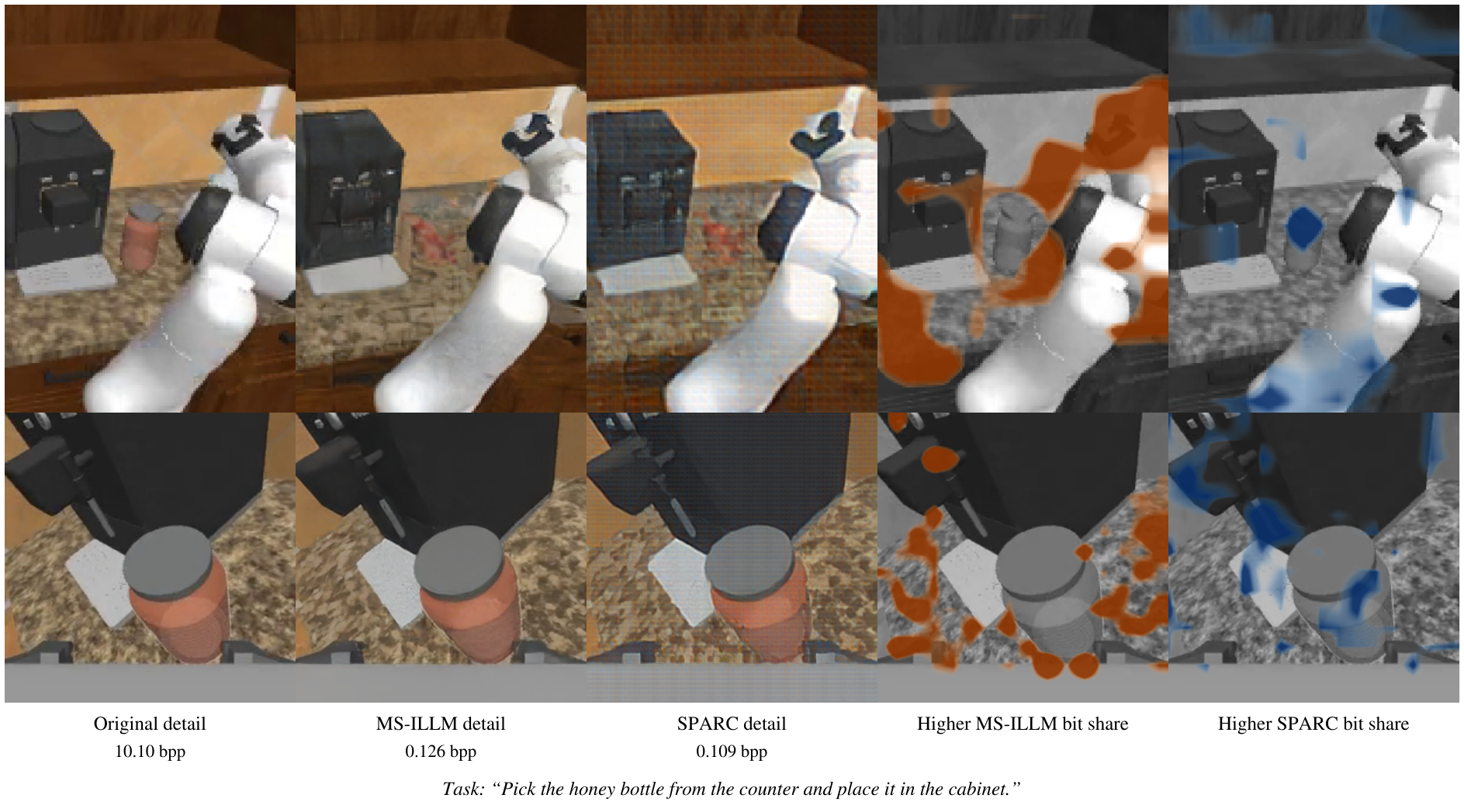}
    \caption{\textbf{Qualitative results on RoboCasa365.} In relative bit allocation map, \textcolor{blue}{Blue} indicates spatial regions where SPARC allocates more bits under an identical total bit budget than MS-ILLM. \textcolor{orange}{Orange} indicates spatial regions where MS-ILLM allocates more bits than SPARC.}
    \label{fig:qualitative}
\end{figure}
\subsection{Qualitative Examples} \label{ssec:qualitative}
In \Cref{fig:qualitative}, we compare visual samples of images from MS-ILLM and SPARC. While MS-ILLM and SPARC produce visually similar images at equivalent compression rates, their bit allocation strategies differ significantly. SPARC prioritizes task-critical regions, such as the gripper and target objects (e.g. an orange can), to facilitate robotic control. Conversely, MS-ILLM distributes bits uniformly across the entire image to optimize global quality, wasting bandwidth on task-irrelevant backgrounds like table textures. More examples can be found in \Cref{sec:ave}.
\subsection{Ablation studies} \label{ssec:ablation_studies}
To verify the contribution of our learned masking strategies to performance, the following ablation studies were conducted on RoboCasa365 with $\pi_{0.5}$, a benchmark with the most dexterous tasks and abundant visual signals. All reported results are averaged over 3 seeds. Other ablation studies can be found in \Cref{sec:ablations}.
\begin{table}[t]
\centering
\caption{\textbf{SPARC vs. other masking strategies.}
We compare SPARC against four alternative masking strategies applied at inference. Keep indicates the keep ratio of the mask. \textit{Current-only} suppresses the past-frame token in the selector.
\textit{Saliency} keeps the top-$K$ latents by the hyperprior's scale $\sigma$.
\textit{Spatially Uniform} applies a fixed deterministic pattern, and
\textit{Random} selects an arbitrary subset.}
\label{tab:masking_strategies}

\resizebox{\linewidth}{!}{%
\begin{tabular}{lrrrrrrrrr}
\toprule
Masking
& Keep & bpp & Succ. (\%)
& Keep & bpp & Succ. (\%)
& Keep & bpp & Succ. (\%) \\
\midrule

Current-only
& 0.235 & 0.015 & 14.2
& 0.667 & 0.053 & 54.4
& 0.644 & 0.080 & 59.2 \\

Saliency
& 0.236 & 0.014 & 0.3
& 0.667 & 0.054 & 10.9
& 0.635 & 0.090 & 8.8 \\

Spatially Uniform
& 0.236 & 0.008 & 0.2
& 0.667 & 0.041 & 0.1
& 0.635 & 0.065 & 1.2 \\

Random
& 0.236 & 0.008 & 0.0
& 0.667 & 0.042 & 0.6
& 0.635 & 0.064 & 0.8 \\

\rowcolor{gray!15}
\textbf{SPARC} (Ours)
& 0.236 & 0.015 & \textbf{14.4}
& 0.667 & 0.053 & \textbf{57.2}
& 0.635 & 0.080 & \textbf{61.3} \\

\bottomrule
\end{tabular}%
}
\end{table}

\paragraph{Masking strategy.} 
In \Cref{tab:masking_strategies}, we compare SPARC against alternative masking strategies to verify that SPARC identifies and preserves task-critical regions rather than masking arbitrarily. To isolate the effect of the masking strategy alone, we fix the trained SPARC model and apply each alternative mask only at inference. As shown, SPARC achieves the highest success rate, outperforming rule-based strategies, as well as saliency-based approaches. These results confirm that SPARC's hyper-latent-driven masking is not incidental but is essential to its performance.
%Notably, the saliency mask, despite being derived from task-relevant signals, performs poorly, suggesting that generic saliency does not align with the compression-aware task-critical regions that SPARC learns to retain.
%
\begin{table}[!t]
  \centering
  \caption{\textbf{Temporal history ablation on RoboCasa
  $\boldsymbol{\pi}_{0.5}$.}
  We analyze how changing temporal history affects the performance.}
  \label{tab:robocasa-pi05-temporal-history-ablation}

  \resizebox{0.75\linewidth}{!}{%
  \begin{tabular}{
    cc
    >{\columncolor{gray!15}}c
    >{\columncolor{gray!15}}c
    cc
    cc
  }
    \toprule

    \multicolumn{2}{c}{$L=0$ (unused)}
    & \multicolumn{2}{>{\columncolor{gray!15}}c}{$L=1$}
    & \multicolumn{2}{c}{$L=2$}
    & \multicolumn{2}{c}{$L=4$} \\

    \cmidrule(lr){1-2}
    \cmidrule(lr){3-4}
    \cmidrule(lr){5-6}
    \cmidrule(lr){7-8}

    bpp & Succ. (\%)
    & bpp & Succ. (\%)
    & bpp & Succ. (\%)
    & bpp & Succ. (\%) \\
    \midrule

    0.0146 & 13.1
    & 0.0146 & 14.4
    & 0.0146 & 13.9
    & 0.0146 & \textbf{14.8} \\

    0.0528 & 54.7
    & 0.0525 & \textbf{57.2}
    & 0.0521 & 55.6
    & 0.0529 & 54.9 \\

    0.0788 & 57.5
    & 0.0796 & \textbf{61.3}
    & 0.0792 & 59.0
    & 0.0789 & 58.5 \\

    \bottomrule
  \end{tabular}%
  }
\end{table}
% \paragraph{Temporal history length.} In \Cref{tab:robocasa-pi05-temporal-history-ablation}, we conduct ablation studies with temporal history length to analyze the effect of temporal history length. Notably, longer history length doesn't necessarily guarantee higher performance. We attribute it to action chunking that generates large movement at one timestep, so that past history includes the scene completely dissimilar to the current frame.
\paragraph{Temporal history length.} In \Cref{tab:robocasa-pi05-temporal-history-ablation}, we analyze the effect of the temporal history length $L$. Leveraging temporal history consistently improves the success rate over the baseline without history ($L{=}0$) at nearly identical bitrates, indicating that the gain stems from temporal context rather than additional bits. $L{=}1$ yields the best overall performance, improving the success rate over L=0 by 1.3, 2.5, and 3.8 percentage points, respectively, with the gain becoming more pronounced at higher bitrates. Notably, a longer history does not necessarily yield higher performance. We attribute this to action chunking, which executes multiple actions per inference step and thus induces large scene changes between consecutive observations. As a result, frames further in the past can differ substantially from the current frame, providing less useful or even misleading context. Moreover, since a single past frame suffices, our method incurs minimal memory and computational overhead.

%\paragraph{Freezing components}
\section{Real-world Experiment} \label{ssec:real_world}
We conduct real-world experiments in a server-client split environment: Intel i7-10875H CPU on local laptop and the enterprise-level GPU workstation (RTX6000Ada) on a remote server. In \Cref{fig:camera_setup}, we set up the environment with multiple cameras from different perspectives and a single 6-DoF AgileX PiPER arm equipped with commercially available two-finger parallel-jaw gripper (80 mm stroke). The dataset is collected using a leader-follower teleoperation setup. We design three challenging long-horizon tasks for training and evaluation. We also apply asynchronous inference and fix the input resolution to $256 \times 256$. \Cref{ssec:real-world_evaluation} demonstrates our evaluation results, and \Cref{ssec:real_world_latency} shows the latency gain of SPARC in the real-world and component-wise breakdown. See \Cref{ssec:real-world experiment-setup} and \Cref{ssec:task_info} for more details, \Cref{sec:real_world_rollout} for example rollouts and \Cref{sec:failure_case} for failure case analysis.

\subsection{Evaluation} \label{ssec:real-world_evaluation} 
\begin{figure}[t]
    \centering

    % ===== Left: Camera setup =====
    \begin{minipage}[t]{0.50\textwidth}
        \centering
        \begin{subfigure}[t]{0.325\linewidth}
            \centering
            \includegraphics[width=\linewidth]{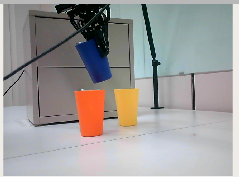}
            % \vspace{0.3em}
            \caption{side-view}
            \label{fig:side_view}
        \end{subfigure}
        \begin{subfigure}[t]{0.325\linewidth}
            \centering
            \includegraphics[width=\linewidth]{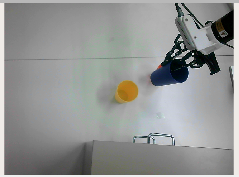}
        % \vspace{0.3em}
            \caption{top-view}
            \label{fig:top_view}
        \end{subfigure}
        \begin{subfigure}[t]{0.325\linewidth}
            \centering
            \includegraphics[width=\linewidth]{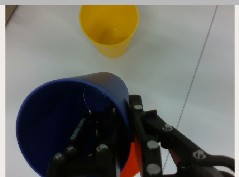}
            % \vspace{0.3em}
            \caption{wrist-view}
            \label{fig:wrist_view}
        \end{subfigure}
        % \vspace{0.3em}
        \caption{\textbf{Camera \& robot setup.}
        Three cameras with different viewpoints (256$\times$256 resolution) and the single AgileX PiPER arm.}
        \label{fig:camera_setup}
    \end{minipage}
    \hfill
    % ===== Right: Real-world results =====
    \begin{minipage}[t]{0.48\textwidth}
        \centering
        \begin{subfigure}[t]{0.38\linewidth}
            \centering
            \includegraphics[width=\linewidth]{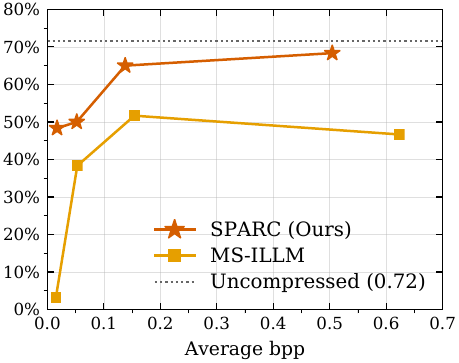}
            % \vspace{-0.3em}
            \caption{sub-tasks}
            \label{fig:rd_curve_subtask}
        \end{subfigure}
        \hspace{0.03\linewidth}
        \begin{subfigure}[t]{0.38\linewidth}
            \centering
            \includegraphics[width=\linewidth]{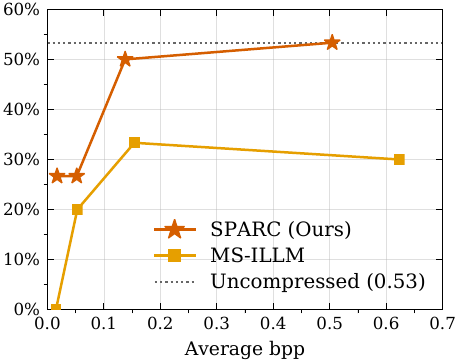}
            \caption{full-tasks}
            \label{fig:rd_curve_full}
        \end{subfigure}
        \caption{\textbf{Real-world results.}
        Average success scores for sub-tasks and full-tasks with 60 training episodes
        and 10 evaluation trials per task.}
        \label{fig:real_world_results}
    \end{minipage}

\end{figure}

In \Cref{fig:rd_curve_subtask} and \Cref{fig:rd_curve_full}, we report the success rate under our setting, showing that the practical gain using our compression method is notable compared to common high-quality JPEG q95 compression ($\sim$1.9 bpp).  We measure the performance of SPARC on three long-horizon tasks, all of which are assessed using full-task success score and sub-task success scores. Each long-horizon task consists of two sub-tasks. We measure both average sub-task success scores and average full-task success scores. Our result shows that SPARC can maintain a greater success rate than MS-ILLM across a wide range of bpps for both sub-tasks and full tasks.
% \paragraph{Other details.} At split-computing architectures, it is too costly to transmit the mask between the sender and the client. To avoid this extra overhead, we generate the mask synchronously at both server and client, extending the architecture of hyper-prior neural codec. Since we create the mask from quantized hyper-latent, the server-side mask and the client-side mask are theoretically guaranteed to be identical and consistent.
% \input{Tables/vram}
% \begin{figure*}[t]
%     \centering
%     % 첫 번째 이미지 (a) - 너비 31% 설정
%     \begin{subfigure}[t]{0.2\linewidth}
%         \centering
%         \includegraphics[width=\linewidth]{Figures/side_view.png}
%         \caption{side-view}
%         \label{fig:sub1}
%     \end{subfigure}
%     % 두 번째 이미지 (b) - 너비 31% 설정
%     \begin{subfigure}[t]{0.2\linewidth}
%         \centering
%         \includegraphics[width=\linewidth]{Figures/top_view.png}
%         \caption{top-view}
%         \label{fig:sub2}
%     \end{subfigure}
%     \begin{subfigure}[t]{0.2\linewidth}
%         \centering
%         \includegraphics[width=\linewidth]{Figures/wrist_view.png}
%         \caption{wrist view}
%         \label{fig:sub3}
%     \end{subfigure}
    
%     \caption{\textbf{Camera setup.} Three cameras with different viewpoints, each with 224$\times$224 resolution.}
%     \label{fig:three_plots}
% \end{figure*}

\subsection{Latency} \label{ssec:real_world_latency}
\begin{table}[t!]
  \centering
  \caption{{\textbf{Per-component latency.} We compare the measured latency (ms) of SPARC with MS-ILLM on $256 \times 256$ input resolution. Each measurement is a median over 50 trials.}}
  \label{tab:codec-component-latency-split}

  \setlength{\tabcolsep}{6pt}
  \scriptsize

  \begin{tabular}{lcccccccc}
    \toprule
    \textbf{Method}
    & \textbf{Keep}
    & \textbf{bpp}
    & \textbf{Main Net}
    & \textbf{Hyper Net}
    & \textbf{Hyper-Entropy}
    & \textbf{Temporal}
    & \textbf{Entropy}
    & \textbf{Total} \\
    \midrule
    \multicolumn{7}{l}{\textit{(a) Encoding}} \\
    SPARC 
    & $0.667$
    & $0.0454$
    & $68.58$
    & $16.68$
    & $3.42$
    & $5.35$
    & $11.28$
    & $106.24$ \\
    MS-ILLM
    & ---
    & $0.0497$
    & $69.09$
    & $16.68$
    & $3.34$
    & ---
    & $12.60$
    & $102.16$ \\
    \midrule
    \multicolumn{7}{l}{\textit{(b) Decoding}} \\
    SPARC 
    & $0.667$
    & $0.0454$
    & $5.55$
    & $0.43$
    & $1.51$
    & ---
    & $10.94$
    & $18.43$ \\
    MS-ILLM
    & ---
    & $0.0497$
    & $5.54$
    & $0.42$
    & $1.49$
    & ---
    & $14.01$
    & $21.46$ \\
    \bottomrule
  \end{tabular}
\end{table}
\begin{table}[!t]
\centering
\caption{\textbf{Per-camera end-to-end latency under varying bandwidth and number of streams.} We show how bandwidth-limited settings highlight the practical effectiveness of SPARC by reducing network latency. $\dagger$ denotes latency exceeding the required transmission deadline.
} \label{tab:latency}
\resizebox{0.7\columnwidth}{!}{%
\begin{tabular}{rcccccc}
\toprule
& \multicolumn{3}{c}{\textbf{JPEG (q95)}}
& \multicolumn{3}{c}{\textbf{SPARC}} \\
\cmidrule(lr){2-4}\cmidrule(lr){5-7}
& Enc/Dec & Network & Total
& Enc/Dec & Network & Total \\
\midrule
\multicolumn{7}{l}{\textit{(a) Varying bandwidth $C$ ($N{=}1$)}} \\
\midrule
250 kbps & 0.5 & --- & ---
         & 124.7 & 33.1 & \textbf{157.8} \\
500 kbps & 0.5 & 446.1 & 446.6$^{\dagger}$
         & 124.7 & 15.8 & \textbf{140.4} \\
750 kbps & 0.5 & 205.6 & 206.1
         & 124.7 & 10.3 & \textbf{135.0} \\
1 Mbps   & 0.5 & 133.6 & 134.1
         & 124.7 & 7.7 & \textbf{132.4} \\
2 Mbps   & 0.5 & 55.6 & \textbf{56.1}
         & 124.7 & 3.8 & 128.5 \\
\midrule
\multicolumn{7}{l}{\textit{(b) Varying streams $N$ ($C{=}100$~Mbps)}} \\
\midrule
$N{=}1$    & 0.5 & 0.96  & \textbf{1.5}
            & 124.7 & 0.08 & 124.7 \\
$N{=}128$  & 0.5 & 1.51  & \textbf{2.0}
            & 124.7 & 0.08 & 124.7 \\
$N{=}256$  & 0.5 & 3.57  & \textbf{4.1}
            & 124.7 & 0.08 & 124.7 \\
$N{=}347$  & 0.5 & 145.0 & 145.5
            & 124.7 & 0.08 & \textbf{124.8} \\
% $N{=}349$  & 0.5 & 1122.0 & 1122.5$^{\dagger}$
%             & 124.7 & 0.08 & \textbf{124.8} \\
$N{=}1024$ & 0.5 & --- & ---
            & 124.7 & 0.10 & \textbf{124.8} \\
$N{=}4096$ & 0.5 & --- & ---
            & 124.7 & 1.01 & \textbf{125.7} \\
\bottomrule
\end{tabular}%
}
\end{table}

\paragraph{Component-wise latency breakdown.} In \Cref{tab:codec-component-latency-split}, we report the component-wise latency comparison between MS-ILLM and SPARC. We omit the decoder-side mask re-materialization time since it can be performed simultaneously with encoder-side mask materialization and transmission, introducing no additional latency. While the temporal mask selector incurs some additional computational overhead, this is offset by the reduction in entropy coding, resulting in similar overall latency. 

\paragraph{End-to-end latency analysis.} In \Cref{tab:latency}, we report the end-to-end latency of our split-computing architecture, where the network latency is computed from the transmitted bitstream size and the available link bandwidth. Since the policy executes action chunks of 10 steps at 30~Hz, each observation must be delivered within 333~ms. As a baseline, we adopt JPEG with a quality factor of 95, the default setting in ROS~2, which records 1.9~bpp on our real-world images. Since SPARC maintains competitive performance down to 0.15~bpp in our real-world experiments, we set its bitrate to 0.15~bpp. The results reveal a clear trade-off between communication and computation. Although SPARC incurs additional neural encoding overhead, its bitrate reduction substantially lowers the transmission latency under bandwidth-constrained conditions, yielding lower latency than JPEG below approximately 1~Mbps or beyond 347 concurrent streams on a 100~Mbps link. In particular, JPEG fails to meet the deadline at 500~kbps or below, whereas SPARC remains within the deadline even at 250~kbps. Moreover, the latency of SPARC stays nearly constant as the number of streams increases, while that of JPEG diverges beyond 347 streams due to link saturation. We note that deploying SPARC on more capable edge-computing platforms, such as NVIDIA Jetson devices with GPU acceleration, could further reduce the encoding latency and extend its advantage to higher-bandwidth settings.
% \footnote{We measure VRAM via \texttt{torch.cuda.reset\_peak\_memory\_stats} per frame.} Numerically, the total latency is reduced by about 27\%. In addition, we compare virtual memory consumed by SPARC and MS-ILLM. The comparison shows that although the temporal mask selector model leads to greater virtual memory, its effect is not significant.
% \newpage
\section{Conclusion} \label{sec:limitations_and_conclusion}
In this work, we propose SPARC, a learned image compression framework that adaptively allocates bitrate across camera views and spatial regions using task-aware spatiotemporal latent masks. By prioritizing control-relevant visual information, SPARC achieves a stronger bitrate–success tradeoff than conventional image/video codecs and learned compression baselines across simulation and real-world experiments. Our analysis further shows that the additional latency introduced by the temporal mask selector on CPU is offset by the reduction in the size of transmitted latent representation resulting in lower entropy coding and decoding overhead. We further believe that additional reduction of latency is possible with more powerful on-device compute resources, making SPARC a practical step toward bandwidth-efficient VLA deployment.
%\section{Limitations} \label{ssec:arch}
% One of limitations of this paper is the absence of language grounding, which requires larger models that may have induced higher latency. Also, erroneously masking critical details can potentially lead to safety issues, requiring some precautions, such as safety lock for unexpected robot movement. The lack of analysis on remote constrained environment, such as deep ocean, is restricted by spatiotemporal constraints, but is indeed the potential source of future work.
%Due to spatiotemporal constraints, our setting does not reflect extremely busy channel constraints in the real-world experiments. Future work should evaluate SPARC under more realistic and challenging network constraints, such as unstable bandwidth, packet loss, and severe communication delays.%

%===============================================================================

\clearpage
% The acknowledgments are automatically included only in the final and preprint versions of the paper.
% \acknowledgments{If a paper is accepted, the final camera-ready version will (and probably should) include acknowledgments. All acknowledgments go at the end of the paper, including thanks to reviewers who gave useful comments, to colleagues who contributed to the ideas, and to funding agencies and corporate sponsors that provided financial support.}
\subsection*{AI use statement}
In this work, we used generative AI tools for helping develop theoretical models or conceptual frameworks, designing or providing feedback on research methodology or experiments, implementing methods, and interpreting results.
We have not used generative AI tools for proposing or refining hypotheses and cleaning and reformatting dataset
and the rest of the required disclosure tasks are not applicable to this work. We have reviewed all AI-assisted work. For example, we checked LLM-generated research ideas for potential plagiarism through a manual literature survey and LLM-generated code was verified and tested for correctness. We take responsibility for the final content of this work,
including text, claims or artifacts produced with the aid of generative AI.

% See the ICLR 2027 AI Policy for Authors for more details. This statement should
% not be more than 1 page.
\subsection*{Reproducibility statement}
For the implementation of SPARC, we submit a link to an
anonymous downloadable source code as supplementary materials. 
%===============================================================================
\clearpage
% \bibliographystyle{IEEEtran}
\bibliography{iclr2027_conference}
\bibliographystyle{iclr2027_conference}
\clearpage
\section*{Appendix}
\appendix
\raggedbottom
\startcontents[appendix]
\printcontents[appendix]{}{1}{\setcounter{tocdepth}{2}}
\newpage
\section{Limitations}
The absence of language grounding constrains interpretive flexibility; incorporating it would require larger models that may introduce non-trivial latency costs. Second, erroneous masking of task-critical details could pose a potential safety concern and may require precautionary measures such as hardware-level safety locks against unexpected robot motion. Finally, while remote and resource-constrained environments---such as deep-ocean settings---represent a compelling frontier for this framework, their evaluation remains out of reach due to spatiotemporal constraints.

\section{Additional Experimental Details}\label{sec:method:impl}
\subsection{Implementation Details}
\label{ssec:add_exp_details}
In \Cref{tab:implementation_details}, we provide additional training, evaluation, and model configurations of temporal mask selector model. Also, our implementation is anonymously available at \url{https://anonymous.4open.science/r/SPARC-B226}
\begin{table}[!ht]
    \centering
    \small
    \setlength{\tabcolsep}{5pt}
    \renewcommand{\arraystretch}{1.15}
    \caption{Additional implementation details}\label{tab:implementation_details}
    \begin{tabular}{ll}
    \toprule
    \textbf{Category} & \textbf{Configuration} \\
    \midrule
    \multicolumn{2}{l}{\textbf{Training}} \\
    Optimizer & AdamW ($\beta_{1}$ = 0.9, $\beta_{2}$ = 0.999) \\
    Weight decay & 0.01\\
    Gradient clipping & 1.0 \\
    Batch size & 12 \\
    Training steps & 15000 \\
    Scheduler & cosine annealing with \texttt{min\_lr = 1e-06} \\
    Image resolution (LIBERO, RoboCasa, Real-World) &  $256 \times 256$  \\  
    Image resolution (VLABench) & $480 \times 480$ \\
    Codec training scope & A single codec for all camera views \\
    Precision & fp32 \\
    Num. Of Evaluation Episodes (LIBERO, RoboCasa, VLABench) & 50 \\
    \midrule
    \multicolumn{2}{l}{\textbf{Evaluation}} \\
    Action chunk size & Official default settings for each benchmark \\
    Action chunk size (Real World) & 10 \\
    \midrule
    \multicolumn{2}{l}{\textbf{SPARC}} \\
    Latent channels & 320 (= MS-ILLM) \\
    Num. of Temporal Mask Selector Parameters & 1.4M \\
    Num. of Encoder Parameters & 7.42M\\
    Num. of Hyperprior-Entropy Parameters & 17.3M \\
    Num. of Decoder Parameters & 156M \\
    Layers & 1 \\
    Attention heads & 8 \\
    Hidden dimension & 1280 ($4\times$ the input dimension) \\
    \bottomrule
    \end{tabular}
\end{table}
\newpage
\subsection{Real-world Experiment Setup}\label{ssec:real-world experiment-setup}
\begin{figure}[h]
    \centering
    \includegraphics[width=0.3\linewidth]{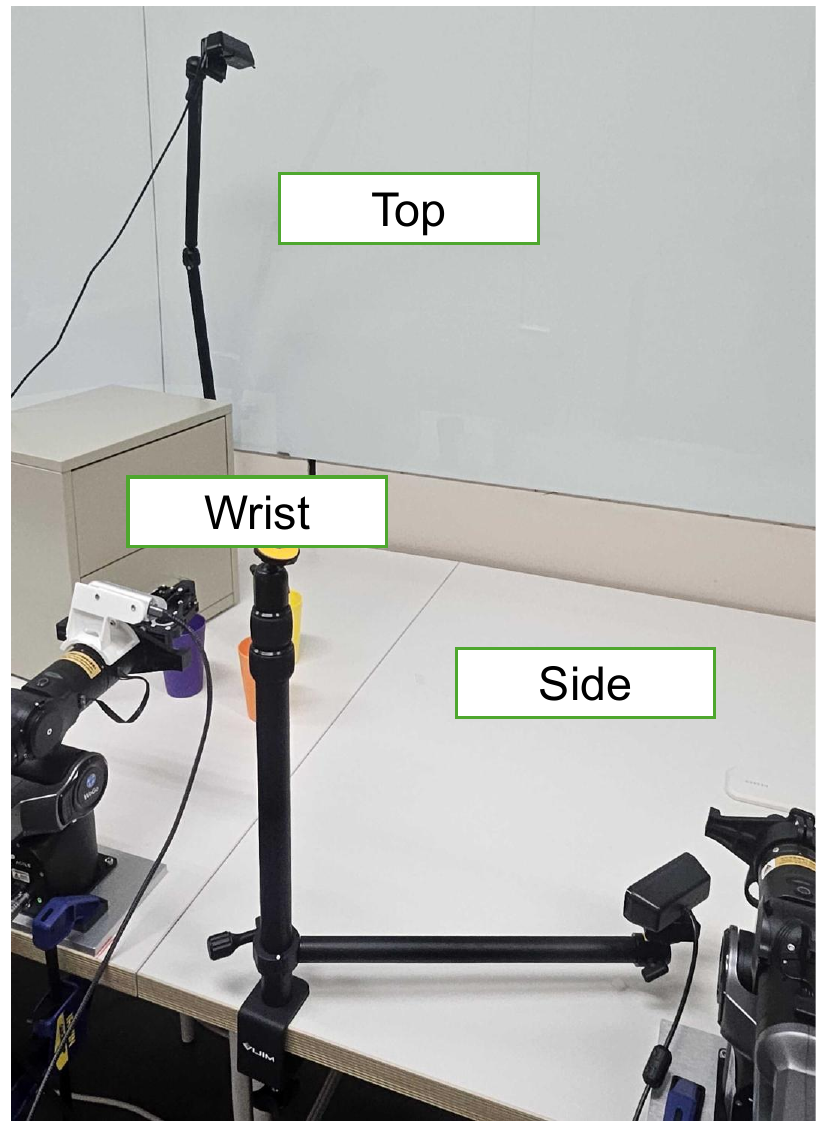}
    \caption{Real-world environment setups.}
    \label{fig:real_world_setup}
\end{figure}
In \Cref{fig:real_world_setup}, we provide an overview of our real world experimental setup. We utilize three cameras (top, side, and wrist) to capture multi-view visual observations. Our platform consists of a single 6-DoF AgileX PiPER arm and three
RGB cameras providing top, side, and wrist views (highlighted in red). Observations are encoded
on-board and transmitted over the emulated link via SSH Tunnel to a remote policy server.

\subsection{Task Information} \label{ssec:task_info}
In \Cref{tab:robocasa_task_list} and \Cref{tab:realworld_task_list}, we provide task composition information for simulation benchmarks and real-world evaluations. For LIBERO and VLABench, we evaluated on all available tasks. For RoboCasa365, we evaluated on 24 selected atomic tasks. For real-robot tasks, we give the details of tasks used for fine-tuning and evaluation.

\begin{table}[!ht]
\centering
\caption{\textbf{RoboCasa atomic tasks.} The 24 atomic tasks used for every simulation result, grouped by skill family. Each task is evaluated with the official pretrained-split scene layouts.}
\label{tab:robocasa_task_list}
\small
\begin{tabularx}{\linewidth}{@{}l>{\raggedright\arraybackslash}Xr@{}}
\toprule
Skill family & Tasks & \# \\
\midrule
Pick-and-place
  & \texttt{PickPlaceCounterToCabinet}, \texttt{PickPlaceCabinetToCounter},
    \texttt{PickPlaceCounterToSink}, \texttt{PickPlaceSinkToCounter},
    \texttt{PickPlaceCounterToMicrowave}, \texttt{PickPlaceMicrowaveToCounter},
    \texttt{PickPlaceCounterToStove}, \texttt{PickPlaceStoveToCounter} & 8 \\
\addlinespace[2pt]
Open / close doors
  & \texttt{OpenCabinet}, \texttt{CloseCabinet}, \texttt{OpenMicrowave}, \texttt{CloseMicrowave} & 4 \\
Open / close drawers
  & \texttt{OpenDrawer}, \texttt{CloseDrawer} & 2 \\
Turning levers
  & \texttt{TurnOnSinkFaucet}, \texttt{TurnOffSinkFaucet} & 2 \\
Twisting knobs
  & \texttt{TurnOnStove}, \texttt{TurnOffStove} & 2 \\
Pressing buttons
  & \texttt{TurnOnMicrowave}, \texttt{TurnOffMicrowave}, \texttt{StartCoffeeMachine} & 3 \\
Insertion
  & \texttt{CoffeeSetupMug}, \texttt{CoffeeServeMug} & 2 \\
Navigation
  & \texttt{NavigateKitchen} & 1 \\
\midrule
Total & & 24 \\
\bottomrule
\end{tabularx}
\end{table}

\begin{table}[!ht]
\centering
\caption{\textbf{Real-robot tasks.} Language instruction given to the policy,
the sub-tasks it chains, and the demonstrations used for fine-tuning
(single-arm PiPER, three cameras: top, side, wrist; 30\,Hz).}
\label{tab:realworld_task_list}
\small
\setlength{\tabcolsep}{5pt}
\begin{tabularx}{\linewidth}{@{}>{\raggedright\arraybackslash}p{2.0cm}YYr@{}}
\toprule
Task & Instruction & Sub-tasks & Demos \\
\midrule
Stack cups
   & \textit{``stack the purple cup on the orange one, and put the yellow one on top''}
   & (1) pick the purple cup $\rightarrow$ place it on the orange cup \newline
     (2) pick the yellow cup $\rightarrow$ place it on the purple cup
   & 60 \\
\addlinespace[3pt]
Cabbage \& carrot to pot
   & \textit{``take the cabbage from the dish and put it in the pot, then add the carrot''}
   & (1) pick the cabbage from the dish $\rightarrow$ place it in the pot \newline
     (2) pick the carrot $\rightarrow$ place it in the pot
   & 60 \\
\addlinespace[3pt]
Carrot to drawer
   & \textit{``pick up the carrot and put it in the upper drawer''}
   & (1) move to the upper drawer $\rightarrow$ open the upper drawer \newline
     (2) pick the carrot $\rightarrow$ place it in the upper drawer
   & 60 \\
\midrule
Total & & & 180 \\
\bottomrule
\end{tabularx}
\end{table}

\newpage
\subsection{Baseline Reproduction}
In this section, we describe how we reproduced the existing codec baselines in our setting. All baselines are evaluated with exactly the pipeline used for SPARC: the policy is frozen and untouched, and the codec is inserted between the simulator and the policy, so every camera observation is encoded to a bitstream and decoded back before it enters the policy's own preprocessing. 
% \paragraph{Common protocol.}
% All camera streams the policy consumes are coded: on RoboCasa365 the left
% agent-view, eye-in-hand and right agent-view cameras for $\pi_{0.5}$ and
% GR00T N1.6, and the left agent-view and eye-in-hand cameras for $\pi_0$; on
% VLABench the main, second and wrist cameras; on LIBERO the agent-view and
% wrist cameras. Frames are coded at the resolution the environment renders
% for the policy, $256{\times}256$ on RoboCasa365 and LIBERO,
% and $480{\times}480$ on VLABench, before the policy resizes them to its own
% input size. The policies act with action chunking, so a codec is invoked
% only when the policy is queried. Consequently one coded frame
% corresponds to one policy call, and an inter-frame codec has to predict
% across that gap rather than across consecutive rendered frames. The rate
% of an episode is the size of the emitted bitstreams in bits divided by the
% number of observation pixels, averaged over the coded streams and the
% policy calls of the episode.
\paragraph{JPEG 2000.}
We use OpenJPEG through Pillow with the irreversible 9/7 wavelet and a single quality layer at a fixed compression ratio $\mathrm{CR}\in\{50,100,200,300,500,1000,2000,5000\}$, i.e.\ a target of
$24/\mathrm{CR}$ bpp for 24-bit RGB; the rate is the size of the emitted codestream.
\paragraph{VVC.}
We use the Fraunhofer VVenC~1.13.0 encoder and VVdeC~3.0.0 decoder, both built from source in release configuration. RGB frames are converted to 8-bit YUV~4:2:0 with the \texttt{faster} preset and a fixed
QP$\in\{22,27,32,37,42,47,51,55\}$, decoded, and converted back to RGB with nearest-neighbor chroma upsampling.
\paragraph{DCVC-RT.}
We use the authors' release: the image model codes I-frames and the video model codes P-frames. Each camera maintains its own independent decoded picture buffer, which is cleared at every episode boundary, so the first
policy call of every episode is coded as an I-frame and every subsequent call as a P-frame predicted from the previously coded frame of the same camera. Frames are replicate-padded to a multiple of $16$ pixels, inference
runs in half precision as in the authors' real-time setting. We sweep the model's continuous rate index $q\in\{0,4,8,16,24,32,40,48,56,63\}$ (larger is higher rate).
\paragraph{HPCM.}
We use the authors' LIC-HPCM release with the HPCM-Large architecture and its MS-SSIM-optimised checkpoints, giving six rate points
($\lambda\in\{2.4,4.58,8.73,16.64,31.73,60.5\}$).The entropy model uses the authors' scale table ($60$ levels between $0.12$ and $64$). HPCM is run in full precision.
\paragraph{MS-ILLM.}
We use the authors' release through its \texttt{torch.hub} entry points.
The release names its operating points by target rate; we evaluate $q=vlo1$, $q=vlo2$, $q{=}1$, $q{=}2$, and $q{=}3$. Frames are reflect-padded to a multiple of $64$ pixels and the reconstruction is cropped back. The rate is the size of the actual rANS bitstream (latent and hyper-latent strings).
% \paragraph{Hardware.}
% VVenC and VVdeC are CPU implementations; all learned codecs run on a GPU.
% Policy rollouts were distributed over a heterogeneous GPU cluster, which
% affects only wall-clock time: every codec above is deterministic given its
% configuration, and the environment seed is fixed, so none of the reported
% rates or success rates depend on the host.

\section{Additional Ablation Results} \label{sec:ablations}
\subsection{Tilted Rate Loss}\label{ssec:alpah}
\begin{table}[H]
\centering
\caption{Performance across different bitrates with varying $\alpha$.}
\label{tab:ablation_alpha}
\begin{tabular}{c cc cc cc}
\toprule
$\alpha$ & Succ. (\%) & bpp & Succ. (\%) & bpp & Succ. (\%) & bpp \\
\midrule
0.0 & 54.9 & 0.0546 & 58.9 & 0.0891 & 15.2 & 0.0146 \\
0.2 & 53.7 & 0.0543 & 59.1 & 0.0863 & \textbf{15.9} & 0.0146 \\
\rowcolor{gray!15}
0.4 & \textbf{57.2} & \textbf{0.0525} & \textbf{61.3} & \textbf{0.0788} & 14.4 & 0.0146 \\
0.6 & 55.6 & 0.0535 & 59.2 & 0.0847 & 0.2 & 0.0081 \\
0.8 & 56.3 & 0.0530 & 59.8 & 0.0819 & 0.3 & \textbf{0.0065} \\
1.0 & 55.8 & 0.0545 & 59.3 & 0.0828 & 0.2 & \textbf{0.0065} \\
\bottomrule
\end{tabular}
\end{table}
We conduct an ablation study of tilted rate loss, as described in \Cref{ssec:training}. In tilted rate loss, the hyper-parameter $\alpha$ determines the degree to which the temporal mask selector aggressively masks regions with high local bitrates. A lower value of $\alpha$ encourages the temporal mask selector to mask based more on the action distortion loss than on local bitrates. In other words, it masks more task-irrelevant regions regardless of how many bits are allocated to them, which leads to an increase in the number of latent channels masked by the temporal mask selector. To verify the advantages of tilted rate loss, we report the average success rate and latency of SPARC with different values of $\alpha$ on the RoboCasa365 benchmark with $\pi_{0.5}$ as the downstream VLA model across a range of different bpp values. For latency measurements, note that all reported points correspond to models initialized from a pretrained model targeting a fixed quality level and trained with fixed hyperparameters to directly analyze the effect of $\alpha$.
\paragraph{Performance.} \Cref{tab:ablation_alpha} shows the effect of scaling $\alpha$ on the bit-success rate trade-off curve. Selecting a moderate value of $\alpha$ ($0.4\leq \alpha \leq0.8$) improves the performance of VLA models by preventing the temporal mask selector from aggressively masking high-frequency details to which a high proportion of bits is allocated, while preserving important details that require high local bitrates. This elucidates the advantage of tilted rate loss in maintaining high performance within a similar bpp range.
% \paragraph{Latency.} \Cref{tab:ablation_alpha} shows that decreasing $a$ reduces latency while maintaining similar success rates. This clearly shows that setting $a$ to 0.4 achieves the largest reduction while maintaining success rates. This indicates that adjusting $a$ can improve latency.
\paragraph{Visualization.}
% \begin{figure}[h]
%     \centering
%     \includegraphics[width=1.0\linewidth]{Figures/vlabench_alpha_bit_allocation_panel.pdf}
%     \caption{Bit allocation maps across a range of $\alpha$.}
%     \label{fig:unseen_tasks_vlabench}
% \end{figure}

%\setlength{\intextsep}{0pt}
\begin{figure}[h]
    \centering
    \includegraphics[
        width=0.80\linewidth
    ]{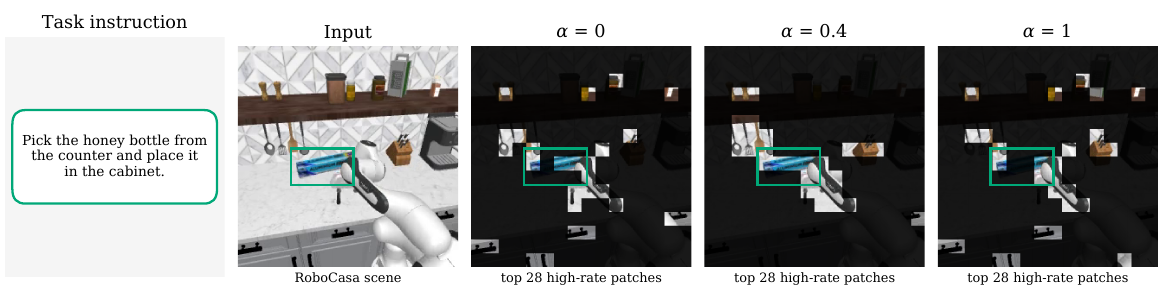}
    \caption{Example visualization of sweeping $\alpha$ on rate preservation}
    \label{fig:vlabench_alpha_bit_allocation}
\end{figure} 
\Cref{fig:vlabench_alpha_bit_allocation} illustrates how the rate-preservation coefficient $\alpha$ reshapes the spatial allocation of the token budget. For a representative RoboCasa manipulation episode with the instruction "Pick the honey bottle from the counter and place it in the cabinet", we visualize the top-28 patches ranked by estimated rate under three settings of $\alpha$, where the green box denotes the task-relevant region containing the target object. At $\alpha$ = 0 and $\alpha$ = 1, a substantial fraction is absorbed by background clutter such as the shelf and surrounding utensils, leaving the target region only sparsely covered. As $\alpha$ increases to 0.4, the retained patches progressively concentrate around the target object and the manipulator, indicating that rate preservation reallocates capacity toward regions that are causally relevant to the instruction rather than merely photometrically complex. This qualitative trend is consistent with the quantitative gains reported in \Cref{tab:ablation_alpha}, and suggests that the improvements stem from better-targeted token allocation rather than from an overall increase in retained information.

\subsection{Comparison with Alternative Training Recipes}\label{ssec:comparison}
% \begin{figure}[t]
%     \centering
%     \begin{minipage}{0.48\textwidth}
%         \centering
%         \includegraphics[width=\linewidth]{Figures/robocasa_pi05_msillm_variants.pdf}
%     \end{minipage}\hfill % 두 minipage 사이의 공간을 최대로 띄움
% \end{figure}
\begin{figure}[h]
    \centering
    \includegraphics[width=0.5\linewidth]{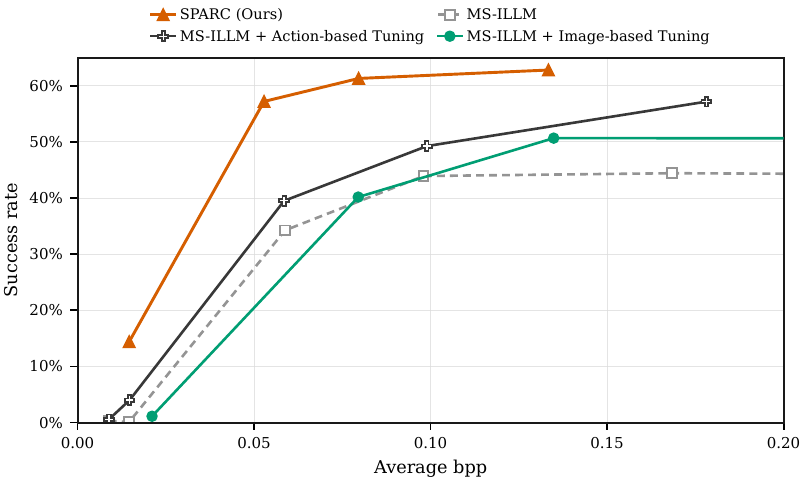}
    \caption{MS-ILLM variants on RoboCasa365 benchmarks.}
    \label{fig:robocasa_msillm_variants}
\end{figure}
We compare with other simple training methods to verify that the training method of SPARC is robot-oriented and effective. To do this, we compare SPARC with the following plausible training recipes.
\begin{itemize}[leftmargin=*,topsep=0pt,parsep=0pt]
    \item \textbf{Action-based Tuning}: Decoder-only fine-tuning with VLA optimization objectives
    \item \textbf{Image-based Tuning}: Original training objectives of MS-ILLM
\end{itemize}
As with SPARC, we apply these training recipes to pretrained MS-ILLM models with different quality levels. We report the average success rate of SPARC on 24 atomic tasks in the RoboCasa365 benchmark with $\pi_{0.5}$ as the downstream VLA model across a range of different bpp values.

\Cref{fig:robocasa_msillm_variants} compares average success rates across a wide range of bpp values among SPARC, MS-ILLM, and MS-ILLM variants trained with the two different recipes. It shows that our method consistently outperforms the baseline across a wide range of bpp values. In particular, SPARC achieves the highest performance at lower bpp values than the other training methods before entering the very low bpp regime, where success rates drop substantially. Moreover, it consistently outperforms other training methods at low bpp values. Other training methods, regardless of whether they are perceptually oriented or action-distortion-oriented, cannot match the performance gain achieved by SPARC, which benefits from the spatial allocation of local bitrates for downstream VLA models.

\subsection{Network robustness} \label{ssec:network_simulation}
\begin{table}[h]
\centering
\caption{\textbf{Network robustness.} On RoboCasa365 benchmark with $\pi_{0.5}$ backbone, we evaluate SPARC under a range of
network-distorted conditions emulated with a FIFO link emulator, covering
bandwidth throttling, fixed RTT, jitter, and frame drop. All results are
averaged over 3 seeds.}
\label{tab:robocasa_network_robustness}

\resizebox{\linewidth}{!}{%
\begin{tabular}{llrrrrrrrrrrr}
\toprule
& & Clean
& \multicolumn{4}{c}{Bandwidth (kbps)}
& \multicolumn{2}{c}{Fixed RTT (ms)}
& \multicolumn{2}{c}{Jitter (ms)}
& \multicolumn{2}{c}{Drop (\%)} \\
\cmidrule(lr){4-7} \cmidrule(lr){8-9} \cmidrule(lr){10-11} \cmidrule(l){12-13}
Method & Metric &
& 50 & 100 & 300 & 500
& 250 & 500
& $250{\pm}50$ & $500{\pm}100$
& 10 & 30 \\
\midrule
\textbf{JPEG} & Succ. (\%)
& 38.67
& 1.67
& 2.00
& 2.17
& 6.08
& 30.17
& 18.67
& 25.25
& 13.25
& 40.58
& 37.58 \\
& bpp
& 1.081
& 1.054
& 1.054
& 1.046
& 1.076
& 1.085
& 1.084
& 1.083
& 1.074
& 1.080
& 1.066 \\
\midrule
\rowcolor{gray!15}
\textbf{SPARC} (Ours) & Succ. (\%)
& \textbf{57.2}
& \textbf{49.9}
& \textbf{50.7}
& \textbf{49.8}
& \textbf{49.0}
& \textbf{49.0}
& \textbf{39.9}
& \textbf{44.4}
& \textbf{32.8}
& \textbf{55.4}
& \textbf{52.3} \\
\rowcolor{gray!15}
& bpp
& 0.053
& 0.053
& 0.053
& 0.053
& 0.053
& 0.053
& 0.052
& 0.052
& 0.052
& 0.053
& 0.053 \\
\bottomrule
\end{tabular}%
}
\end{table}

In \Cref{tab:robocasa_network_robustness}, we demonstrate the robustness of SPARC against JPEG, which is a common compression method on robot foundation models, In a variety of limited channel conditions, unlike JPEG, SPARC maintains its performance under matched or slightly higher success rate.

\subsection{Freezing components}\label{ssec:freeze_components}
% \begin{figure}[h]
%     \centering
%     \includegraphics[width=1.0\linewidth]{Figures/robocasa_pi05_freeze_component_ablation.pdf}
%     \caption{Freezing components}
%     \label{fig:ablation_per_task}
% \end{figure}
\begin{figure}[H]
    \centering
    \includegraphics[width=0.6\linewidth]{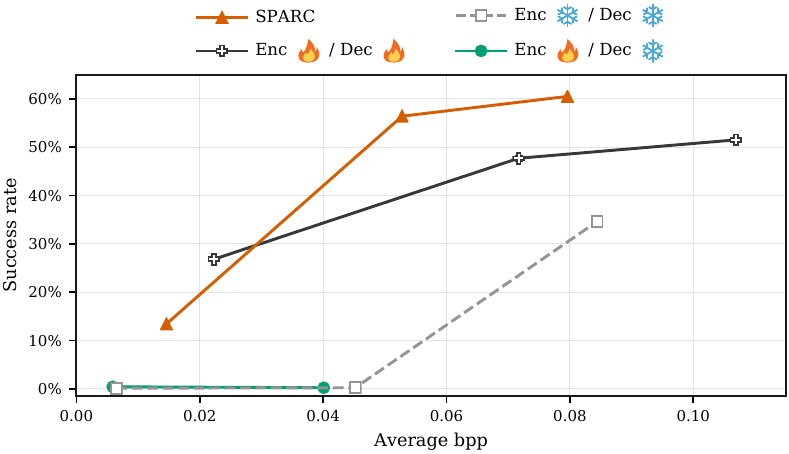}
    \caption{Freezing encoder or decoder on SPARC}
    \label{fig:freeze_architecture_abl}
\end{figure}
We test the effects of freezing key components on the performance of SPARC to verify why we choose to freeze encoder, and train decoder. \Cref{fig:freeze_architecture_abl} compares average success rates across a range of bpp values between SPARC and variants in which the encoder and/or decoder are frozen or fine-tuned. SPARC reaches a success rate of 56$\%$ at around 0.035 bpp and 60$\%$ at 0.08 bpp. It outperforms the fully fine-tuned variant across most of the bpp range, and the fully fine-tuned variant attains lower success rates (47$\%$ and 51$\%$) even at higher bitrates (0.072 and 0.107 bpp). SPARC falls behind the fully fine-tuned variant only in the very-low-bpp regime (below roughly 0.03 bpp), where its success rate drops substantially. Freezing the decoder has a severe impact. The variant with a trainable encoder and a frozen decoder fails completely, with 0$\%$ success at all evaluated bitrates. When both the encoder and decoder are frozen, the success rate stays at 0$\%$ up to about 0.045 bpp and reaches only 34$\%$ at 0.085 bpp. These results indicate that adapting the decoder is essential for preserving action-relevant information for downstream VLA models. They also show that SPARC's training design achieves a better rate–performance trade-off than naively fine-tuning all components.

\section{Transferability Analysis}\label{sec:trans}
% \begin{wrapfigure}[10]{r}{0.4\linewidth} % r: 오른쪽 정렬 (왼쪽 정렬은 l), 0.4\linewidth: 텍스트 단 폭의 40% 공간 할당
%     \vspace{-10pt}
%     \centering
%     % wrapfigure가 할당한 공간 안에서 이미지가 꽉 차도록 1.0\linewidth로 설정합니다.
%     \includegraphics[width=1.0\linewidth]{Figures/robocasa_pi0_xvla_transfer_bpp_success.pdf}
%     \label{fig:transferability_pi0_pi05}
% \end{wrapfigure}
\begin{figure}[!h]
    \centering
    \includegraphics[width=0.625\linewidth]{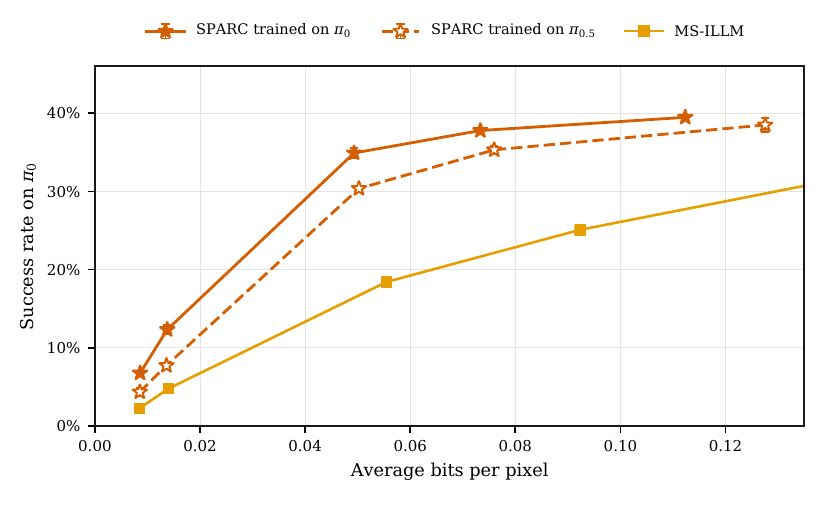}
    \caption{\textbf{Transfer to another VLA backbone.} We train SPARC on $\pi_{0.5}$ and run inference with different VLA backbone, $\pi_{0}$.}
    \label{fig:transferability}
\end{figure}
%Table로 바꾸기
\Cref{fig:transferability} evaluates whether SPARC's learned compression strategy generalizes to a VLA backbone unseen during training. We train SPARC on $\pi_{0.5}$ and evaluate it on $\pi_0$, comparing against the oracle setting (SPARC trained directly on $\pi_0$) and the MS-ILLM baseline. Despite never observing $\pi_0$ during training, SPARC trained on $\pi_{0.5}$ consistently outperforms MS-ILLM across the full bitrate range, and the gap to the oracle narrows as bitrate increases. We attribute this to the fact that the visual regions critical to task success---e.g., the gripper and target object---are largely determined by the manipulation task itself rather than by the specific VLA backbone, so a codec trained to preserve task-relevant information on one backbone transfers reasonably well to another. The residual gap to the oracle likely reflects backbone-specific differences in attention granularity rather than a failure of transfer.
\section{LIBERO-Plus Benchmark}\label{sec:libero-plus}
\begin{figure}[h]
    \centering
    \includegraphics[width=1.0\linewidth]{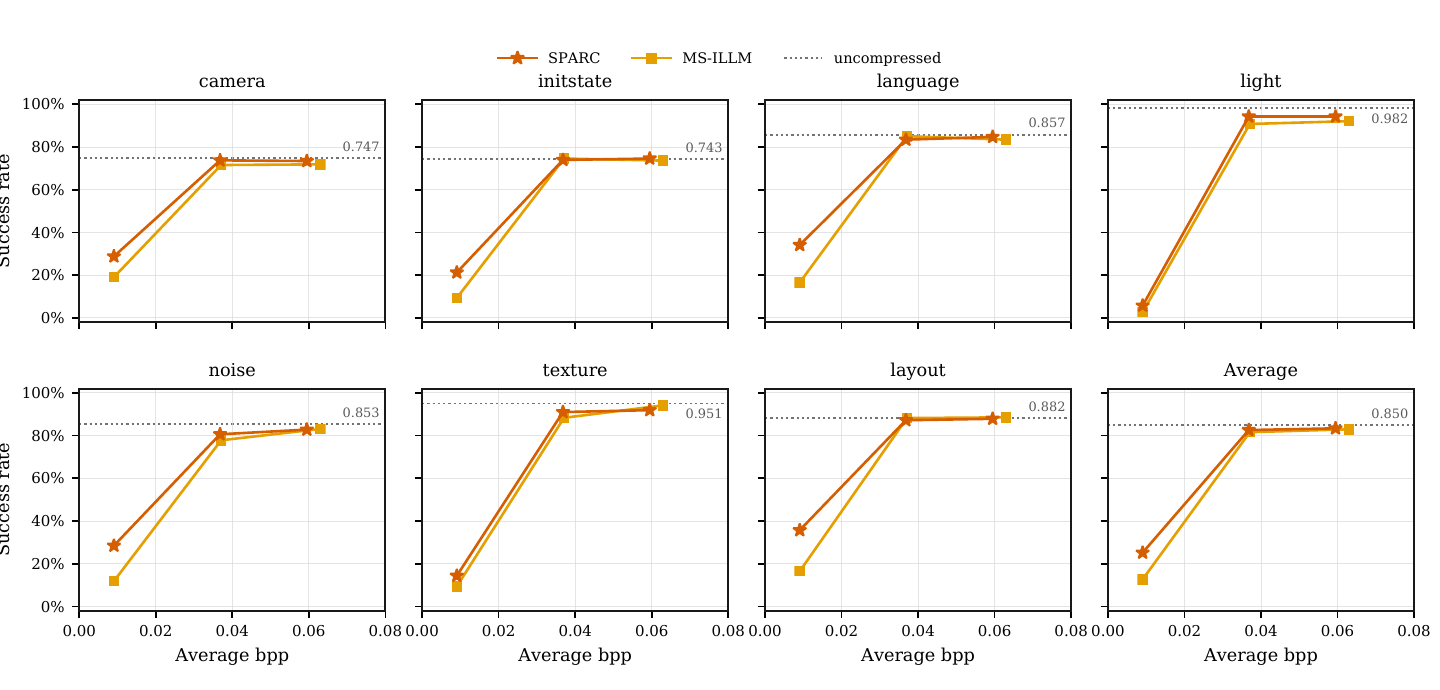}
    \caption{\textbf{LIBERO-Plus results.} \textit{Average} represents the average of other seven perturbation dimensions.}
    \label{fig:libero_plus}
\end{figure}
To assess robustness beyond the standard LIBERO suite, we evaluate on LIBERO-Plus \citep{liberoplus}, which perturbs each episode along seven axes---camera pose, initial state, language instruction, lighting, sensor noise, texture, and object layout---as well as an aggregate setting combining all perturbations (\texttt{Average}). \Cref{fig:libero_plus} shows that the gains observed on standard LIBERO transfer consistently to this more challenging setting: across nearly all perturbation types, SPARC matches or exceeds MS-ILLM at every bitrate and closes the gap to the uncompressed baseline more quickly. The advantage is most pronounced in the low-bitrate regime ($\sim$0.01 bpp), where SPARC roughly doubles the success rate of MS-ILLM under \texttt{initstate}, \texttt{noise}, \texttt{texture}, and the \texttt{Average} aggregate, indicating that SPARC's compression better preserves task-relevant information precisely when bit budgets are tightest. The two methods converge as bitrate increases, with both approaching the uncompressed success rate by 0.03--0.05 bpp in most categories. The \texttt{light} perturbation is a partial exception, where the uncompressed baseline is already near-ceiling (0.982) and both compressed variants converge to it almost immediately. Overall, these results indicate that SPARC's gains are not an artifact of the standard LIBERO setup but generalize across a broad range of visual and semantic perturbations.
% \section{Ablation Study}\label{sec:ablations}
% \subsection{Analysis of Key Components}\label{ssec:model_components}
% \input{Figures_tex/ablation_results} 
% We conduct ablation studies on key design choices. We compare SPARC with several variants excluding one of the following key design choices: (a) FiLM with robot states, (b) Past feature memory for temporal mask selector and (c) Phase 1 warming up stage.
% We set the VLABench benchmark and $\pi_{0.5}$ model as downstream VLA model.
% \paragraph{Performance.} \Cref{fig:ablation} and \Cref{tab:vlabench-pi05-sparc-ablation} show that SPARC clearly achieve the highest success rate at higher bpps where the task performance is meaningful, while other variants show comparable performance at lower bpps. The breakdown of ablation results on \Cref{tab:vlabench-pi05-sparc-ablation} also shows that SPARC achieves the highest bpp at similar bpp range. It achieves lower success rate at very low bpp. However, all variants mark very low success rate at this range ($\approx10.0\%$). At this compression rate, it is difficult to reserve minimum information for the performance of VLA models. Considering the practicality, we can focus on the performance of VLA models at higher bpps with meaningfully high performances. As previously stated, SPARC achieves the best performance at this range.
\newpage
\section{Task-criticality of the mask selector} \label{ssec:task-critical}
\begin{figure}[h]
\centering
\includegraphics[width=0.8\linewidth]{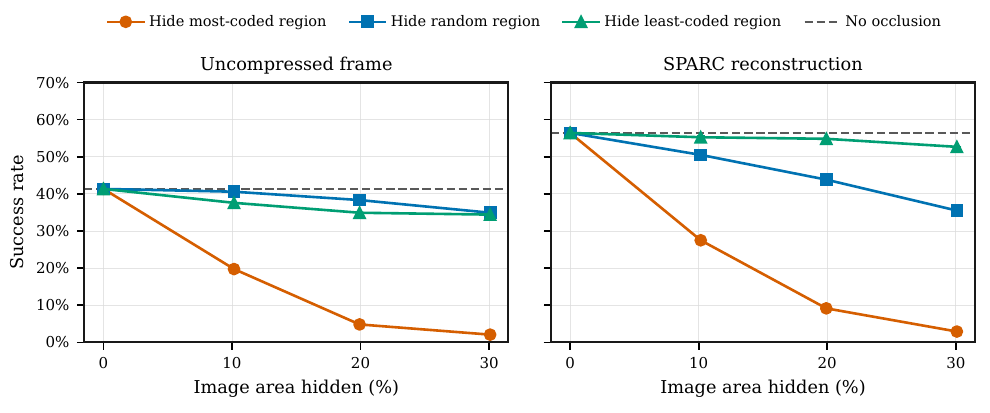}
\caption{Effect of occluding masked regions.}
\label{fig:check_critical_region}
\end{figure}
% \vspace{5pt}
% \begin{wrapfigure}[12]{r}{0.3\linewidth} % r: 오른쪽 정렬 (왼쪽 정렬은 l), 0.4\linewidth: 텍스트 단 폭의 40% 공간 할당
%     \centering
%     \caption{\textbf{Effect of occluding masked regions.}}
%     % wrapfigure가 할당한 공간 안에서 이미지가 꽉 차도록 1.0\linewidth로 설정합니다.
%     \includegraphics[width=1.0\linewidth]{Figures/robocasa_pi05_sparc_occlusion.pdf}
%     \label{fig:check_critical_region}
% \end{wrapfigure}
% wrapfigure 전체를 주석처리하고
% \begin{wrapfigure}[5]{r}{0.2\linewidth}
%     \centering
%     \includegraphics[width=1.0\linewidth]{Figures/robocasa_pi05_sparc_occlusion.pdf}
%     \caption{\textbf{Effect of occluding masked regions.}}
%     \label{fig:check_critical_region}
% \end{wrapfigure}
% \begin{wrapfigure}[10]{r}{0.43\linewidth}
%     \centering
%     \includegraphics[
%         width=\linewidth,
%         trim={20pt 10pt 5pt 0pt},
%         clip
%     ]{Figures/robocasa_pi05_sparc_occlusion.pdf}
%     \label{fig:transferability_pi0_pi05}
% \end{wrapfigure}

% \begin{wrapfigure}{r}{0.35\linewidth}
%     \centering
%     \vspace{-10pt}
%     \includegraphics[
%         width=\linewidth,
%         trim={20pt 0pt 5pt 0pt},
%         clip
%     ]{Figures/robocasa_pi05_sparc_occlusion.pdf}
%     \vspace{-10pt}
% \end{wrapfigure}
\begin{figure}[ht]
    \centering
    \includegraphics[width=0.625\linewidth]{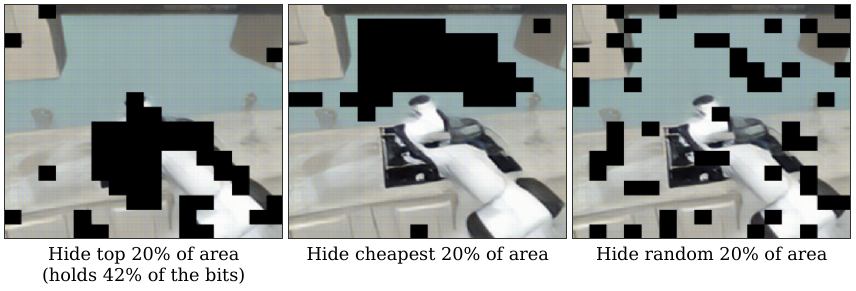}
    \caption{Input images examples.}
    \label{fig:task_critical_sample}
\end{figure}
\Cref{fig:check_critical_region} examines whether the regions prioritized by the mask selector are indeed task-critical. We rank image regions by the bitrate allocated by the mask selector and occlude the most-coded, least-coded, or randomly chosen regions, covering 10--30\% of the image area. Occlusion is applied both to uncompressed frames (left) and to SPARC reconstructions (right). In both settings, hiding the most-coded regions is by far the most harmful: the success rate is roughly halved already at 10\% occlusion and collapses from 41\% to 2\% on uncompressed frames and from 57\% to 3\% on SPARC reconstructions at 30\% occlusion. Hiding random regions causes only a moderate drop, to about 35\% and 36\%, respectively. Hiding the least-coded regions is nearly harmless for SPARC reconstructions (56\% $\rightarrow$ 53\%), whereas on uncompressed frames it degrades performance comparably to random occlusion (41\% $\rightarrow$ 34\%). This gap is likely because the least-coded regions are already reconstructed at low fidelity by SPARC, so occluding them removes little additional information. Notably, the most-coded regions are equally critical on uncompressed frames, indicating that the selector identifies content that is intrinsically important for the downstream task rather than an artifact of compression. These results confirm that the mask selector concentrates bits on the regions that matter most for downstream task performance.
\newpage
\section{Visual Quality Assessment}\label{sec:vqa}
To assess how image quality changes with our training recipe, we compare the quality of images produced by MS-ILLM and SPARC using image quality assessment metrics. On RoboCasa365 benchmarks, we collect 800 sample images of SPARC and MS-ILLM, where images samples of SPARC are optimized for $\pi_{0.5}$ across a range of different bpps and measure their average qualities using three well-known image quality assessment methods: (1) PSNR (Peak Signal-to-Noise Ratio), (2) SSIM (Structural Similarity Index Measure) and (3) LPIPS (Learned Perceptual Image Patch Similarity). 
\begin{figure}[h]
\centering
\begin{subfigure}[t!]{0.32\linewidth}
    \centering
    \caption{LPIPS $\downarrow$}
    \includegraphics[width=\linewidth]{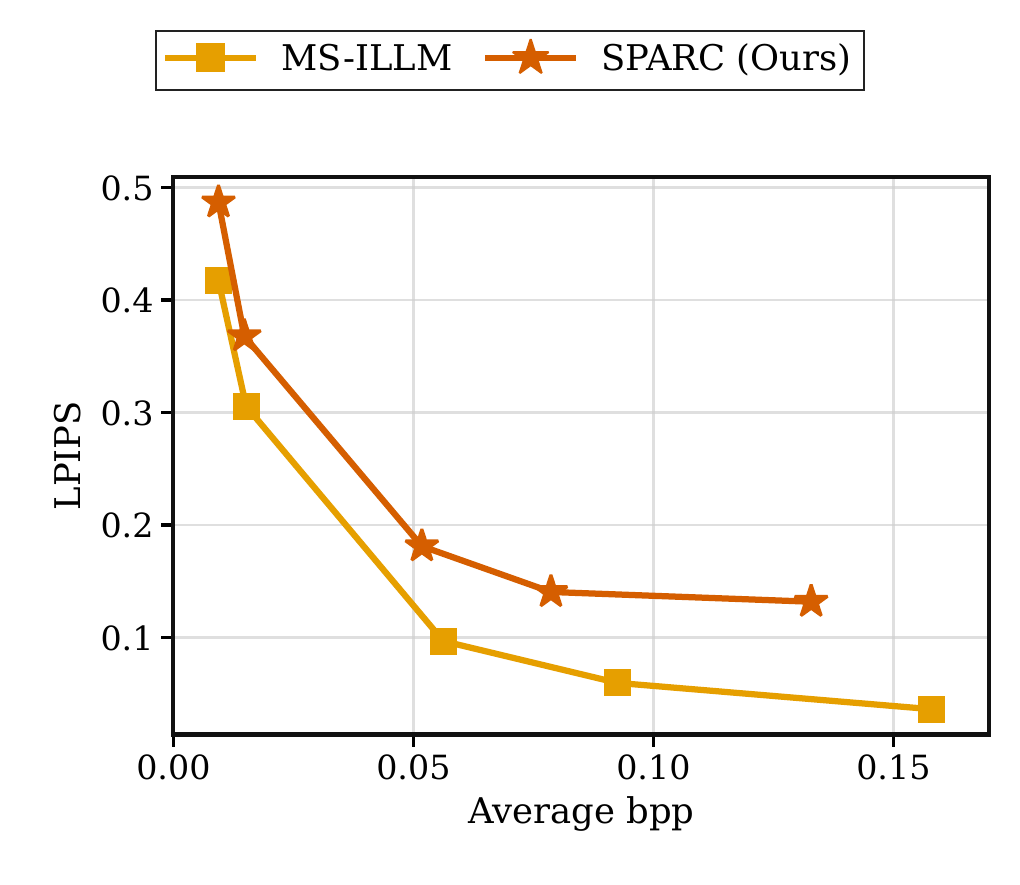}
    \label{fig:robocasa_sparc_msillm_lpips}
\end{subfigure}
\hfill
\begin{subfigure}[t!]{0.32\linewidth}
    \centering
    \caption{PSNR $\uparrow$}
    \includegraphics[width=\linewidth]{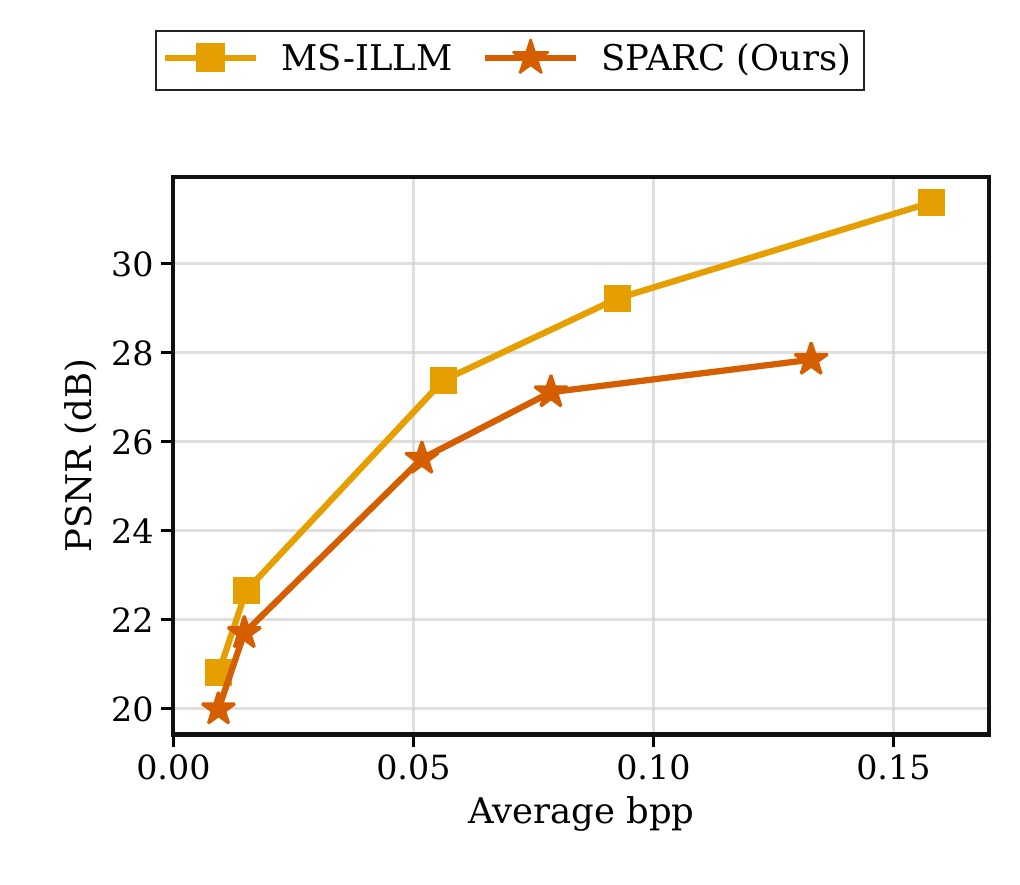}
    \label{fig:robocasa_sparc_msillm_psnr}
\end{subfigure}
\hfill
\begin{subfigure}[t!]{0.32\linewidth}
    \centering
    \caption{SSIM $\uparrow$}
    \includegraphics[width=\linewidth]{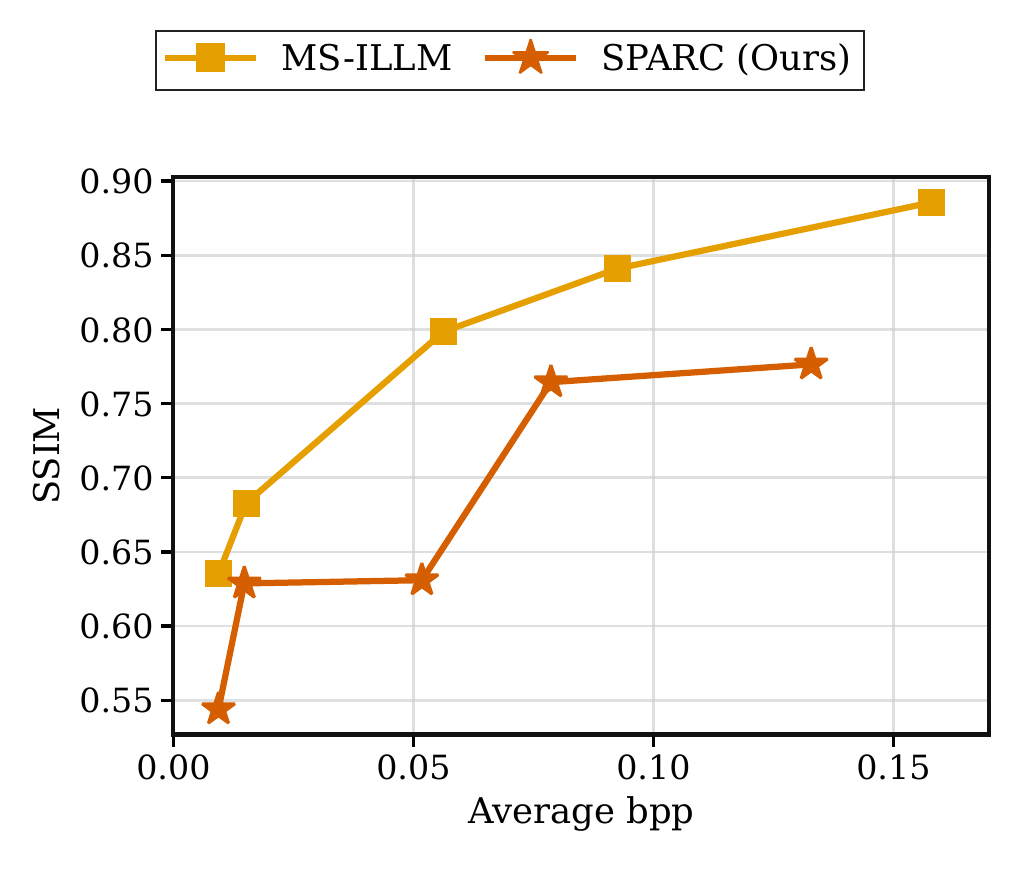}
    \label{fig:robocasa_sparc_msillm_ssim}
\end{subfigure}

\caption{Visual quality assessment between SPARC and MS-ILLM}
\label{fig:robocasa_sparc_vs_msillm}
\end{figure}

\Cref{fig:robocasa_sparc_vs_msillm} compares SPARC and MS-ILLM in terms of image quality. We observe that SPARC generates images with lower perceptual quality across a wide range of bpp values compared to MS-ILLM. An interesting characteristic shared by all three metrics is that the improvement in the quality of SPARC images follows that of MS-ILLM at low bpp values, but soon stagnates beyond a certain bpp range (0.03 $\sim$ 0.05 bpp), where the performance of downstream VLA models becomes sufficiently high. This implies that, beyond a certain point, improving perceptual quality does not contribute to further improvements in VLA model performance. SPARC is instead optimized for reconstructing task-relevant details for downstream VLA models.
\newpage
\section{Per-task Analysis}\label{sec:fca}
\begin{figure}[h]
    \centering
    \includegraphics[width=0.95\linewidth]{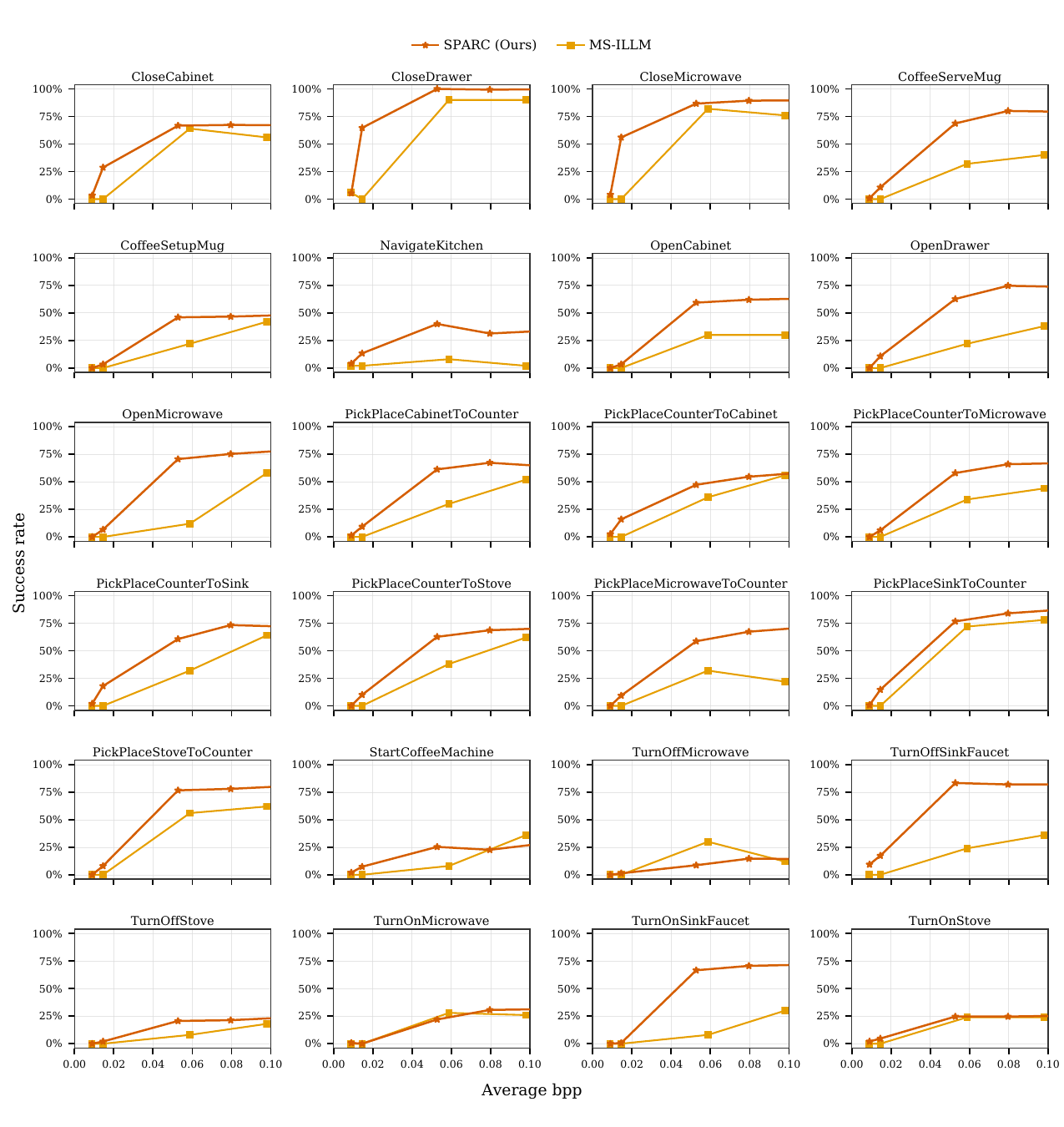}
    \caption{Per-task analysis of RoboCasa365 benchmarks}
    \label{fig:ablation_per_task}
\end{figure}
To analyze the advantages of SPARC in depth, we conduct a per-task analysis of SPARC compared to MS-ILLM on RoboCasa365 benchmarks with $\pi_{0.5}$ VLA model as the downstream VLA model. Following the evaluation pipeline in \Cref{ssec:setup}, we analyze 24 atomic tasks in RoboCasa365 benchmark.
In \Cref{fig:ablation_per_task}, we report the task success rate across a wide range of bpp values. We observe that SPARC achieves higher performance than MS-ILLM on most tasks, but the magnitude of the performance gain differs across tasks. SPARC tends to achieve greater performance gains on tasks for which identifying objects and the environment with more textural high-frequency details is important, such as \emph{CloseMicrowave}, \emph{CoffeeServeMug}, \emph{CloseDrawer} and \emph{TurnOffSinkFaucet}. In contrast, the performance gain from SPARC is not significant for tasks that require more complex dexterous operations, such as \emph{TurnOffStove}, \emph{TurnOnMicrowave}, and \emph{TurnOffMicrowave}. We hypothesize that removing complicated background details and focusing on task-specific regions helps identify target objects so that VLA models are not distracted by unnecessary background details. However, dexterous tasks that test fine-grained manipulation details \emph{rather} require more detailed information about how to manipulate the target objects. Nonetheless, SPARC is trained not to mask important latent features, which explains why the performance of SPARC is at least similar to or better than that of MS-ILLM on these tasks.
\newpage
\section{Real-World Rollout Examples}\label{sec:real_world_rollout}
% In the main text: SPARC vs. MS-ILLM comparison
\begin{figure*}[p]
\centering
\setlength{\tabcolsep}{0pt}
\renewcommand{\arraystretch}{0.6}
\begin{tabular}{@{}m{0.155\textwidth}@{\hspace{3pt}}m{0.75\textwidth}@{}}
\multicolumn{2}{@{}p{\linewidth}@{}}{\small \textbf{Carrot to drawer} \textit{``pick up the carrot and put it in the upper drawer''}} \\ \noalign{\vskip 2pt}
\parbox[c]{0.15\textwidth}{\raggedright\small \textcolor{ForestGreen}{\ding{51}}\,\textbf{SPARC}\\[-1pt]{\scriptsize\color{gray}0.051\,bpp}} & \begin{tabular}{@{}m{6pt}@{\hspace{1pt}}m{\dimexpr\linewidth-7pt\relax}@{}}\rotatebox{90}{\tiny\color{gray}top} & \includegraphics[width=\linewidth]{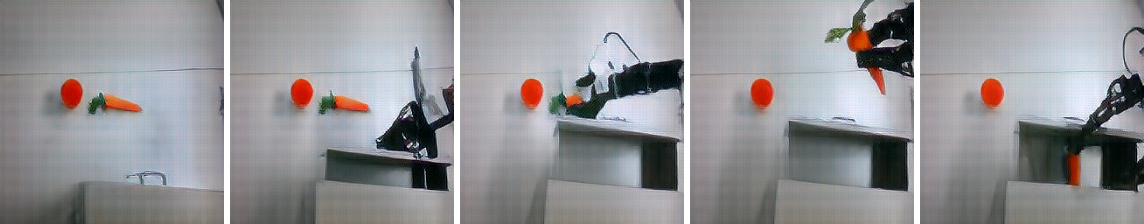} \\[1pt] \rotatebox{90}{\tiny\color{gray}side} & \includegraphics[width=\linewidth]{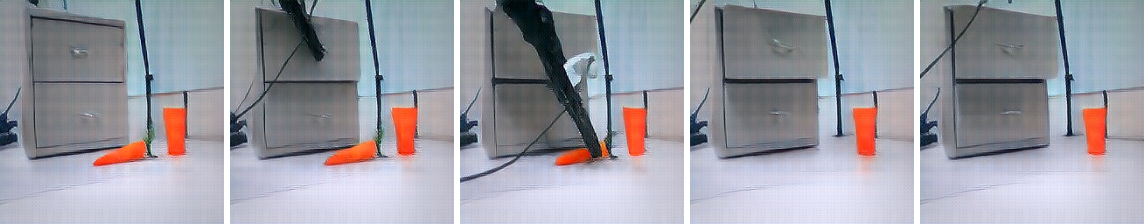} \\[1pt] \rotatebox{90}{\tiny\color{gray}wrist} & \includegraphics[width=\linewidth]{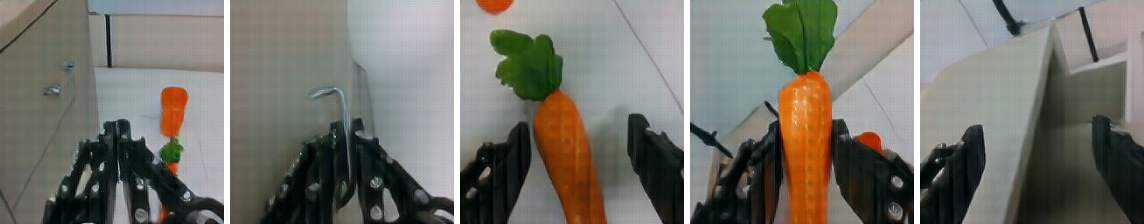}\end{tabular} \\ \noalign{\vskip 3pt}
\parbox[c]{0.15\textwidth}{\raggedright\small \textcolor{BrickRed}{\ding{55}}\,\textbf{MS-ILLM}\\[-1pt]{\scriptsize\color{gray}0.052\,bpp}\\[-1pt]{\scriptsize\itshape misses the grasp}} & \begin{tabular}{@{}m{6pt}@{\hspace{1pt}}m{\dimexpr\linewidth-7pt\relax}@{}}\rotatebox{90}{\tiny\color{gray}top} & \includegraphics[width=\linewidth]{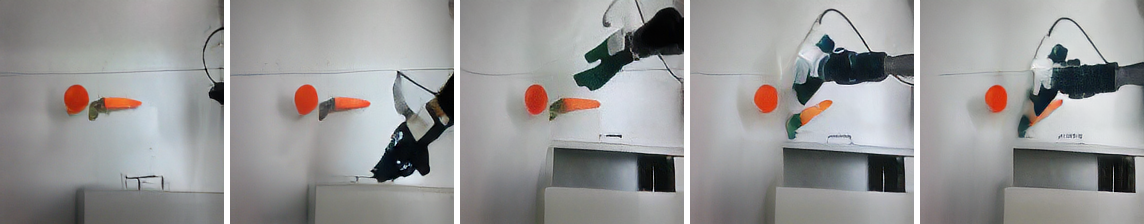} \\[1pt] \rotatebox{90}{\tiny\color{gray}side} & \includegraphics[width=\linewidth]{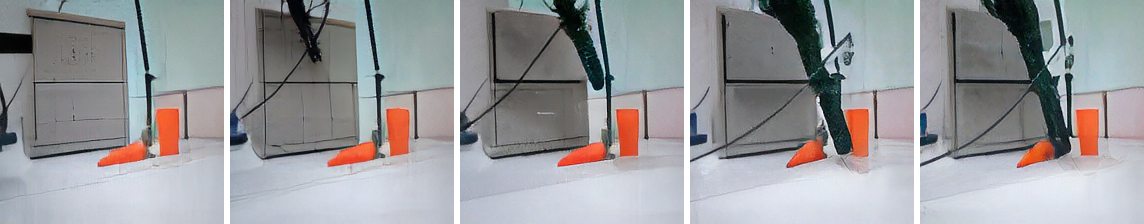} \\[1pt] \rotatebox{90}{\tiny\color{gray}wrist} & \includegraphics[width=\linewidth]{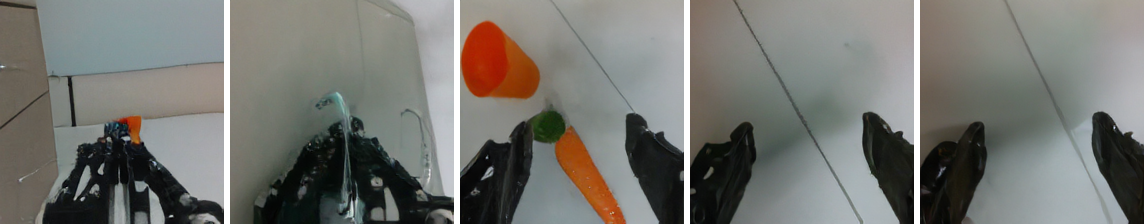}\end{tabular} \\ \noalign{\vskip 3pt}
\end{tabular}
\caption{\textbf{Real-world rollouts: SPARC vs.\ MS-ILLM.} For each task, SPARC (ours; top) succeeds while MS-ILLM (bottom) fails at a comparable bitrate. The italic note under each label names the failure mode. Each episode is shown by its top, side, and wrist cameras (three rows) at the same timesteps; time flows left to right, frames are the decoded images the $\pi_{0.5}$ policy received on the Piper arm, and the gray label gives the measured bitrate.}
\label{fig:realworld_comparison}
\end{figure*}

\begin{figure*}[p]
\centering
\setlength{\tabcolsep}{0pt}
\renewcommand{\arraystretch}{0.6}
\begin{tabular}{@{}m{0.155\textwidth}@{\hspace{3pt}}m{0.75\textwidth}@{}}
\multicolumn{2}{@{}p{\linewidth}@{}}{\small \textbf{Cabbage \& carrot to pot} \textit{``take the cabbage from the dish, and put it in the pot, then add the carrot''}} \\ \noalign{\vskip 2pt}
\parbox[c]{0.15\textwidth}{\raggedright\small \textcolor{ForestGreen}{\ding{51}}\,\textbf{SPARC}\\[-1pt]{\scriptsize\color{gray}0.152\,bpp}} & \begin{tabular}{@{}m{6pt}@{\hspace{1pt}}m{\dimexpr\linewidth-7pt\relax}@{}}\rotatebox{90}{\tiny\color{gray}top} & \includegraphics[width=\linewidth]{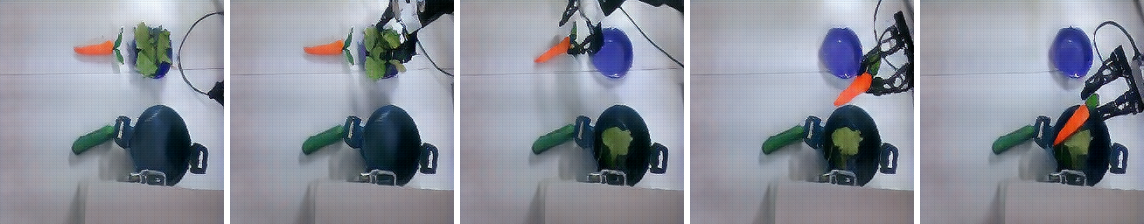} \\[1pt] \rotatebox{90}{\tiny\color{gray}side} & \includegraphics[width=\linewidth]{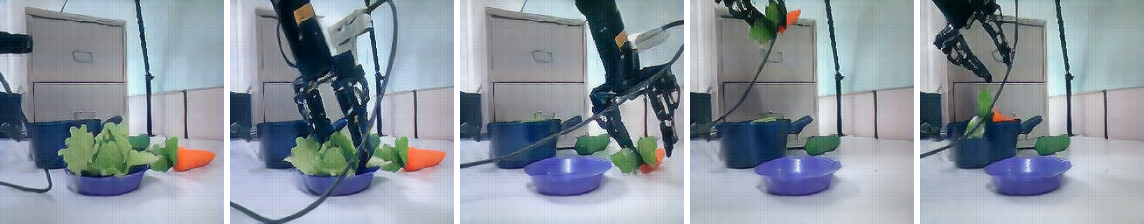} \\[1pt] \rotatebox{90}{\tiny\color{gray}wrist} & \includegraphics[width=\linewidth]{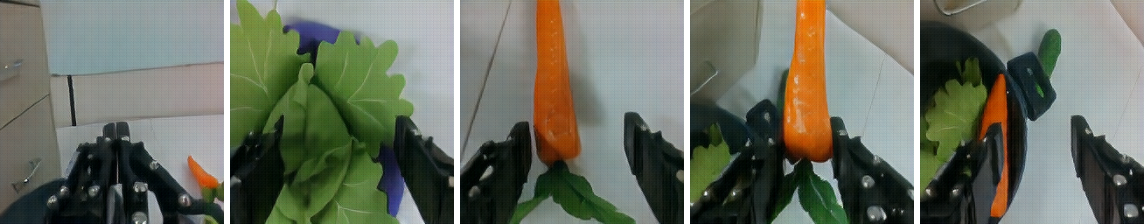}\end{tabular} \\ \noalign{\vskip 3pt}
\parbox[c]{0.15\textwidth}{\raggedright\small \textcolor{BrickRed}{\ding{55}}\,\textbf{MS-ILLM}\\[-1pt]{\scriptsize\color{gray}0.178\,bpp}\\[-1pt]{\scriptsize\itshape cannot lift the cabbage}} & \begin{tabular}{@{}m{6pt}@{\hspace{1pt}}m{\dimexpr\linewidth-7pt\relax}@{}}\rotatebox{90}{\tiny\color{gray}top} & \includegraphics[width=\linewidth]{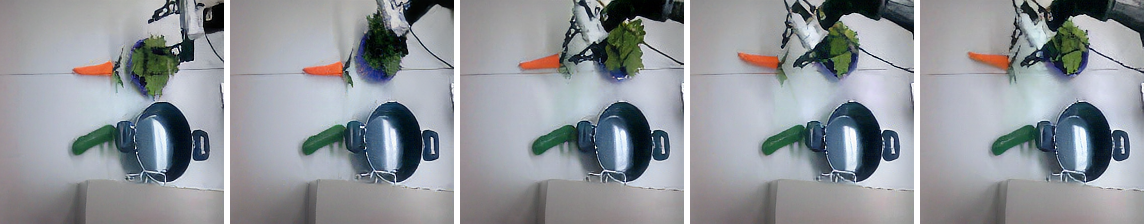} \\[1pt] \rotatebox{90}{\tiny\color{gray}side} & \includegraphics[width=\linewidth]{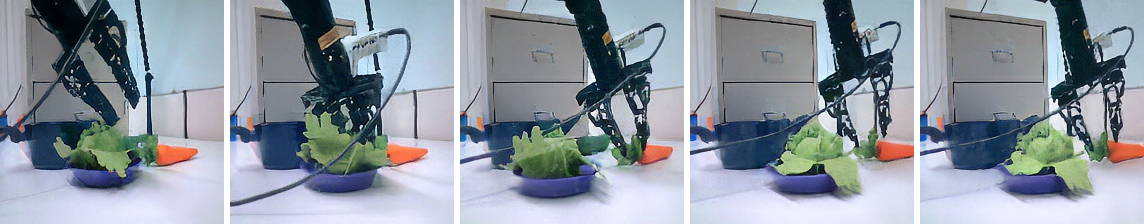} \\[1pt] \rotatebox{90}{\tiny\color{gray}wrist} & \includegraphics[width=\linewidth]{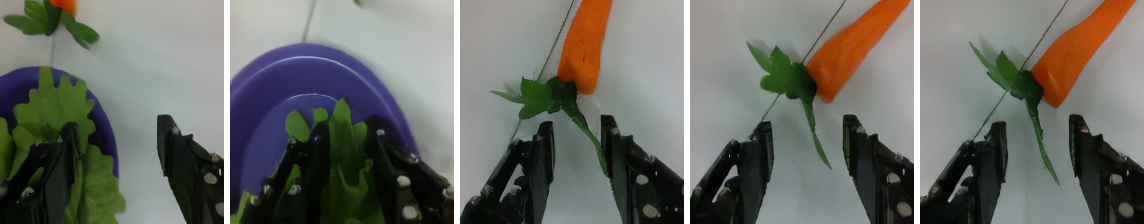}\end{tabular} \\ \noalign{\vskip 3pt}
\end{tabular}
\caption{\textbf{Real-world rollouts: SPARC vs.\ MS-ILLM.} (continued from Figure~\ref{fig:realworld_comparison}).}
\label{fig:realworld_comparison_2}
\end{figure*}

\begin{figure*}[p]
\centering
\setlength{\tabcolsep}{0pt}
\renewcommand{\arraystretch}{0.6}
\begin{tabular}{@{}m{0.155\textwidth}@{\hspace{3pt}}m{0.75\textwidth}@{}}
\multicolumn{2}{@{}p{\linewidth}@{}}{\small \textbf{Stack cups} \textit{``stack the purple cup on the orange one, and put the yellow one on top''}} \\ \noalign{\vskip 2pt}
\parbox[c]{0.15\textwidth}{\raggedright\small \textcolor{ForestGreen}{\ding{51}}\,\textbf{SPARC}\\[-1pt]{\scriptsize\color{gray}0.047\,bpp}} & \begin{tabular}{@{}m{6pt}@{\hspace{1pt}}m{\dimexpr\linewidth-7pt\relax}@{}}\rotatebox{90}{\tiny\color{gray}top} & \includegraphics[width=\linewidth]{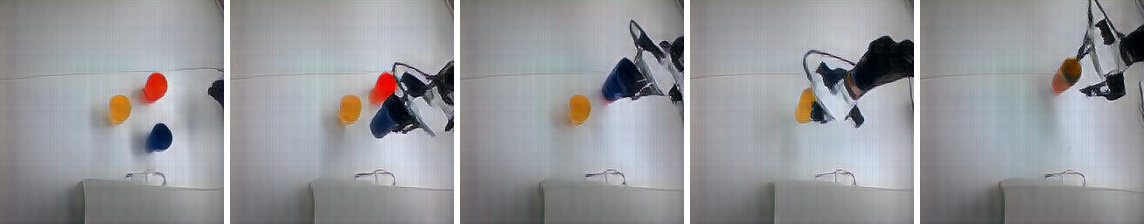} \\[1pt] \rotatebox{90}{\tiny\color{gray}side} & \includegraphics[width=\linewidth]{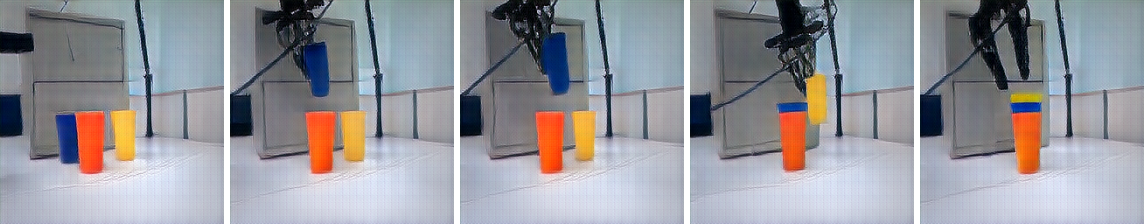} \\[1pt] \rotatebox{90}{\tiny\color{gray}wrist} & \includegraphics[width=\linewidth]{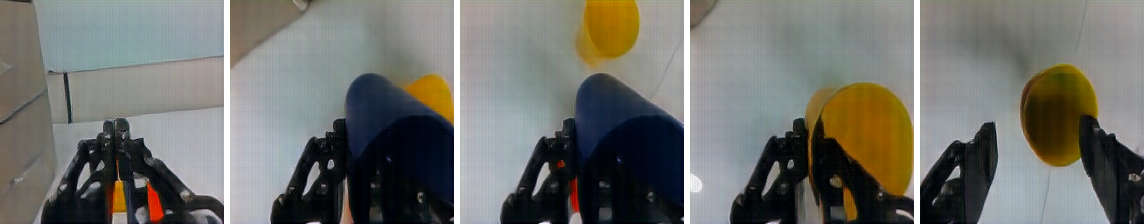}\end{tabular} \\ \noalign{\vskip 3pt}
\parbox[c]{0.15\textwidth}{\raggedright\small \textcolor{BrickRed}{\ding{55}}\,\textbf{MS-ILLM}\\[-1pt]{\scriptsize\color{gray}0.052\,bpp}\\[-1pt]{\scriptsize\itshape never places the yellow cup}} & \begin{tabular}{@{}m{6pt}@{\hspace{1pt}}m{\dimexpr\linewidth-7pt\relax}@{}}\rotatebox{90}{\tiny\color{gray}top} & \includegraphics[width=\linewidth]{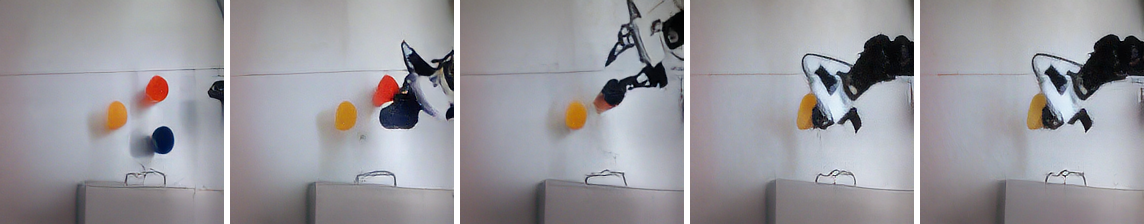} \\[1pt] \rotatebox{90}{\tiny\color{gray}side} & \includegraphics[width=\linewidth]{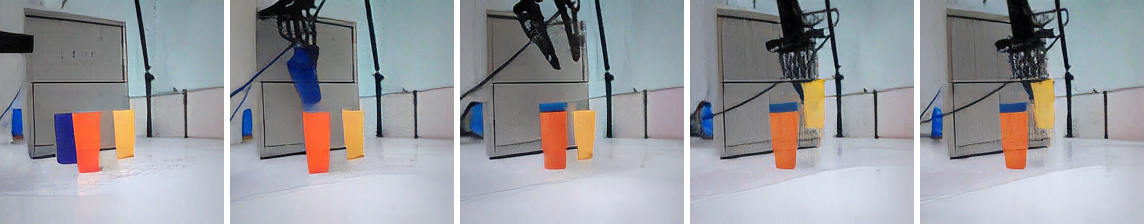} \\[1pt] \rotatebox{90}{\tiny\color{gray}wrist} & \includegraphics[width=\linewidth]{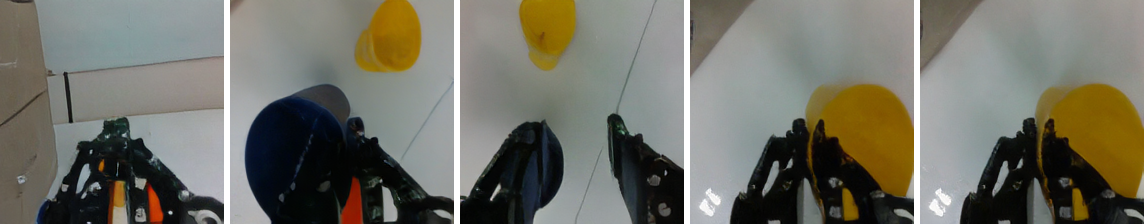}\end{tabular} \\ \noalign{\vskip 3pt}
\end{tabular}
\caption{\textbf{Real-world rollouts: SPARC vs.\ MS-ILLM.} (continued from Figure~\ref{fig:realworld_comparison}).}
\label{fig:realworld_comparison_3}
\end{figure*}

% Appendix
This section shows sample real-world rollouts of the $\pi_{0.5}$ policy on the Piper arm when its camera
observations are compressed by SPARC or by MS-ILLM, on the three tasks \textit{cabbage \& carrot to pot},
\textit{carrot to drawer}, and \textit{stack cups}. Each episode is shown from its top, side, and wrist
cameras at the same five timesteps, and the frames are the decoded images the policy actually received. The
label next to each episode gives the method, its outcome, and the measured bitrate; for failed episodes it
also names the failure mode.

\begin{figure*}[p]
\centering
\setlength{\tabcolsep}{0pt}
\renewcommand{\arraystretch}{0.6}
\begin{tabular}{@{}m{0.155\textwidth}@{\hspace{3pt}}m{0.75\textwidth}@{}}
\multicolumn{2}{@{}p{\linewidth}@{}}{\small \textbf{Carrot to drawer} \textit{``pick up the carrot and put it in the upper drawer''}} \\ \noalign{\vskip 2pt}
\parbox[c]{0.15\textwidth}{\raggedright\small \textcolor{ForestGreen}{\ding{51}}\,\textbf{SPARC}\\[-1pt]{\scriptsize\color{gray}0.051\,bpp}} & \begin{tabular}{@{}m{6pt}@{\hspace{1pt}}m{\dimexpr\linewidth-7pt\relax}@{}}\rotatebox{90}{\tiny\color{gray}top} & \includegraphics[width=\linewidth]{realworld/strips/scr4r_task0_q1_trial001_ep0_top.png} \\[1pt] \rotatebox{90}{\tiny\color{gray}side} & \includegraphics[width=\linewidth]{realworld/strips/scr4r_task0_q1_trial001_ep0_side.png} \\[1pt] \rotatebox{90}{\tiny\color{gray}wrist} & \includegraphics[width=\linewidth]{realworld/strips/scr4r_task0_q1_trial001_ep0_wrist.png}\end{tabular} \\ \noalign{\vskip 3pt}
\parbox[c]{0.15\textwidth}{\raggedright\small \textcolor{ForestGreen}{\ding{51}}\,\textbf{SPARC}\\[-1pt]{\scriptsize\color{gray}0.135\,bpp}} & \begin{tabular}{@{}m{6pt}@{\hspace{1pt}}m{\dimexpr\linewidth-7pt\relax}@{}}\rotatebox{90}{\tiny\color{gray}top} & \includegraphics[width=\linewidth]{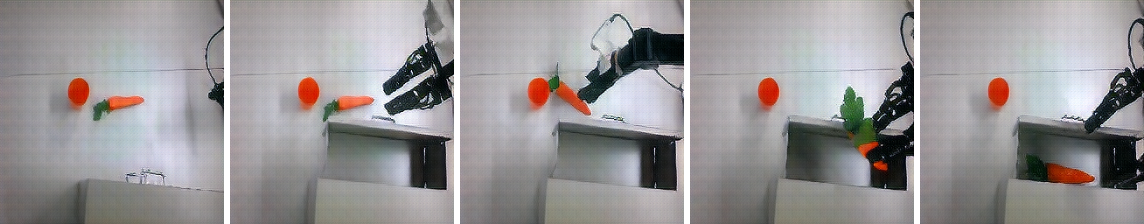} \\[1pt] \rotatebox{90}{\tiny\color{gray}side} & \includegraphics[width=\linewidth]{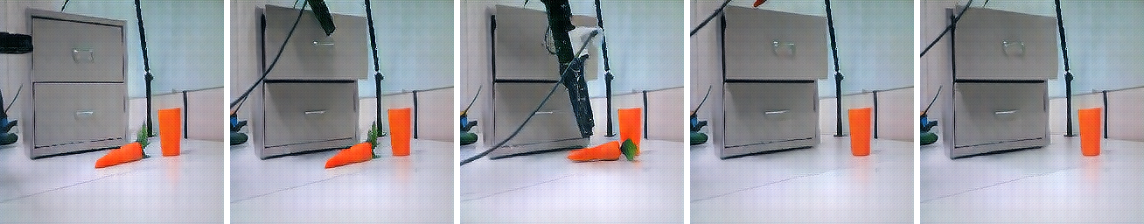} \\[1pt] \rotatebox{90}{\tiny\color{gray}wrist} & \includegraphics[width=\linewidth]{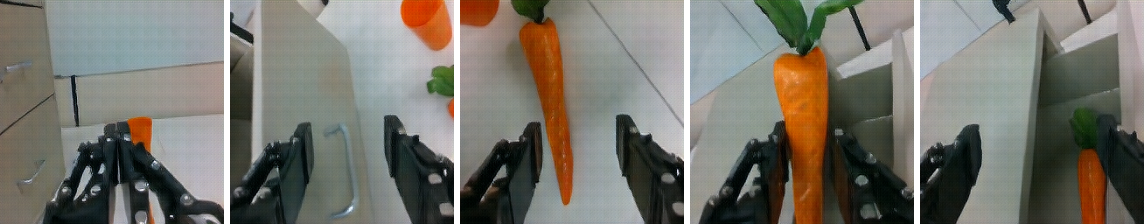}\end{tabular} \\ \noalign{\vskip 3pt}
\end{tabular}
\caption{\textbf{Real-world rollouts with SPARC: successful episodes} across tasks and bitrates. Each episode is shown by its top, side, and wrist cameras (three rows) at the same timesteps; time flows left to right, frames are the decoded images the $\pi_{0.5}$ policy received on the Piper arm, and the gray label gives the measured bitrate.}
\label{fig:realworld_sparc_success}
\end{figure*}

\begin{figure*}[p]
\centering
\setlength{\tabcolsep}{0pt}
\renewcommand{\arraystretch}{0.6}
\begin{tabular}{@{}m{0.155\textwidth}@{\hspace{3pt}}m{0.75\textwidth}@{}}
\multicolumn{2}{@{}p{\linewidth}@{}}{\small \textbf{Carrot to drawer} \textit{``pick up the carrot and put it in the upper drawer''}} \\ \noalign{\vskip 2pt}
\parbox[c]{0.15\textwidth}{\raggedright\small \textcolor{BrickRed}{\ding{55}}\,\textbf{SPARC}\\[-1pt]{\scriptsize\color{gray}0.504\,bpp}\\[-1pt]{\scriptsize\itshape misses the grasp}} & \begin{tabular}{@{}m{6pt}@{\hspace{1pt}}m{\dimexpr\linewidth-7pt\relax}@{}}\rotatebox{90}{\tiny\color{gray}top} & \includegraphics[width=\linewidth]{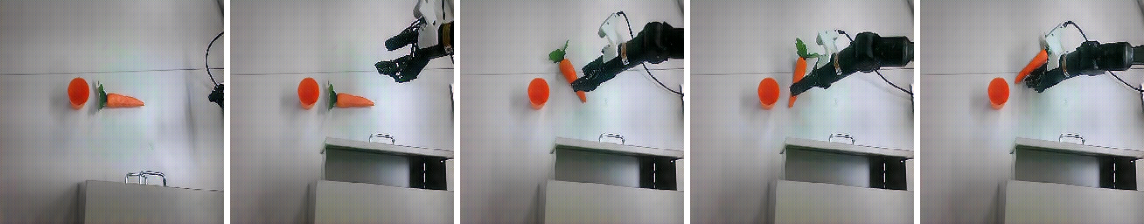} \\[1pt] \rotatebox{90}{\tiny\color{gray}side} & \includegraphics[width=\linewidth]{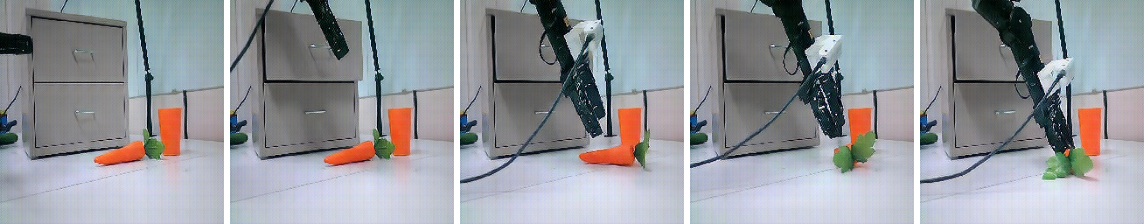} \\[1pt] \rotatebox{90}{\tiny\color{gray}wrist} & \includegraphics[width=\linewidth]{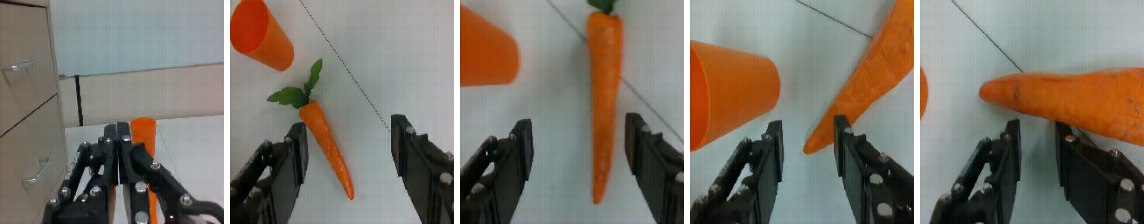}\end{tabular} \\ \noalign{\vskip 3pt}
\noalign{\vskip 5pt}
\multicolumn{2}{@{}p{\linewidth}@{}}{\small \textbf{Cabbage \& carrot to pot} \textit{``take the cabbage from the dish, and put it in the pot, then add the carrot''}} \\ \noalign{\vskip 2pt}
\parbox[c]{0.15\textwidth}{\raggedright\small \textcolor{BrickRed}{\ding{55}}\,\textbf{SPARC}\\[-1pt]{\scriptsize\color{gray}0.057\,bpp}\\[-1pt]{\scriptsize\itshape never picks up the carrot}} & \begin{tabular}{@{}m{6pt}@{\hspace{1pt}}m{\dimexpr\linewidth-7pt\relax}@{}}\rotatebox{90}{\tiny\color{gray}top} & \includegraphics[width=\linewidth]{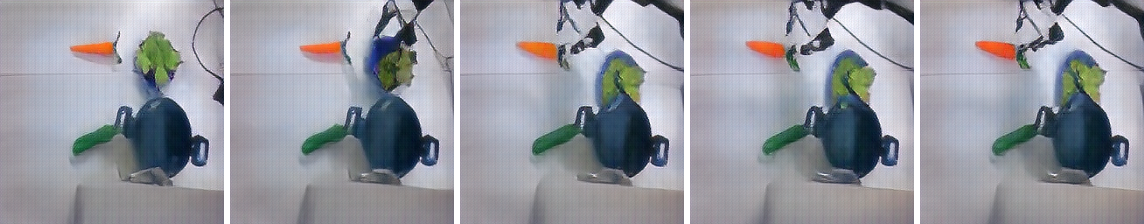} \\[1pt] \rotatebox{90}{\tiny\color{gray}side} & \includegraphics[width=\linewidth]{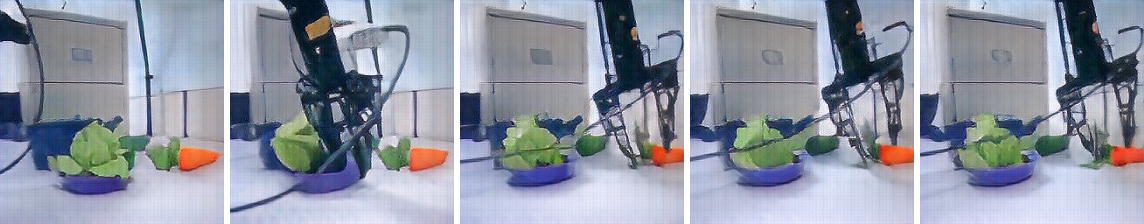} \\[1pt] \rotatebox{90}{\tiny\color{gray}wrist} & \includegraphics[width=\linewidth]{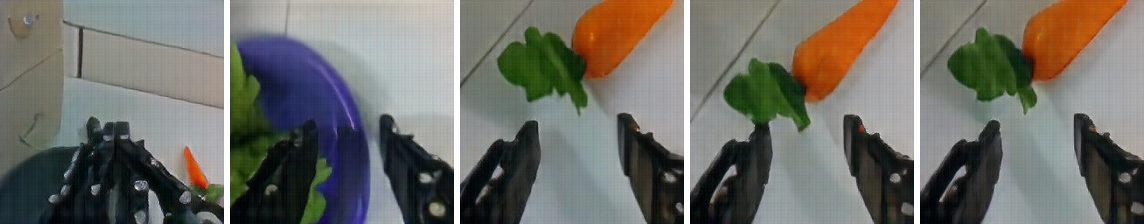}\end{tabular} \\ \noalign{\vskip 3pt}
\noalign{\vskip 5pt}
\multicolumn{2}{@{}p{\linewidth}@{}}{\small \textbf{Stack cups} \textit{``stack the purple cup on the orange one, and put the yellow one on top''}} \\ \noalign{\vskip 2pt}
\parbox[c]{0.15\textwidth}{\raggedright\small \textcolor{BrickRed}{\ding{55}}\,\textbf{SPARC}\\[-1pt]{\scriptsize\color{gray}0.127\,bpp}\\[-1pt]{\scriptsize\itshape stacks on the wrong cup}} & \begin{tabular}{@{}m{6pt}@{\hspace{1pt}}m{\dimexpr\linewidth-7pt\relax}@{}}\rotatebox{90}{\tiny\color{gray}top} & \includegraphics[width=\linewidth]{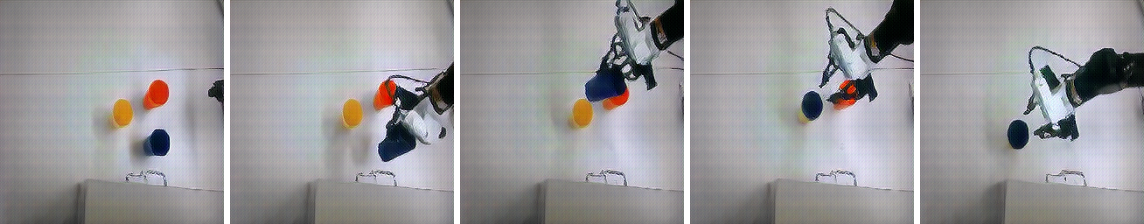} \\[1pt] \rotatebox{90}{\tiny\color{gray}side} & \includegraphics[width=\linewidth]{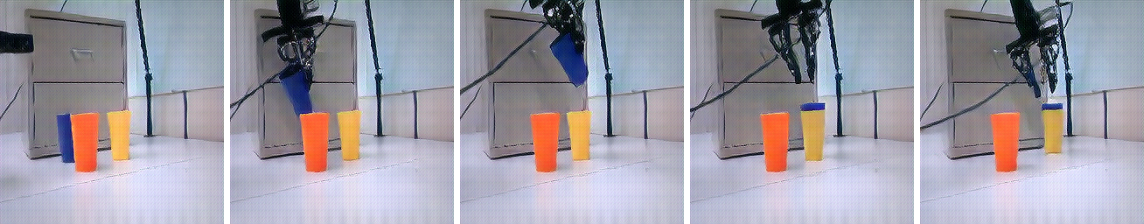} \\[1pt] \rotatebox{90}{\tiny\color{gray}wrist} & \includegraphics[width=\linewidth]{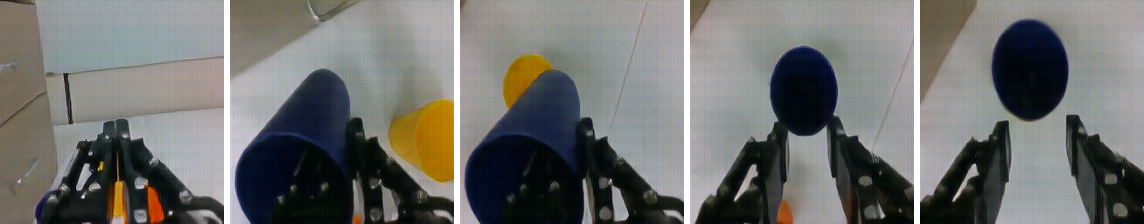}\end{tabular} \\ \noalign{\vskip 3pt}
\end{tabular}
\caption{\textbf{Real-world rollouts with SPARC: failure cases.} The italic note under each label names the failure mode. Each episode is shown by its top, side, and wrist cameras (three rows) at the same timesteps; time flows left to right, frames are the decoded images the $\pi_{0.5}$ policy received on the Piper arm, and the gray label gives the measured bitrate.}
\label{fig:realworld_sparc_failure}
\end{figure*}

\begin{figure*}[p]
\centering
\setlength{\tabcolsep}{0pt}
\renewcommand{\arraystretch}{0.6}
\begin{tabular}{@{}m{0.155\textwidth}@{\hspace{3pt}}m{0.75\textwidth}@{}}
\multicolumn{2}{@{}p{\linewidth}@{}}{\small \textbf{Carrot to drawer} \textit{``pick up the carrot and put it in the upper drawer''}} \\ \noalign{\vskip 2pt}
\parbox[c]{0.15\textwidth}{\raggedright\small \textcolor{BrickRed}{\ding{55}}\,\textbf{MS-ILLM}\\[-1pt]{\scriptsize\color{gray}0.052\,bpp}\\[-1pt]{\scriptsize\itshape misses the grasp}} & \begin{tabular}{@{}m{6pt}@{\hspace{1pt}}m{\dimexpr\linewidth-7pt\relax}@{}}\rotatebox{90}{\tiny\color{gray}top} & \includegraphics[width=\linewidth]{realworld/strips/msillm_task0_q1_trial002_ep5_top.png} \\[1pt] \rotatebox{90}{\tiny\color{gray}side} & \includegraphics[width=\linewidth]{realworld/strips/msillm_task0_q1_trial002_ep5_side.png} \\[1pt] \rotatebox{90}{\tiny\color{gray}wrist} & \includegraphics[width=\linewidth]{realworld/strips/msillm_task0_q1_trial002_ep5_wrist.png}\end{tabular} \\ \noalign{\vskip 3pt}
\parbox[c]{0.15\textwidth}{\raggedright\small \textcolor{BrickRed}{\ding{55}}\,\textbf{MS-ILLM}\\[-1pt]{\scriptsize\color{gray}0.574\,bpp}\\[-1pt]{\scriptsize\itshape misses the grasp}} & \begin{tabular}{@{}m{6pt}@{\hspace{1pt}}m{\dimexpr\linewidth-7pt\relax}@{}}\rotatebox{90}{\tiny\color{gray}top} & \includegraphics[width=\linewidth]{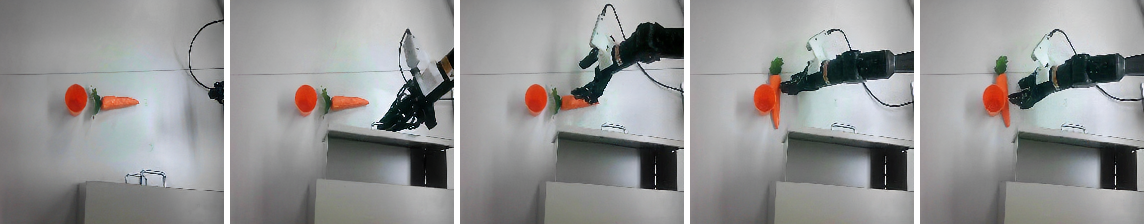} \\[1pt] \rotatebox{90}{\tiny\color{gray}side} & \includegraphics[width=\linewidth]{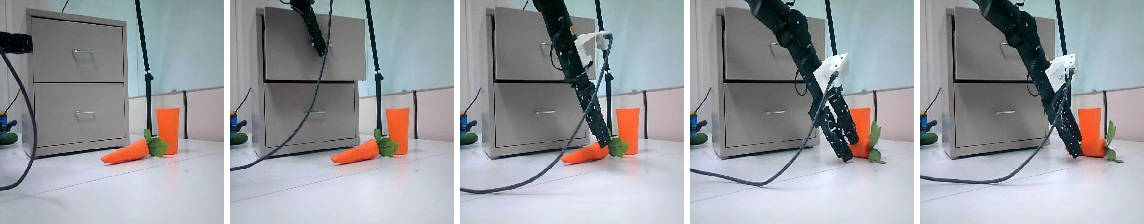} \\[1pt] \rotatebox{90}{\tiny\color{gray}wrist} & \includegraphics[width=\linewidth]{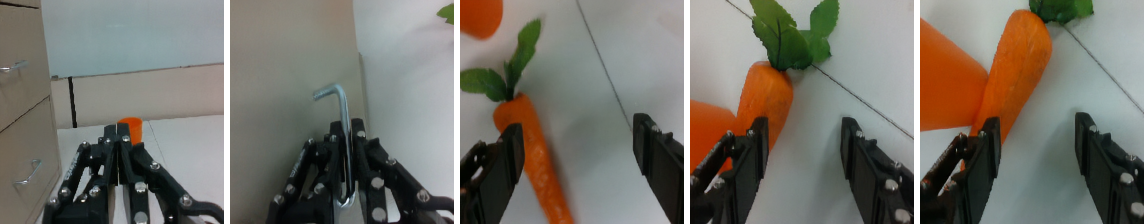}\end{tabular} \\ \noalign{\vskip 3pt}
\end{tabular}
\caption{\textbf{Real-world rollouts with MS-ILLM: failure cases} across tasks and bitrates. The italic note under each label names the failure mode. Each episode is shown by its top, side, and wrist cameras (three rows) at the same timesteps; time flows left to right, frames are the decoded images the $\pi_{0.5}$ policy received on the Piper arm, and the gray label gives the measured bitrate.}
\label{fig:realworld_msillm_failure}
\end{figure*}

Figure~\ref{fig:realworld_comparison} pairs, for each task, a successful SPARC episode with a failed MS-ILLM
episode at a comparable bitrate. Figure~\ref{fig:realworld_sparc_success} collects successful SPARC episodes
across tasks and bitrates, and Figure~\ref{fig:realworld_sparc_failure} shows representative SPARC failures.
Figure~\ref{fig:realworld_msillm_failure} shows MS-ILLM failures across tasks and bitrates for comparison.
\newpage
       % failure analysis section + table

\begin{figure*}[p]
\centering
\setlength{\tabcolsep}{0pt}
\renewcommand{\arraystretch}{0.6}
\begin{tabular}{@{}m{0.155\textwidth}@{\hspace{3pt}}m{0.75\textwidth}@{}}
\multicolumn{2}{@{}p{\linewidth}@{}}{\small \textbf{Carrot to drawer} \textit{``pick up the carrot and put it in the upper drawer''}} \\ \noalign{\vskip 2pt}
\parbox[c]{0.15\textwidth}{\raggedright\small \textcolor{ForestGreen}{\ding{51}}\,\textbf{SPARC}\\[-1pt]{\scriptsize\color{gray}0.496\,bpp}} & \begin{tabular}{@{}m{6pt}@{\hspace{1pt}}m{\dimexpr\linewidth-7pt\relax}@{}}\rotatebox{90}{\tiny\color{gray}top} & \includegraphics[width=\linewidth]{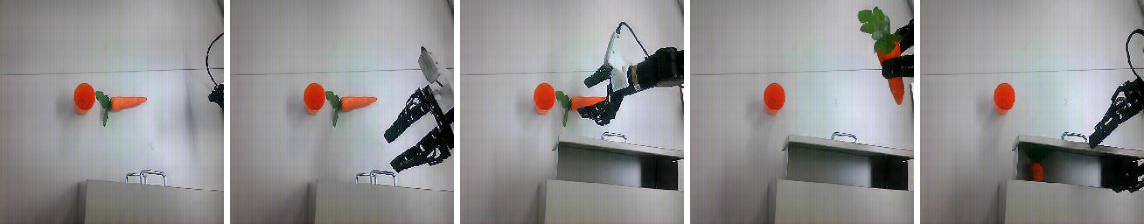} \\[1pt] \rotatebox{90}{\tiny\color{gray}side} & \includegraphics[width=\linewidth]{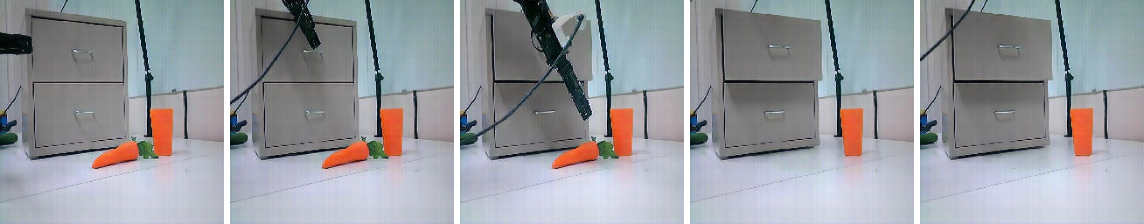} \\[1pt] \rotatebox{90}{\tiny\color{gray}wrist} & \includegraphics[width=\linewidth]{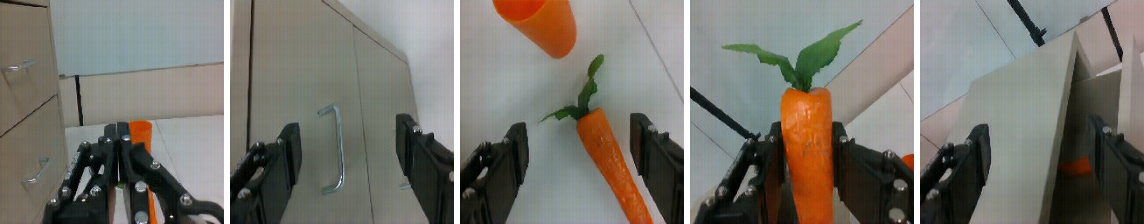}\end{tabular} \\ \noalign{\vskip 3pt}
\parbox[c]{0.15\textwidth}{\raggedright\small \textcolor{ForestGreen}{\ding{51}}\,\textbf{SPARC}\\[-1pt]{\scriptsize\color{gray}0.017\,bpp}} & \begin{tabular}{@{}m{6pt}@{\hspace{1pt}}m{\dimexpr\linewidth-7pt\relax}@{}}\rotatebox{90}{\tiny\color{gray}top} & \includegraphics[width=\linewidth]{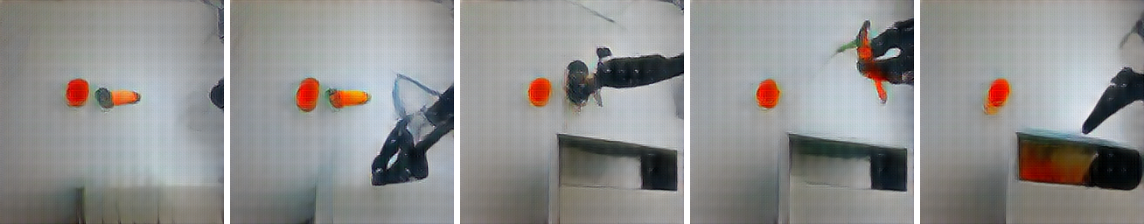} \\[1pt] \rotatebox{90}{\tiny\color{gray}side} & \includegraphics[width=\linewidth]{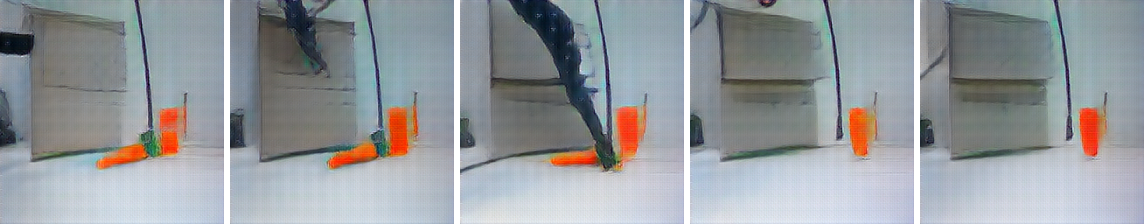} \\[1pt] \rotatebox{90}{\tiny\color{gray}wrist} & \includegraphics[width=\linewidth]{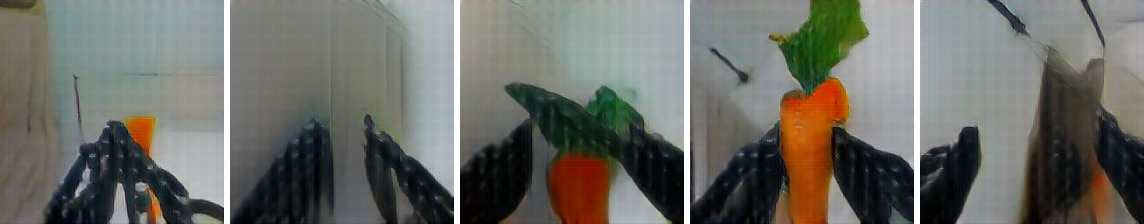}\end{tabular} \\ \noalign{\vskip 3pt}
\noalign{\vskip 5pt}
\multicolumn{2}{@{}p{\linewidth}@{}}{\small \textbf{Cabbage \& carrot to pot} \textit{``take the cabbage from the dish, and put it in the pot, then add the carrot''}} \\ \noalign{\vskip 2pt}
\parbox[c]{0.15\textwidth}{\raggedright\small \textcolor{ForestGreen}{\ding{51}}\,\textbf{SPARC}\\[-1pt]{\scriptsize\color{gray}0.057\,bpp}} & \begin{tabular}{@{}m{6pt}@{\hspace{1pt}}m{\dimexpr\linewidth-7pt\relax}@{}}\rotatebox{90}{\tiny\color{gray}top} & \includegraphics[width=\linewidth]{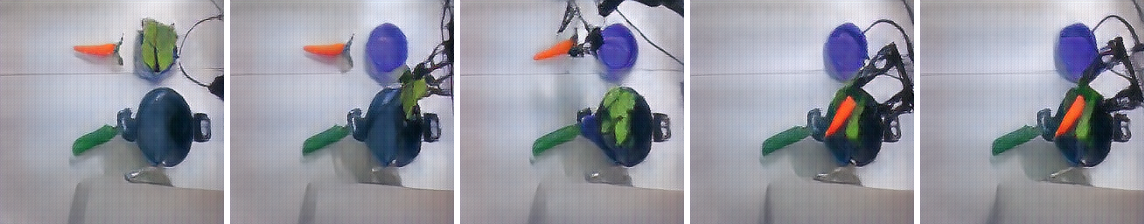} \\[1pt] \rotatebox{90}{\tiny\color{gray}side} & \includegraphics[width=\linewidth]{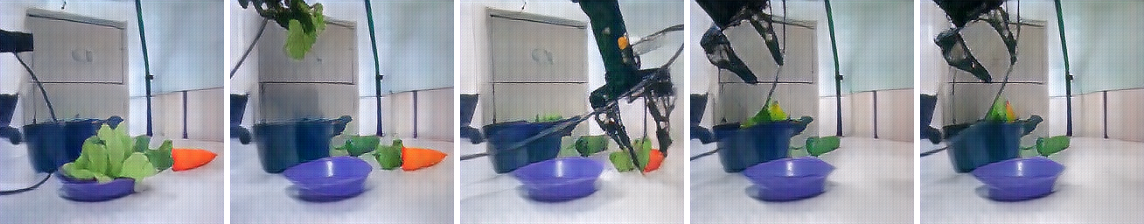} \\[1pt] \rotatebox{90}{\tiny\color{gray}wrist} & \includegraphics[width=\linewidth]{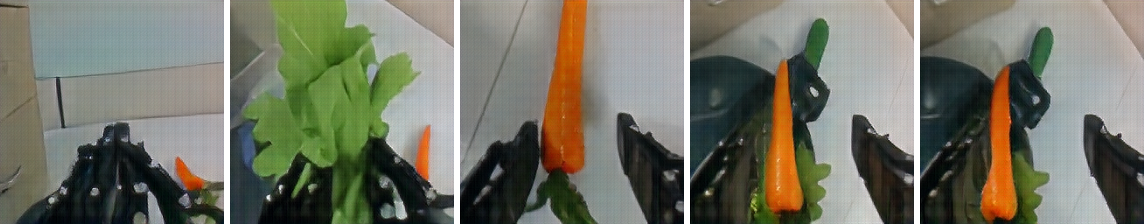}\end{tabular} \\ \noalign{\vskip 3pt}
\end{tabular}
\caption{\textbf{Real-world rollouts with SPARC: successful episodes} (continued from Figure~\ref{fig:realworld_sparc_success}).}
\label{fig:realworld_sparc_success_2}
\end{figure*}

\begin{figure*}[p]
\centering
\setlength{\tabcolsep}{0pt}
\renewcommand{\arraystretch}{0.6}
\begin{tabular}{@{}m{0.155\textwidth}@{\hspace{3pt}}m{0.75\textwidth}@{}}
\multicolumn{2}{@{}p{\linewidth}@{}}{\small \textbf{Cabbage \& carrot to pot} \textit{``take the cabbage from the dish, and put it in the pot, then add the carrot''}} \\ \noalign{\vskip 2pt}
\parbox[c]{0.15\textwidth}{\raggedright\small \textcolor{ForestGreen}{\ding{51}}\,\textbf{SPARC}\\[-1pt]{\scriptsize\color{gray}0.152\,bpp}} & \begin{tabular}{@{}m{6pt}@{\hspace{1pt}}m{\dimexpr\linewidth-7pt\relax}@{}}\rotatebox{90}{\tiny\color{gray}top} & \includegraphics[width=\linewidth]{realworld/strips/scr4r_task1_q3_trial001_ep5_top.png} \\[1pt] \rotatebox{90}{\tiny\color{gray}side} & \includegraphics[width=\linewidth]{realworld/strips/scr4r_task1_q3_trial001_ep5_side.png} \\[1pt] \rotatebox{90}{\tiny\color{gray}wrist} & \includegraphics[width=\linewidth]{realworld/strips/scr4r_task1_q3_trial001_ep5_wrist.png}\end{tabular} \\ \noalign{\vskip 3pt}
\parbox[c]{0.15\textwidth}{\raggedright\small \textcolor{ForestGreen}{\ding{51}}\,\textbf{SPARC}\\[-1pt]{\scriptsize\color{gray}0.019\,bpp}} & \begin{tabular}{@{}m{6pt}@{\hspace{1pt}}m{\dimexpr\linewidth-7pt\relax}@{}}\rotatebox{90}{\tiny\color{gray}top} & \includegraphics[width=\linewidth]{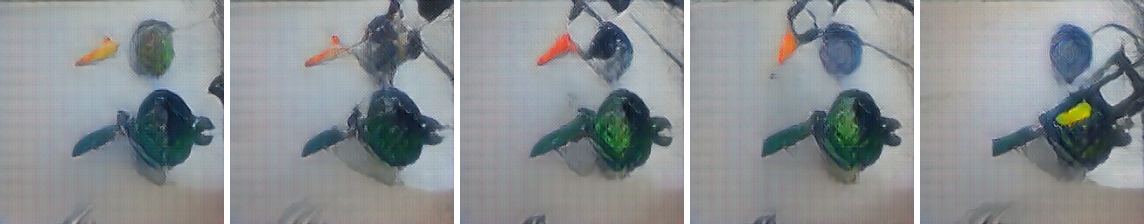} \\[1pt] \rotatebox{90}{\tiny\color{gray}side} & \includegraphics[width=\linewidth]{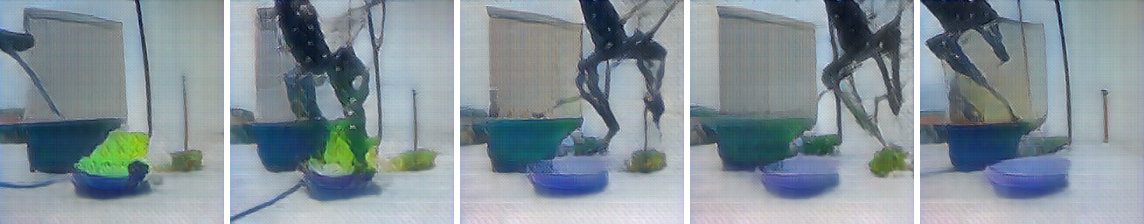} \\[1pt] \rotatebox{90}{\tiny\color{gray}wrist} & \includegraphics[width=\linewidth]{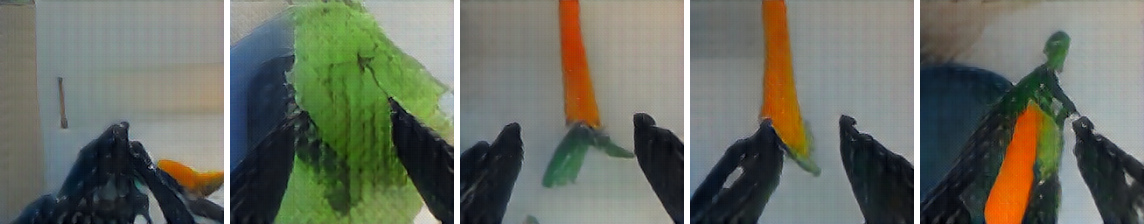}\end{tabular} \\ \noalign{\vskip 3pt}
\noalign{\vskip 5pt}
\multicolumn{2}{@{}p{\linewidth}@{}}{\small \textbf{Stack cups} \textit{``stack the purple cup on the orange one, and put the yellow one on top''}} \\ \noalign{\vskip 2pt}
\parbox[c]{0.15\textwidth}{\raggedright\small \textcolor{ForestGreen}{\ding{51}}\,\textbf{SPARC}\\[-1pt]{\scriptsize\color{gray}0.047\,bpp}} & \begin{tabular}{@{}m{6pt}@{\hspace{1pt}}m{\dimexpr\linewidth-7pt\relax}@{}}\rotatebox{90}{\tiny\color{gray}top} & \includegraphics[width=\linewidth]{realworld/strips/scr4r_task2_q1_trial002_ep4_top.png} \\[1pt] \rotatebox{90}{\tiny\color{gray}side} & \includegraphics[width=\linewidth]{realworld/strips/scr4r_task2_q1_trial002_ep4_side.png} \\[1pt] \rotatebox{90}{\tiny\color{gray}wrist} & \includegraphics[width=\linewidth]{realworld/strips/scr4r_task2_q1_trial002_ep4_wrist.png}\end{tabular} \\ \noalign{\vskip 3pt}
\end{tabular}
\caption{\textbf{Real-world rollouts with SPARC: successful episodes} (continued from Figure~\ref{fig:realworld_sparc_success}).}
\label{fig:realworld_sparc_success_3}
\end{figure*}

\begin{figure*}[p]
\centering
\setlength{\tabcolsep}{0pt}
\renewcommand{\arraystretch}{0.6}
\begin{tabular}{@{}m{0.155\textwidth}@{\hspace{3pt}}m{0.75\textwidth}@{}}
\multicolumn{2}{@{}p{\linewidth}@{}}{\small \textbf{Stack cups} \textit{``stack the purple cup on the orange one, and put the yellow one on top''}} \\ \noalign{\vskip 2pt}
\parbox[c]{0.15\textwidth}{\raggedright\small \textcolor{ForestGreen}{\ding{51}}\,\textbf{SPARC}\\[-1pt]{\scriptsize\color{gray}0.443\,bpp}} & \begin{tabular}{@{}m{6pt}@{\hspace{1pt}}m{\dimexpr\linewidth-7pt\relax}@{}}\rotatebox{90}{\tiny\color{gray}top} & \includegraphics[width=\linewidth]{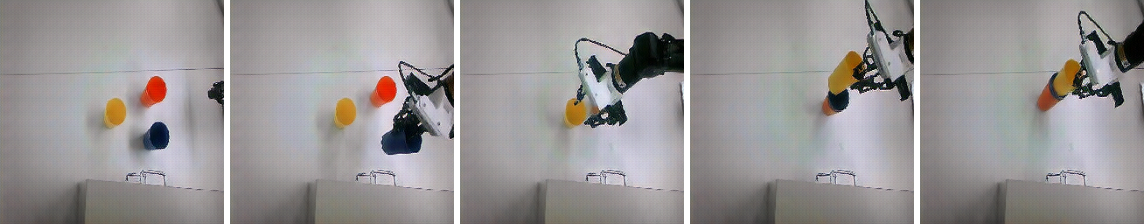} \\[1pt] \rotatebox{90}{\tiny\color{gray}side} & \includegraphics[width=\linewidth]{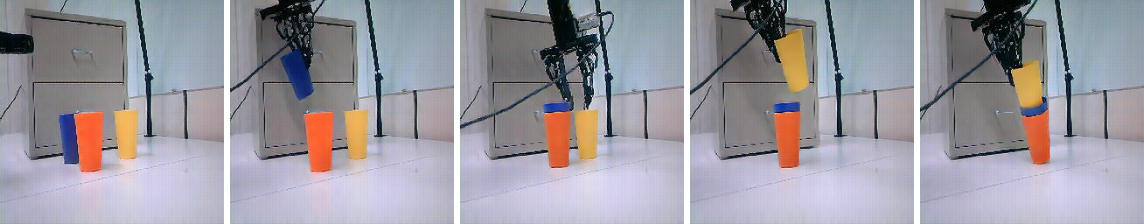} \\[1pt] \rotatebox{90}{\tiny\color{gray}wrist} & \includegraphics[width=\linewidth]{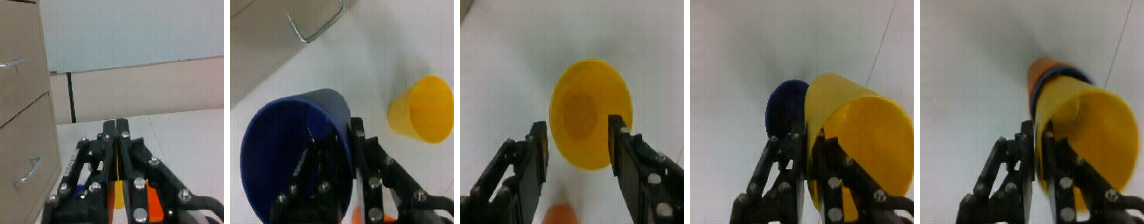}\end{tabular} \\ \noalign{\vskip 3pt}
\parbox[c]{0.15\textwidth}{\raggedright\small \textcolor{ForestGreen}{\ding{51}}\,\textbf{SPARC}\\[-1pt]{\scriptsize\color{gray}0.016\,bpp}} & \begin{tabular}{@{}m{6pt}@{\hspace{1pt}}m{\dimexpr\linewidth-7pt\relax}@{}}\rotatebox{90}{\tiny\color{gray}top} & \includegraphics[width=\linewidth]{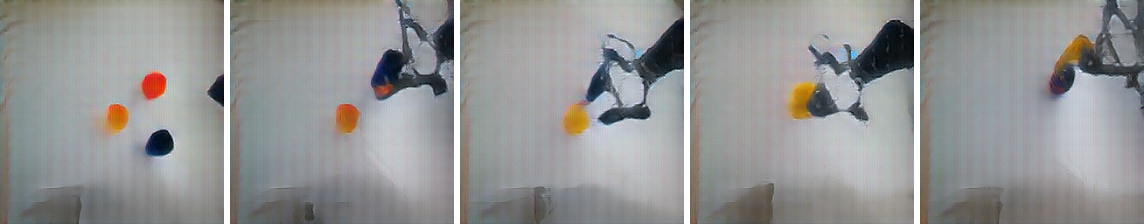} \\[1pt] \rotatebox{90}{\tiny\color{gray}side} & \includegraphics[width=\linewidth]{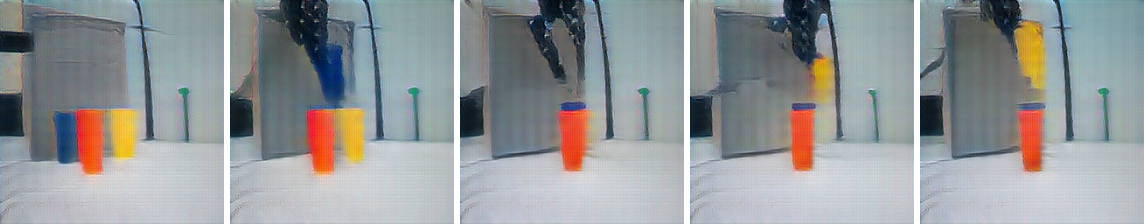} \\[1pt] \rotatebox{90}{\tiny\color{gray}wrist} & \includegraphics[width=\linewidth]{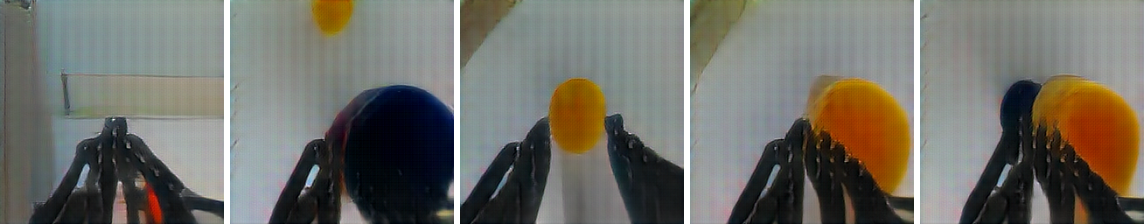}\end{tabular} \\ \noalign{\vskip 3pt}
\end{tabular}
\caption{\textbf{Real-world rollouts with SPARC: successful episodes} (continued from Figure~\ref{fig:realworld_sparc_success}).}
\label{fig:realworld_sparc_success_4}
\end{figure*}

  % SPARC successes

  % SPARC failures

\begin{figure*}[p]
\centering
\setlength{\tabcolsep}{0pt}
\renewcommand{\arraystretch}{0.6}
\begin{tabular}{@{}m{0.155\textwidth}@{\hspace{3pt}}m{0.75\textwidth}@{}}
\multicolumn{2}{@{}p{\linewidth}@{}}{\small \textbf{Carrot to drawer} \textit{``pick up the carrot and put it in the upper drawer''}} \\ \noalign{\vskip 2pt}
\parbox[c]{0.15\textwidth}{\raggedright\small \textcolor{BrickRed}{\ding{55}}\,\textbf{MS-ILLM}\\[-1pt]{\scriptsize\color{gray}0.015\,bpp}\\[-1pt]{\scriptsize\itshape never opens the drawer}} & \begin{tabular}{@{}m{6pt}@{\hspace{1pt}}m{\dimexpr\linewidth-7pt\relax}@{}}\rotatebox{90}{\tiny\color{gray}top} & \includegraphics[width=\linewidth]{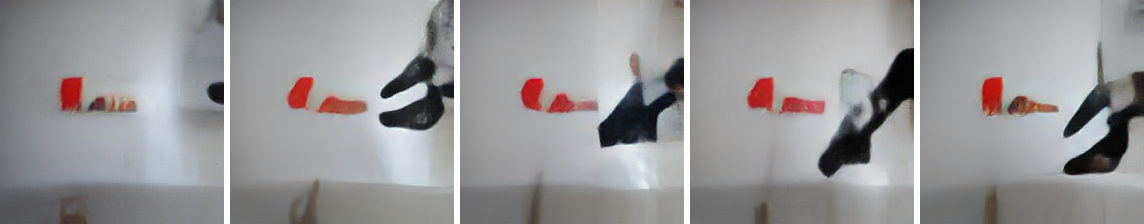} \\[1pt] \rotatebox{90}{\tiny\color{gray}side} & \includegraphics[width=\linewidth]{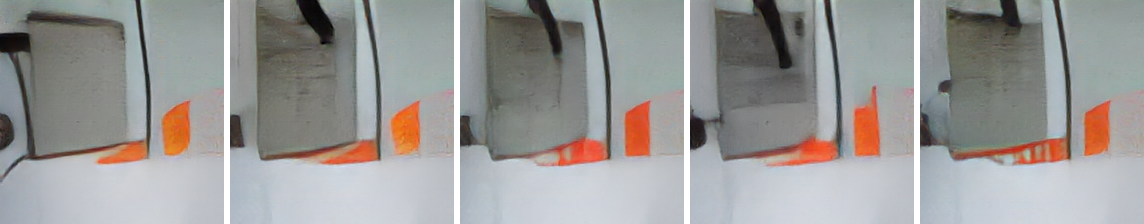} \\[1pt] \rotatebox{90}{\tiny\color{gray}wrist} & \includegraphics[width=\linewidth]{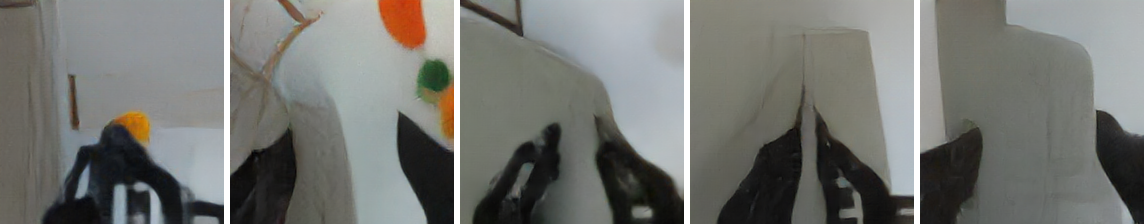}\end{tabular} \\ \noalign{\vskip 3pt}
\noalign{\vskip 5pt}
\multicolumn{2}{@{}p{\linewidth}@{}}{\small \textbf{Cabbage \& carrot to pot} \textit{``take the cabbage from the dish, and put it in the pot, then add the carrot''}} \\ \noalign{\vskip 2pt}
\parbox[c]{0.15\textwidth}{\raggedright\small \textcolor{BrickRed}{\ding{55}}\,\textbf{MS-ILLM}\\[-1pt]{\scriptsize\color{gray}0.064\,bpp}\\[-1pt]{\scriptsize\itshape never places the carrot}} & \begin{tabular}{@{}m{6pt}@{\hspace{1pt}}m{\dimexpr\linewidth-7pt\relax}@{}}\rotatebox{90}{\tiny\color{gray}top} & \includegraphics[width=\linewidth]{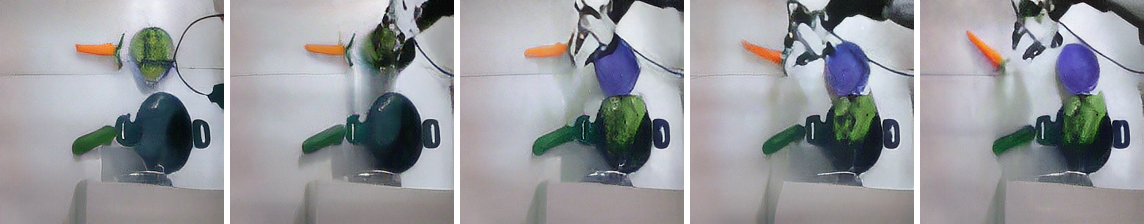} \\[1pt] \rotatebox{90}{\tiny\color{gray}side} & \includegraphics[width=\linewidth]{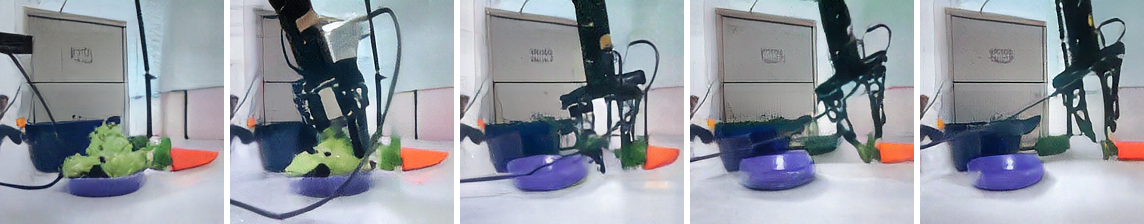} \\[1pt] \rotatebox{90}{\tiny\color{gray}wrist} & \includegraphics[width=\linewidth]{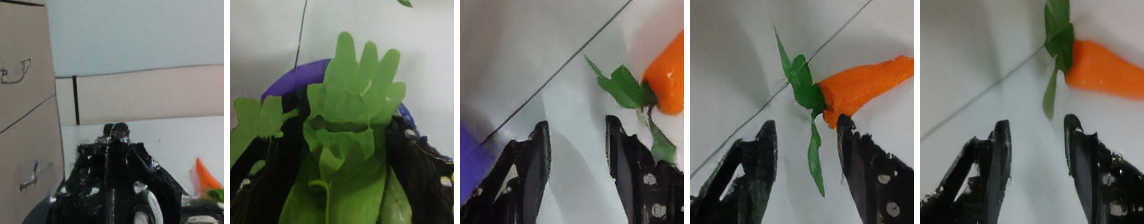}\end{tabular} \\ \noalign{\vskip 3pt}
\parbox[c]{0.15\textwidth}{\raggedright\small \textcolor{BrickRed}{\ding{55}}\,\textbf{MS-ILLM}\\[-1pt]{\scriptsize\color{gray}0.178\,bpp}\\[-1pt]{\scriptsize\itshape cannot lift the cabbage}} & \begin{tabular}{@{}m{6pt}@{\hspace{1pt}}m{\dimexpr\linewidth-7pt\relax}@{}}\rotatebox{90}{\tiny\color{gray}top} & \includegraphics[width=\linewidth]{realworld/strips/msillm_task1_q3_trial001_ep3_top.png} \\[1pt] \rotatebox{90}{\tiny\color{gray}side} & \includegraphics[width=\linewidth]{realworld/strips/msillm_task1_q3_trial001_ep3_side.png} \\[1pt] \rotatebox{90}{\tiny\color{gray}wrist} & \includegraphics[width=\linewidth]{realworld/strips/msillm_task1_q3_trial001_ep3_wrist.png}\end{tabular} \\ \noalign{\vskip 3pt}
\end{tabular}
\caption{\textbf{Real-world rollouts with MS-ILLM: failure cases} (continued from Figure~\ref{fig:realworld_msillm_failure}).}
\label{fig:realworld_msillm_failure_2}
\end{figure*}

\begin{figure*}[p]
\centering
\setlength{\tabcolsep}{0pt}
\renewcommand{\arraystretch}{0.6}
\begin{tabular}{@{}m{0.155\textwidth}@{\hspace{3pt}}m{0.75\textwidth}@{}}
\multicolumn{2}{@{}p{\linewidth}@{}}{\small \textbf{Cabbage \& carrot to pot} \textit{``take the cabbage from the dish, and put it in the pot, then add the carrot''}} \\ \noalign{\vskip 2pt}
\parbox[c]{0.15\textwidth}{\raggedright\small \textcolor{BrickRed}{\ding{55}}\,\textbf{MS-ILLM}\\[-1pt]{\scriptsize\color{gray}0.683\,bpp}\\[-1pt]{\scriptsize\itshape never places the carrot}} & \begin{tabular}{@{}m{6pt}@{\hspace{1pt}}m{\dimexpr\linewidth-7pt\relax}@{}}\rotatebox{90}{\tiny\color{gray}top} & \includegraphics[width=\linewidth]{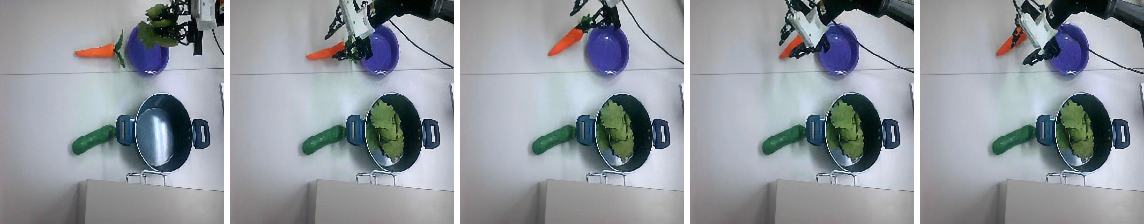} \\[1pt] \rotatebox{90}{\tiny\color{gray}side} & \includegraphics[width=\linewidth]{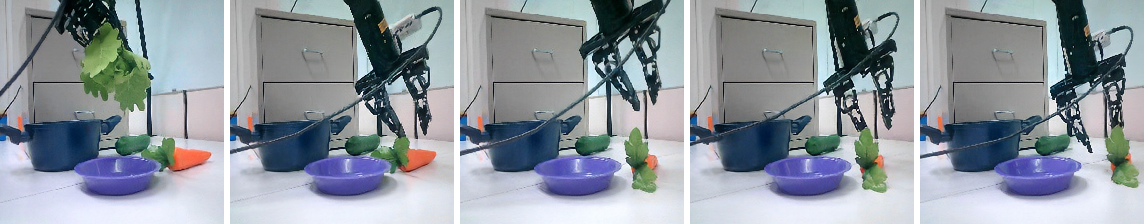} \\[1pt] \rotatebox{90}{\tiny\color{gray}wrist} & \includegraphics[width=\linewidth]{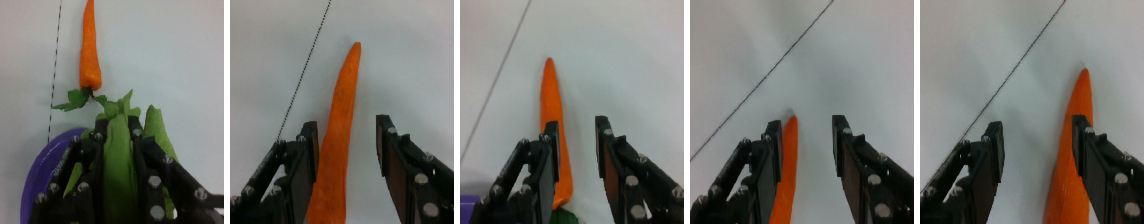}\end{tabular} \\ \noalign{\vskip 3pt}
\noalign{\vskip 5pt}
\multicolumn{2}{@{}p{\linewidth}@{}}{\small \textbf{Stack cups} \textit{``stack the purple cup on the orange one, and put the yellow one on top''}} \\ \noalign{\vskip 2pt}
\parbox[c]{0.15\textwidth}{\raggedright\small \textcolor{BrickRed}{\ding{55}}\,\textbf{MS-ILLM}\\[-1pt]{\scriptsize\color{gray}0.052\,bpp}\\[-1pt]{\scriptsize\itshape never places the yellow cup}} & \begin{tabular}{@{}m{6pt}@{\hspace{1pt}}m{\dimexpr\linewidth-7pt\relax}@{}}\rotatebox{90}{\tiny\color{gray}top} & \includegraphics[width=\linewidth]{realworld/strips/msillm_task2_q1_trial001_ep4_top.png} \\[1pt] \rotatebox{90}{\tiny\color{gray}side} & \includegraphics[width=\linewidth]{realworld/strips/msillm_task2_q1_trial001_ep4_side.png} \\[1pt] \rotatebox{90}{\tiny\color{gray}wrist} & \includegraphics[width=\linewidth]{realworld/strips/msillm_task2_q1_trial001_ep4_wrist.png}\end{tabular} \\ \noalign{\vskip 3pt}
\parbox[c]{0.15\textwidth}{\raggedright\small \textcolor{BrickRed}{\ding{55}}\,\textbf{MS-ILLM}\\[-1pt]{\scriptsize\color{gray}0.149\,bpp}\\[-1pt]{\scriptsize\itshape knocks the yellow cup over}} & \begin{tabular}{@{}m{6pt}@{\hspace{1pt}}m{\dimexpr\linewidth-7pt\relax}@{}}\rotatebox{90}{\tiny\color{gray}top} & \includegraphics[width=\linewidth]{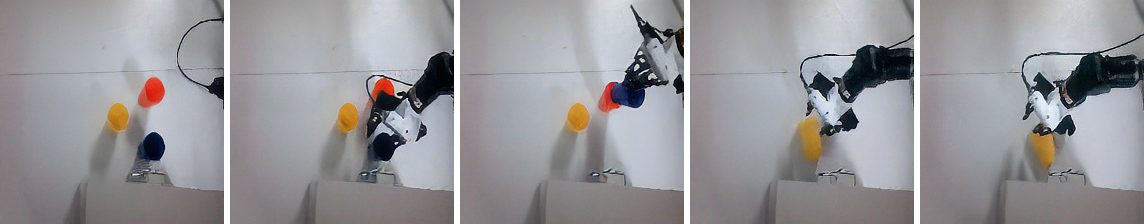} \\[1pt] \rotatebox{90}{\tiny\color{gray}side} & \includegraphics[width=\linewidth]{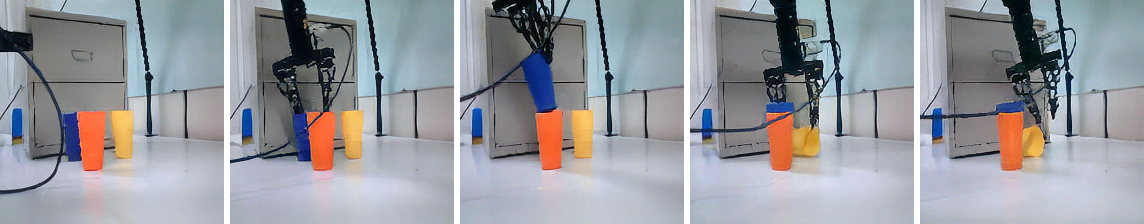} \\[1pt] \rotatebox{90}{\tiny\color{gray}wrist} & \includegraphics[width=\linewidth]{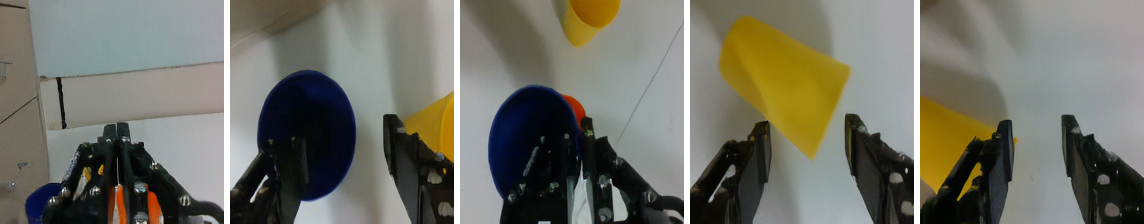}\end{tabular} \\ \noalign{\vskip 3pt}
\end{tabular}
\caption{\textbf{Real-world rollouts with MS-ILLM: failure cases} (continued from Figure~\ref{fig:realworld_msillm_failure}).}
\label{fig:realworld_msillm_failure_3}
\end{figure*}

\begin{figure*}[p]
\centering
\setlength{\tabcolsep}{0pt}
\renewcommand{\arraystretch}{0.6}
\begin{tabular}{@{}m{0.155\textwidth}@{\hspace{3pt}}m{0.75\textwidth}@{}}
\multicolumn{2}{@{}p{\linewidth}@{}}{\small \textbf{Stack cups} \textit{``stack the purple cup on the orange one, and put the yellow one on top''}} \\ \noalign{\vskip 2pt}
\parbox[c]{0.15\textwidth}{\raggedright\small \textcolor{BrickRed}{\ding{55}}\,\textbf{MS-ILLM}\\[-1pt]{\scriptsize\color{gray}0.564\,bpp}\\[-1pt]{\scriptsize\itshape never places the yellow cup}} & \begin{tabular}{@{}m{6pt}@{\hspace{1pt}}m{\dimexpr\linewidth-7pt\relax}@{}}\rotatebox{90}{\tiny\color{gray}top} & \includegraphics[width=\linewidth]{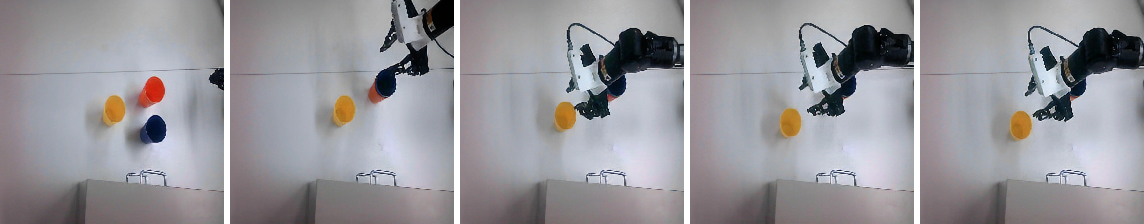} \\[1pt] \rotatebox{90}{\tiny\color{gray}side} & \includegraphics[width=\linewidth]{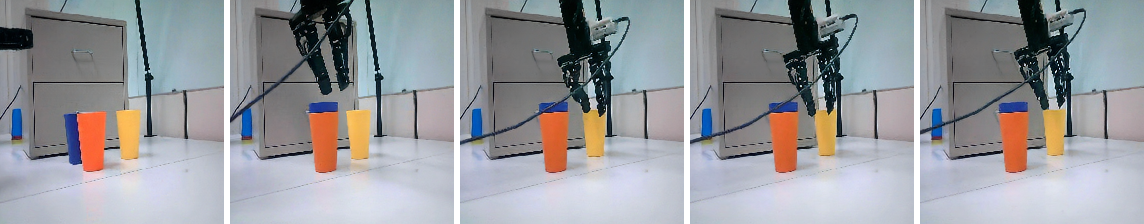} \\[1pt] \rotatebox{90}{\tiny\color{gray}wrist} & \includegraphics[width=\linewidth]{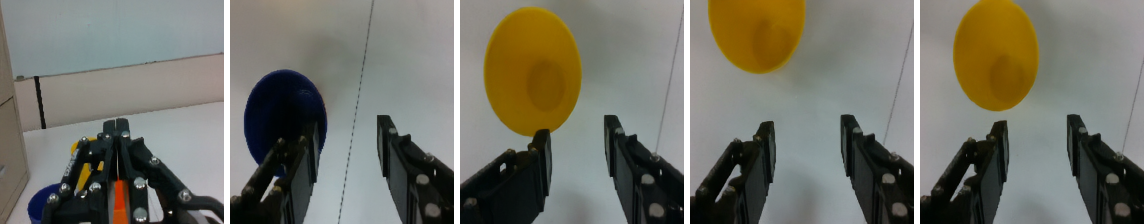}\end{tabular} \\ \noalign{\vskip 3pt}
\parbox[c]{0.15\textwidth}{\raggedright\small \textcolor{BrickRed}{\ding{55}}\,\textbf{MS-ILLM}\\[-1pt]{\scriptsize\color{gray}0.015\,bpp}\\[-1pt]{\scriptsize\itshape fails to grasp a cup}} & \begin{tabular}{@{}m{6pt}@{\hspace{1pt}}m{\dimexpr\linewidth-7pt\relax}@{}}\rotatebox{90}{\tiny\color{gray}top} & \includegraphics[width=\linewidth]{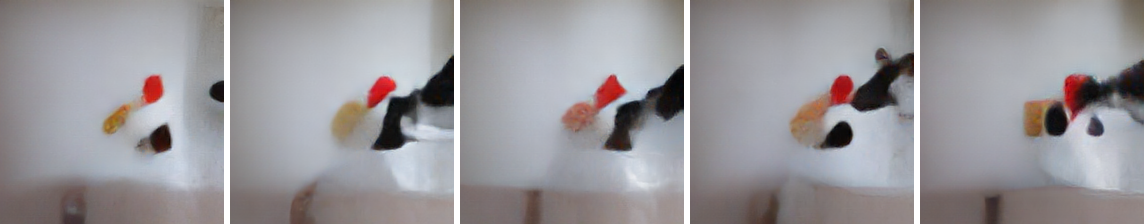} \\[1pt] \rotatebox{90}{\tiny\color{gray}side} & \includegraphics[width=\linewidth]{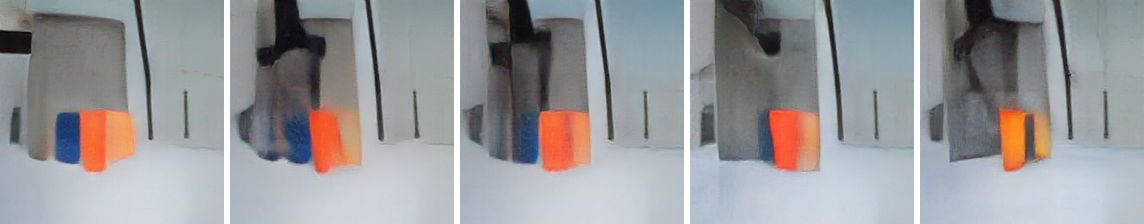} \\[1pt] \rotatebox{90}{\tiny\color{gray}wrist} & \includegraphics[width=\linewidth]{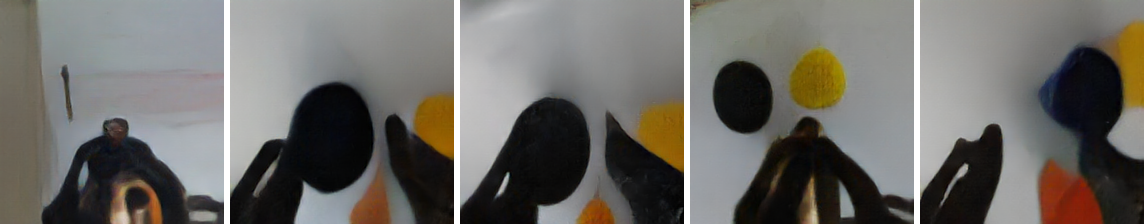}\end{tabular} \\ \noalign{\vskip 3pt}
\end{tabular}
\caption{\textbf{Real-world rollouts with MS-ILLM: failure cases} (continued from Figure~\ref{fig:realworld_msillm_failure}).}
\label{fig:realworld_msillm_failure_4}
\end{figure*}

 % MS-ILLM failures
\clearpage
\newpage
\section{Failure Case Analysis} \label{sec:failure_case}
\paragraph{SPARC rarely fails outright.}
On the two pick-and-place tasks, SPARC fails the first stage in only 2 of 80 attempts, against 31 of 80
for MS-ILLM. This matches the uncompressed policy, which also never fails the first stage there. SPARC's
remaining failures on these tasks are therefore second-stage failures, the same kind the uncompressed
policy makes in 40--50\% of its attempts. The gap is widest at the lowest bitrate: MS-ILLM fails the first task in 28 of 30 attempts, and SPARC in 9 of 30, 8 of which are on \textit{stack cups}. SPARC reaches this
while spending 3--19\% fewer bits than MS-ILLM at the three higher quality levels.

\paragraph{Failure modes.}
We inspected the decoded frames of every episode and grouped the failures. SPARC's failures are concentrated in the precise placement of \textit{stack cups}, where the gripper must line up a cup rim with another in the side view. This task is also the hardest for the uncompressed policy (30\% zero-score attempts), so part of this failure rate reflects the policy rather than the codec. MS-ILLM's failures instead extend to the first stage of every task and, at the lowest bitrate, to losing the scene altogether.
\newpage

\section{Additional Visual Examples}\label{sec:ave}
Following \Cref{ssec:qualitative}, we provide additional visualization examples for tasks where the average performance gain of SPARC is high or low. For both visualizations, the bit allocation map shows that SPARC allocates more bits to task-relevant regions (i.e. near the microwave and the gripper's end-effector), supporting the analysis in \Cref{sec:fca}. In addition, SPARC also tends to allocate more bits to objects that are generally not used in other tasks. This may imply that allocating more bits to static objects provides reference points and is required for recognizing distinctive scene features in simulation environments where object locations are randomized. Overall, these examples show that SPARC is capable of allocating more bits to task-centric regions to improve the performance of VLA models.
% Requires \usepackage{graphicx}. Copy Figures/ alongside your main .tex.
% Use figure* instead of figure for page-wide figures in a two-column paper.
% Include either the eight-example or all-example file, not both.
\begin{figure}[!htbp]
    \centering
    \includegraphics[width=\linewidth]{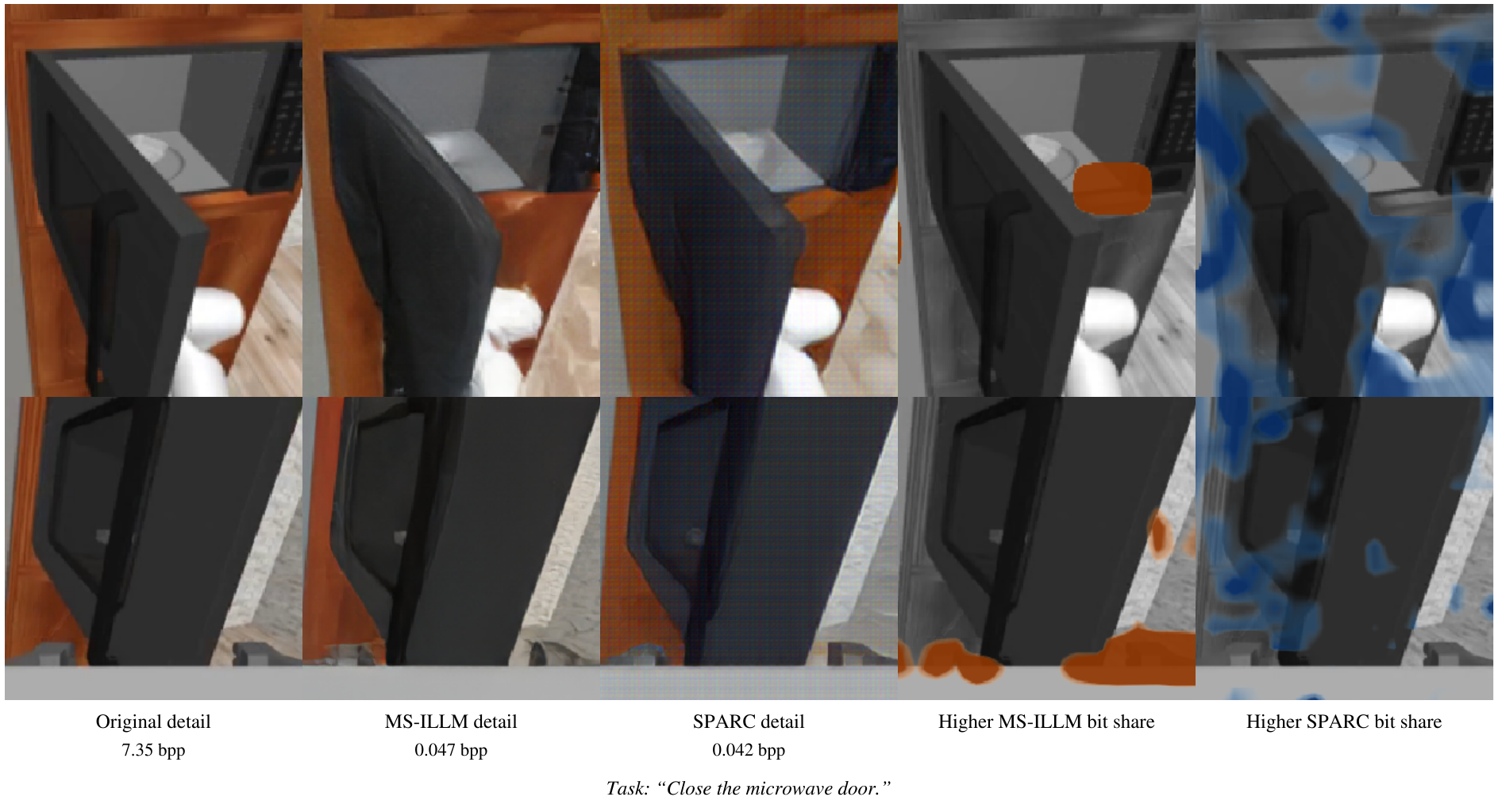}
    \caption{CloseMicrowave,``Close the microwave door.''.}
    \label{fig:qual_bidirectional_adaptive_closemicrowave_q1_detail}
\end{figure}

\begin{figure}[!htbp]
    \centering
    \includegraphics[width=\linewidth]{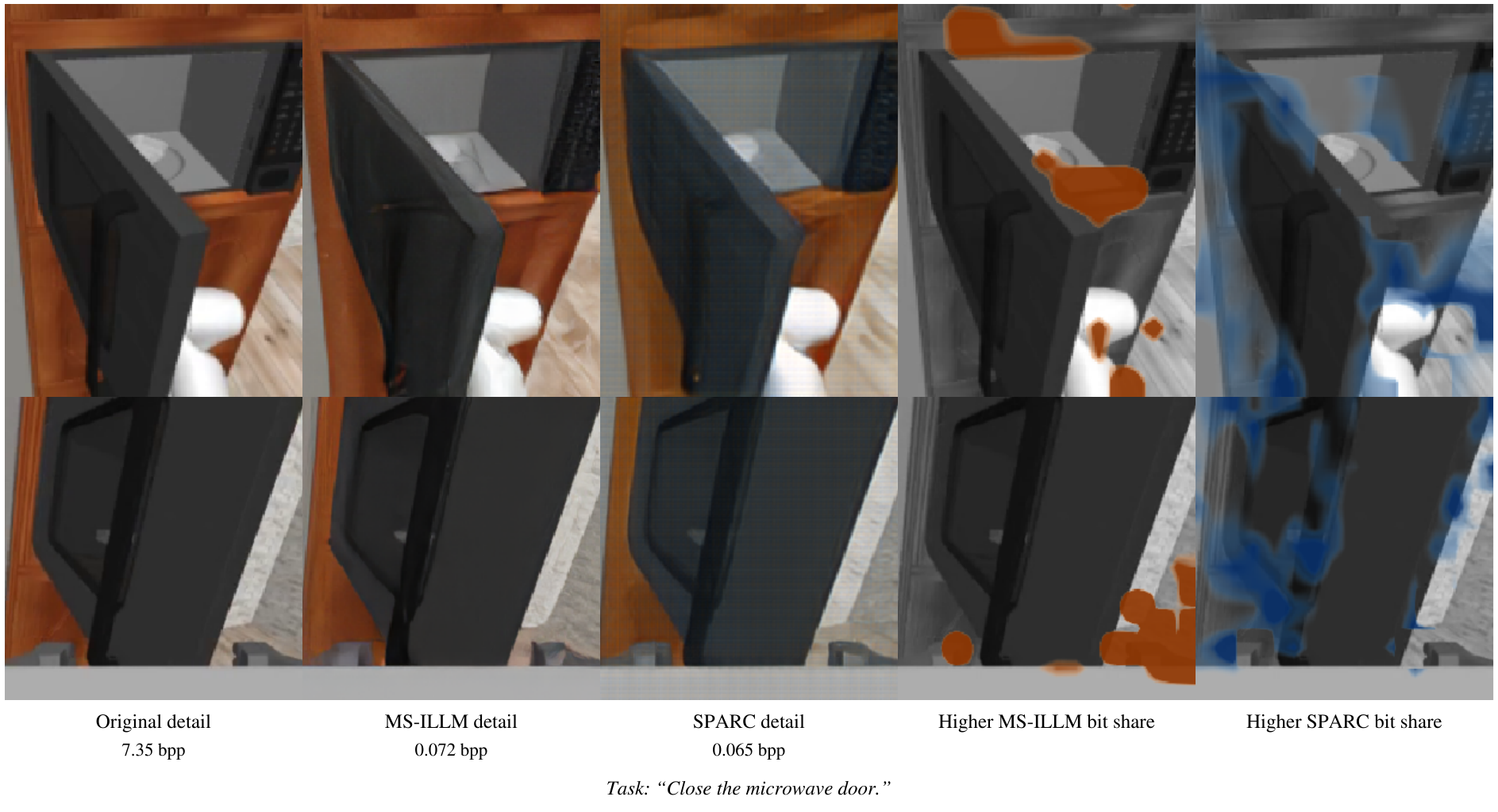}
    \caption{CloseMicrowave, ``Close the microwave door.''.}
    \label{fig:qual_bidirectional_adaptive_closemicrowave_q2_detail}
\end{figure}

\begin{figure}[!htbp]
    \centering
    \includegraphics[width=\linewidth]{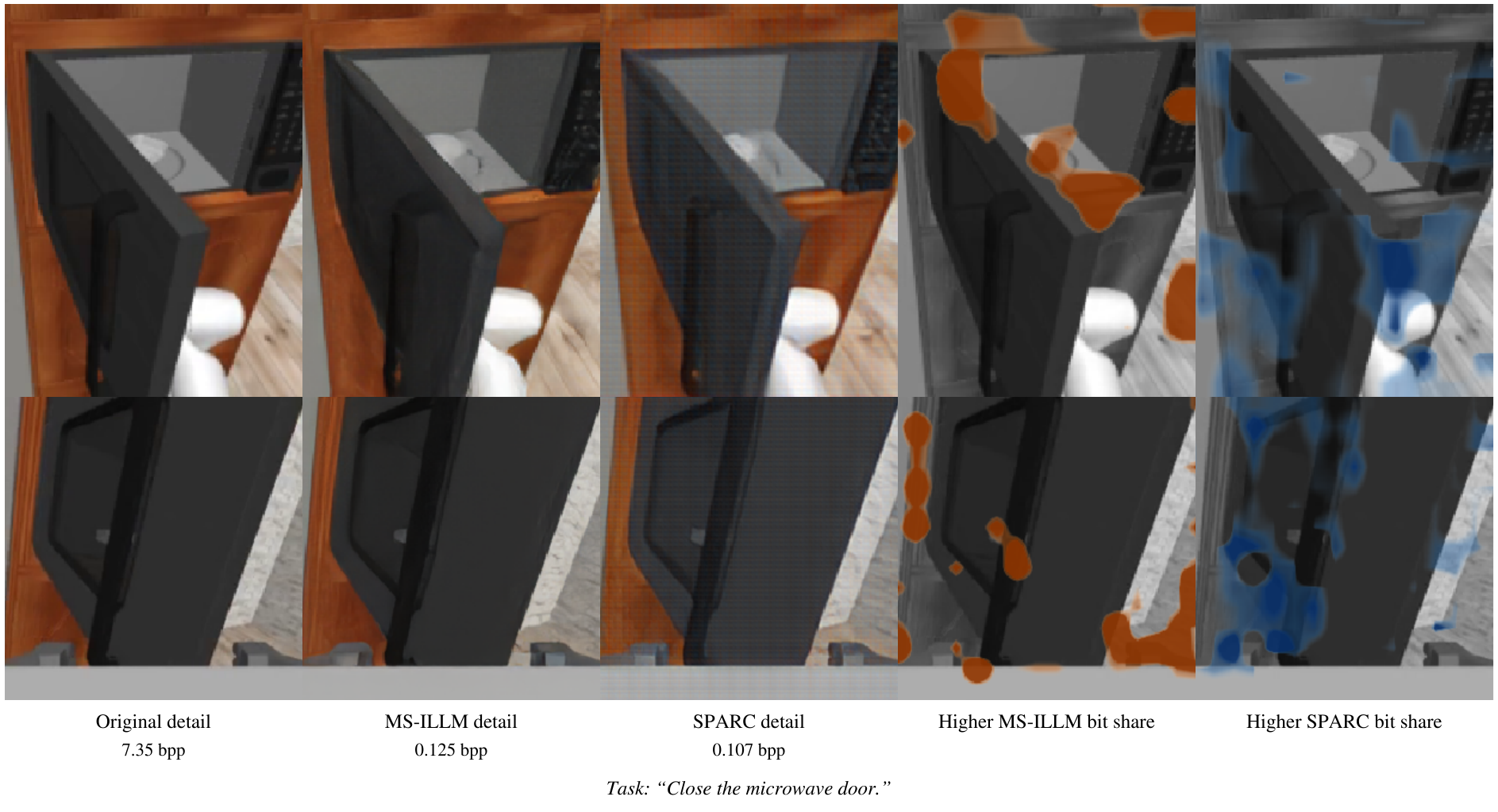}
    \caption{CloseMicrowave, ``Close the microwave door.''.}
    \label{fig:qual_bidirectional_adaptive_closemicrowave_q3_detail}
\end{figure}

\begin{figure}[!htbp]
    \centering
    \includegraphics[width=\linewidth]{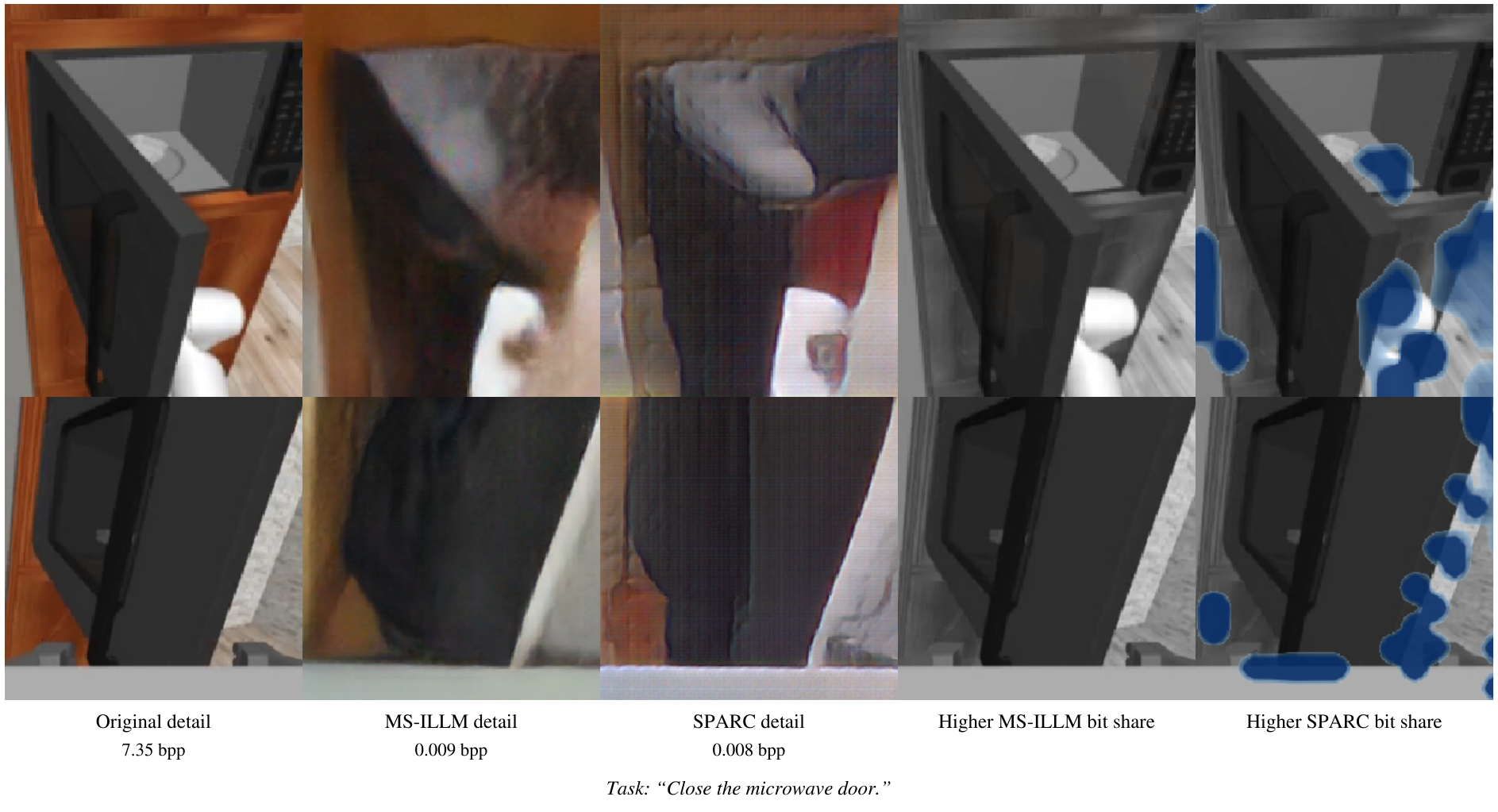}
    \caption{CloseMicrowave, ``Close the microwave door.''.}
    \label{fig:qual_bidirectional_adaptive_closemicrowave_qvlo1_detail}
\end{figure}

\begin{figure}[!htbp]
    \centering
    \includegraphics[width=\linewidth]{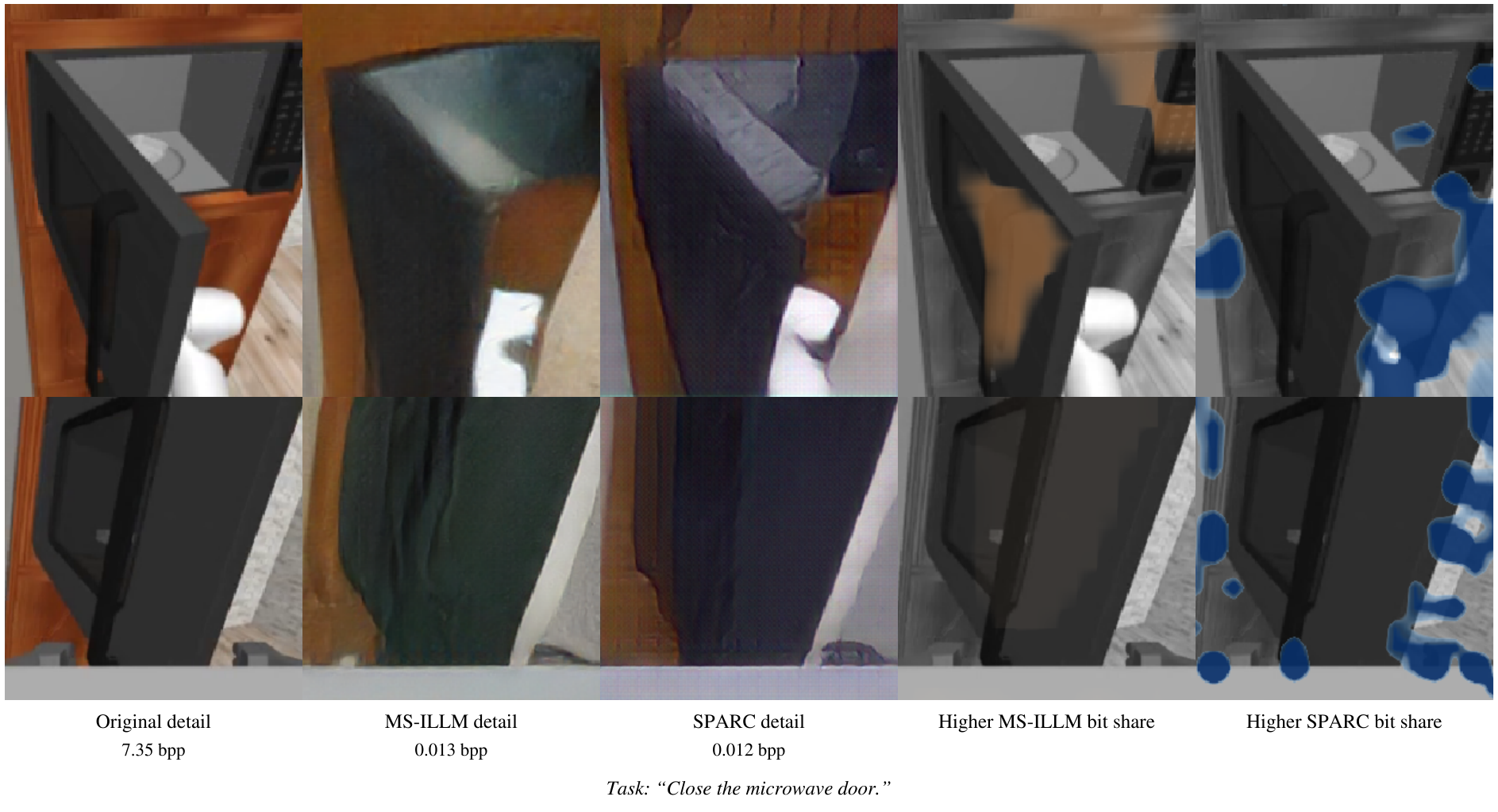}
    \caption{CloseMicrowave, ``Close the microwave door.''.}
    \label{fig:qual_bidirectional_adaptive_closemicrowave_qvlo2_detail}
\end{figure}

\begin{figure}[!htbp]
    \centering
    \includegraphics[width=\linewidth]{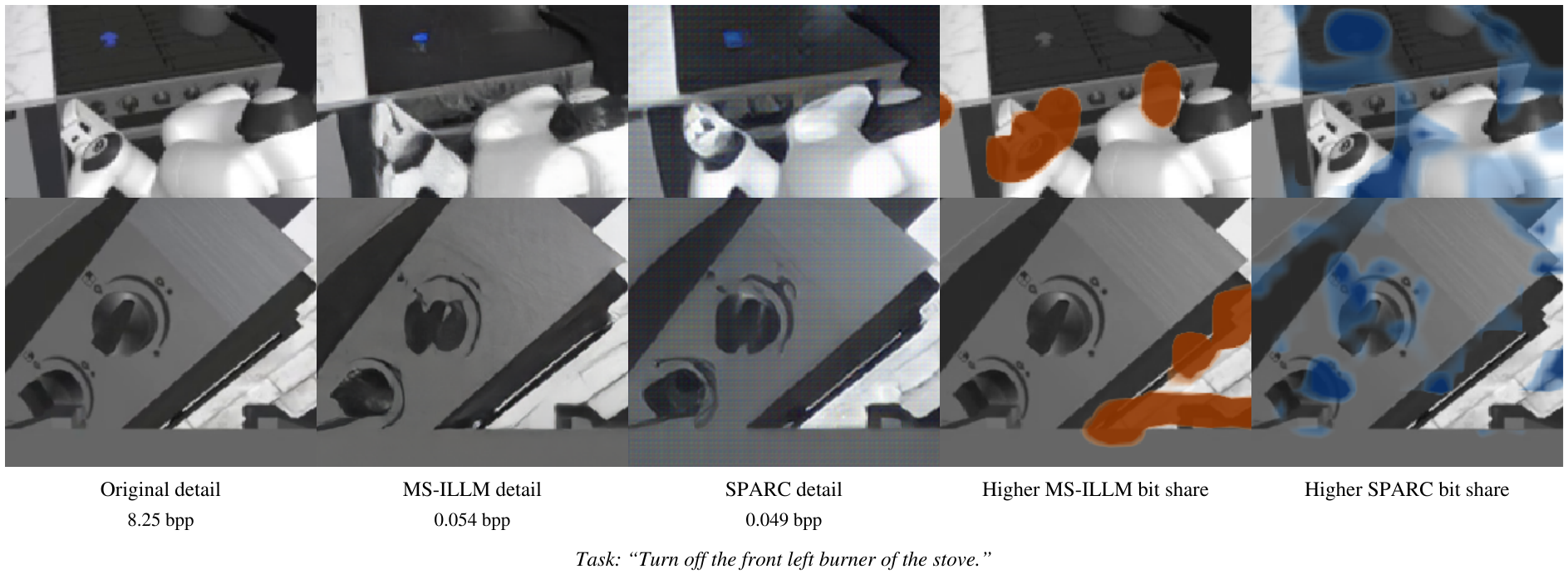}
    \caption{TurnOffStove, ``Turn off the front left burner of the stove.''.}
    \label{fig:qual_bidirectional_adaptive_turnoffstove_q1_detail}
\end{figure}

\begin{figure}[!htbp]
    \centering
    \includegraphics[width=\linewidth]{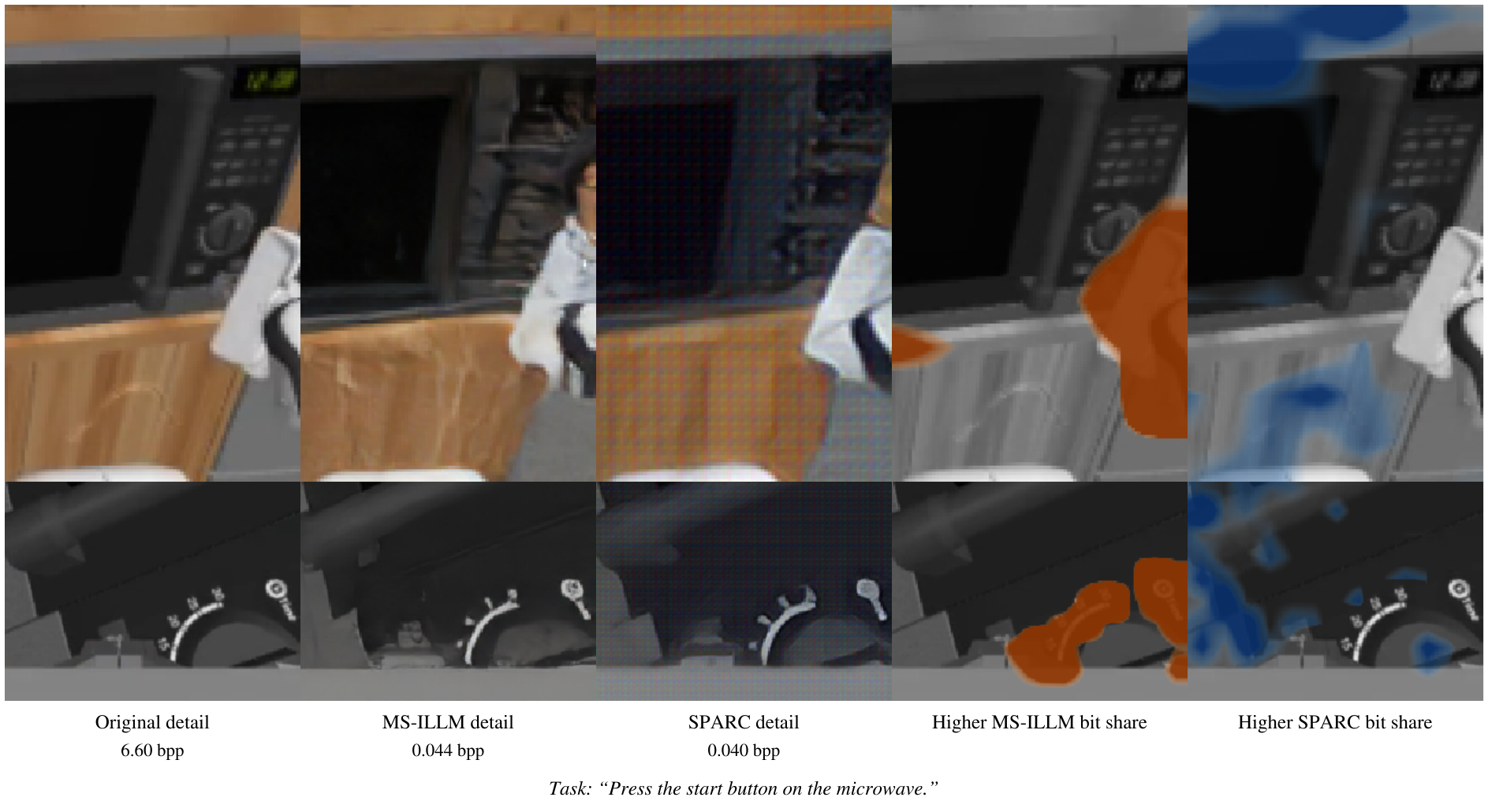}
    \caption{TurnOnMicrowave, ``Press the start button on the microwave.''.}
    \label{fig:qual_bidirectional_adaptive_turnonmicrowave_q1_detail}
\end{figure}

\begin{figure}[!htbp]
    \centering
    \includegraphics[width=\linewidth]{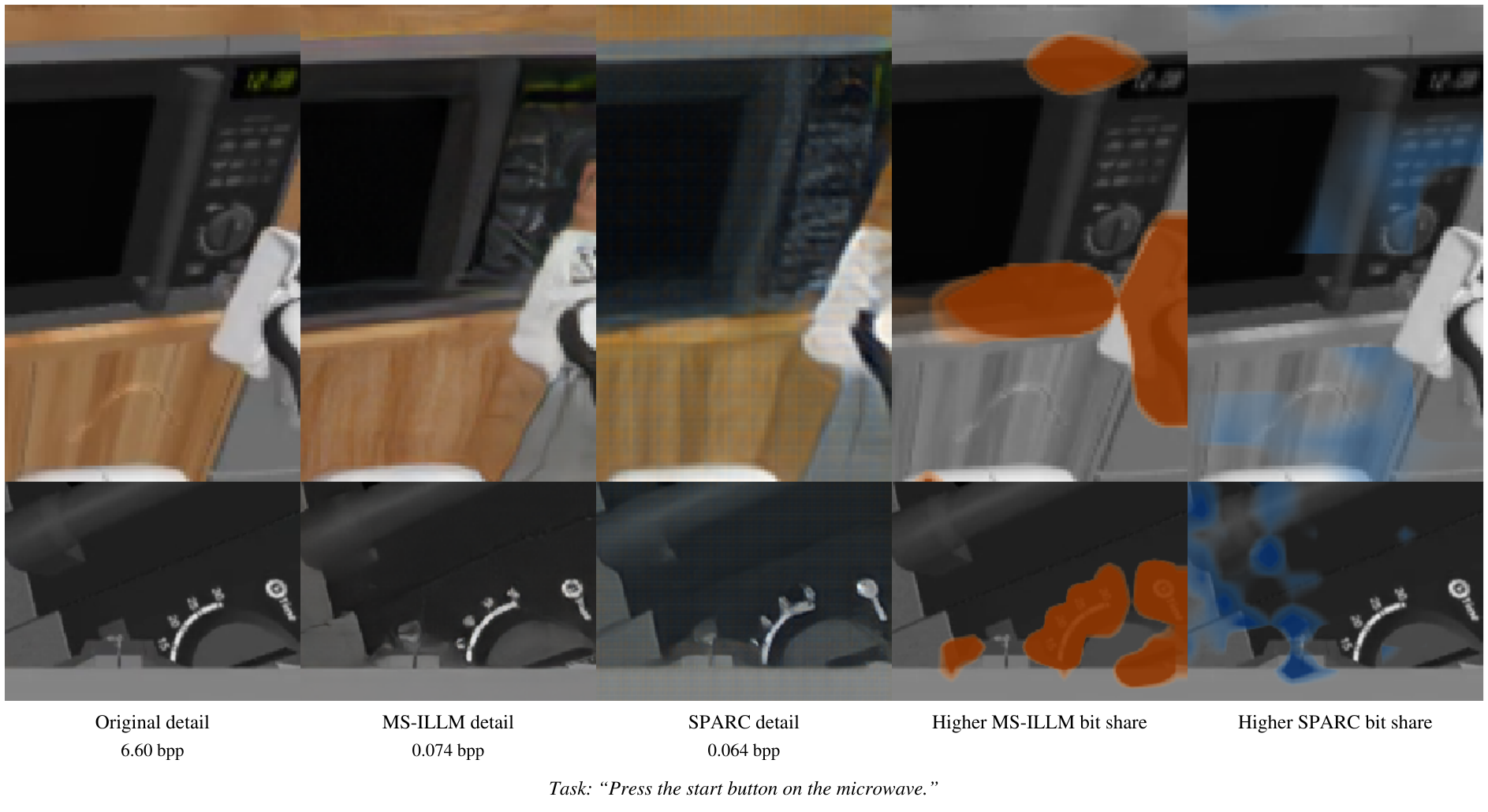}
    \caption{TurnOnMicrowave, ``Press the start button on the microwave.''.}
    \label{fig:qual_bidirectional_adaptive_turnonmicrowave_q2_detail}
\end{figure}

\begin{figure}[!htbp]
    \centering
    \includegraphics[width=\linewidth]{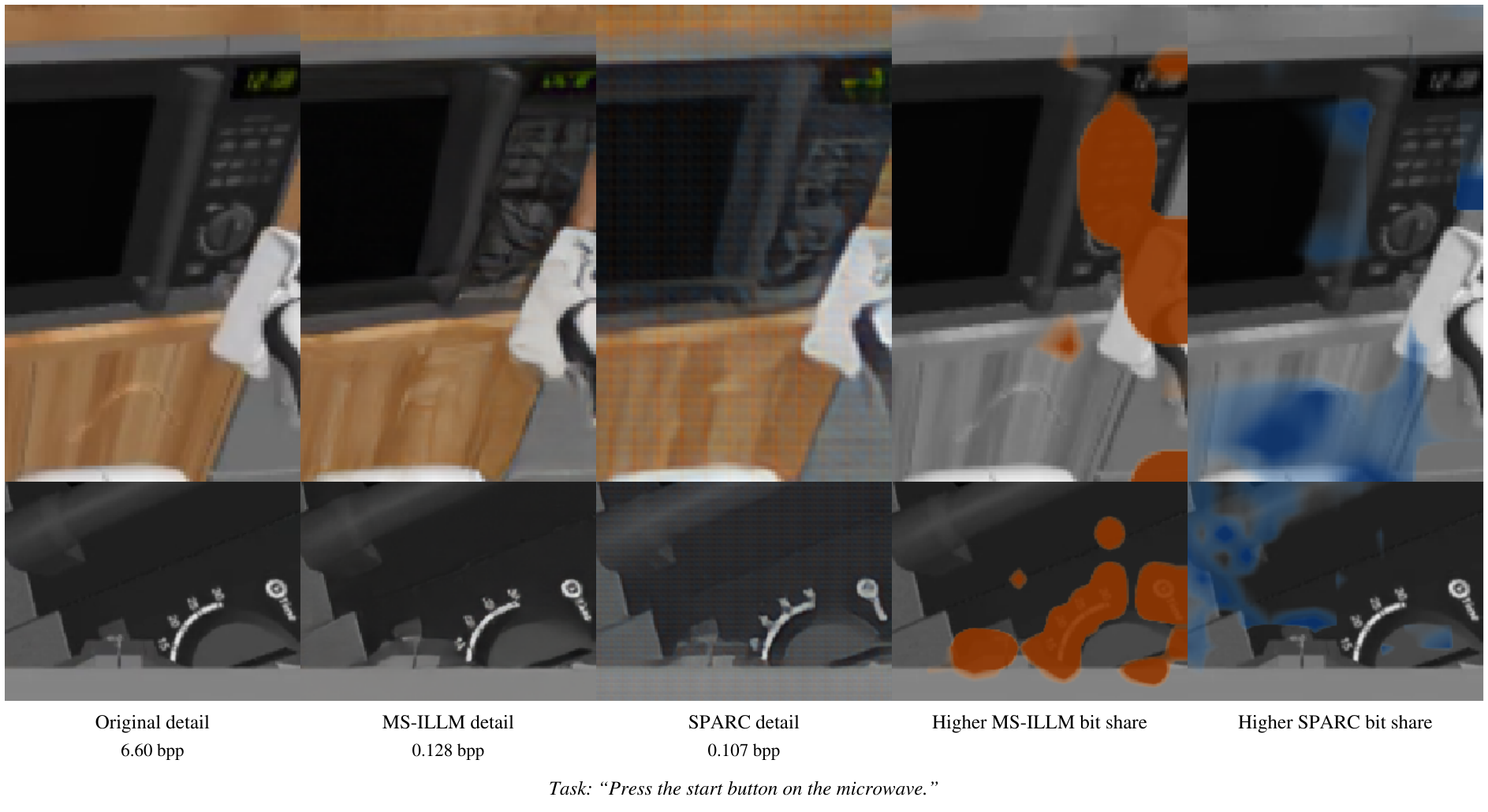}
    \caption{TurnOnMicrowave, ``Press the start button on the microwave.''.}
    \label{fig:qual_bidirectional_adaptive_turnonmicrowave_q3_detail}
\end{figure}

\begin{figure}[!htbp]
    \centering
    \includegraphics[width=\linewidth]{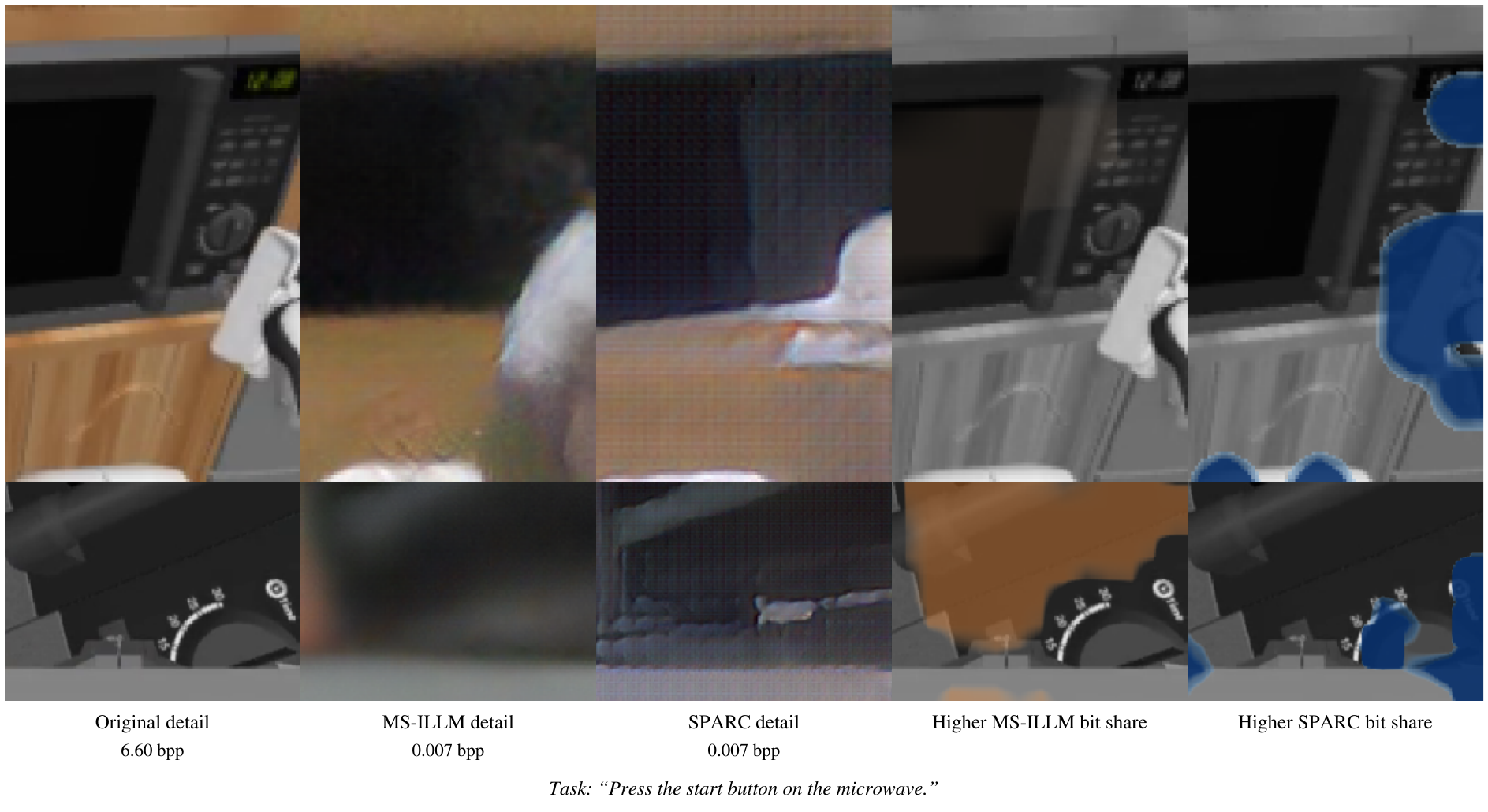}
    \caption{TurnOnMicrowave, ``Press the start button on the microwave.''.}
    \label{fig:qual_bidirectional_adaptive_turnonmicrowave_qvlo1_detail}
\end{figure}

\begin{figure}[!htbp]
    \centering
    \includegraphics[width=\linewidth]{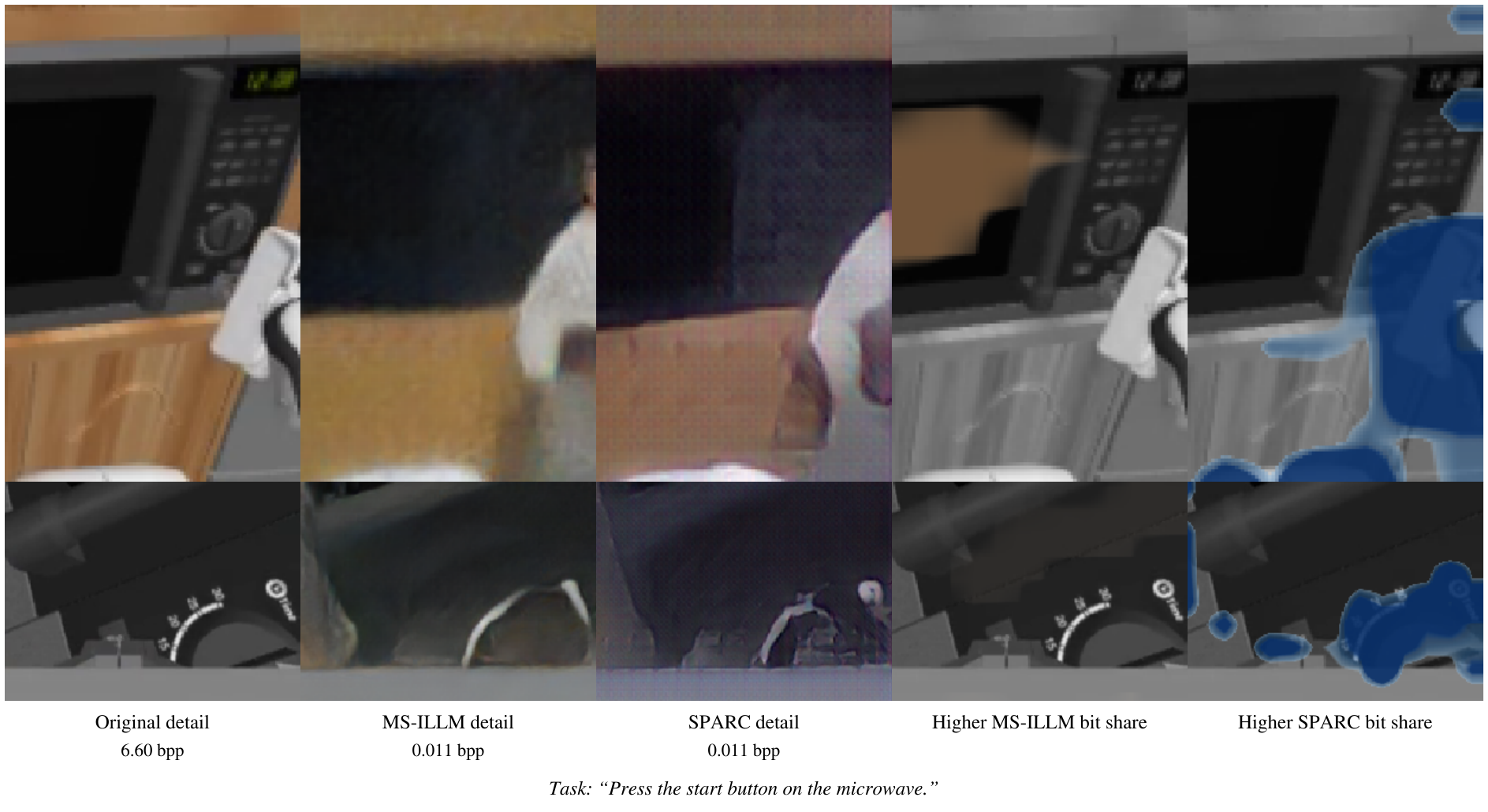}
    \caption{TurnOnMicrowave, ``Press the start button on the microwave.''.}
    \label{fig:qual_bidirectional_adaptive_turnonmicrowave_qvlo2_detail}
\end{figure}

\begin{figure}[!htbp]
    \centering
    \includegraphics[width=\linewidth]{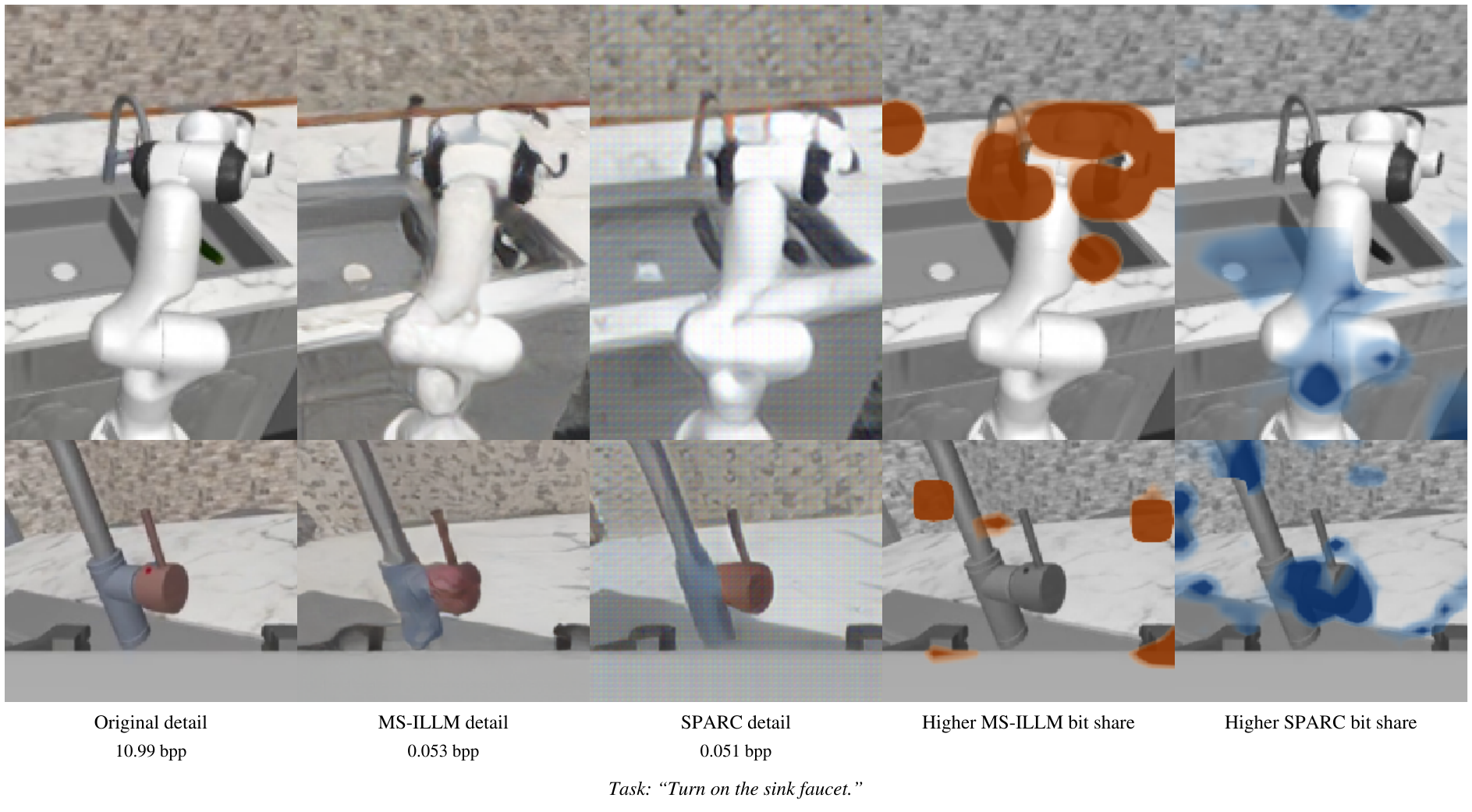}
    \caption{TurnOnSinkFaucet, ``Turn on the sink faucet.''.}
    \label{fig:qual_bidirectional_adaptive_turnonsinkfaucet_q1_detail}
\end{figure}

\clearpage

\end{document}